\documentclass[10pt,journal,compsoc]{IEEEtran}
\usepackage{helvet}
\usepackage{courier}
\usepackage{graphicx}
\usepackage{amsmath,amsthm,amssymb}
\usepackage{textcomp,booktabs}
\usepackage{footmisc}
\usepackage[normalem]{ulem}
\usepackage{multirow}
\usepackage{setspace}
\usepackage{bigstrut}
\usepackage{array}
\usepackage{amsfonts}
\usepackage{bm}
\usepackage{url}
\usepackage{balance}
\usepackage{mathrsfs}
\usepackage{adjustbox}
\usepackage{caption}
\usepackage{cuted}
\usepackage{capt-of}
\usepackage{xcolor, color, colortbl}
\usepackage{rotating}
\usepackage{commath}
\usepackage{algorithmic}
\usepackage[linesnumbered,ruled,vlined]{algorithm2e}
\usepackage{dsfont}
\usepackage{bbding}
\usepackage{threeparttable}
\usepackage{listings}
\usepackage{subfig}
\usepackage{makecell}

\usepackage{ulem}
\usepackage{bbm}

\makeatletter
\newcommand{\rmnum}[1]{\romannumeral #1}
\newcommand{\Rmnum}[1]{\expandafter\@slowromancap\romannumeral #1@}
\makeatother

\ifCLASSOPTIONcompsoc
  \usepackage[nocompress]{cite}
\else
  \usepackage{cite}
\fi

\ifCLASSINFOpdf
\else
\fi

\newcommand{\etal}{\textit{et~al.}\xspace}
\newcommand{\ie}{\textit{i.e.}\xspace}

\newcommand{\sg}{\text{sg}\xspace}

\newcommand{\fullmethod}{Generalized Robust Test-Time Adaptation\xspace}
\newcommand{\method}{GRoTTA\xspace}

\newcommand{\fullsetting}{Practical Test-Time Adaptation\xspace}

\newcommand{\setting}{PTTA\xspace}
\newcommand{\sampling}{CBS\xspace}

\newcommand{\argmax}{\mathop{\arg\max}}

\newcommand{\one}{\mathbbm{1}}
\newcommand{\onec}{\mathbbm{1}_\mathcal{C}}
\newcommand{\oneb}{\mathbbm{1}_B}

\newcommand{\bank}{\mathcal{M}}
\newcommand{\banksize}{\mathcal{N}}

\newcommand{\numclass}{\mathcal{C}}
\newcommand{\R}{\mathbb{R}}

\newcommand{\ptea}{\theta^{T}}
\newcommand{\pstu}{\theta^{S}}

\newcommand{\batch}{\bm{\mathcal{X}}}
\newcommand{\varL}{\mathcal{L}}
\newcommand{\st}{\text{s.t.}}

\newcommand{\VspaceBefore}{\vspace{-3mm}}
\newcommand{\VspaceAfter}{\vspace{-4mm}}
\newcommand{\imb}{\gamma}

\newcommand{\Figure}{Fig.\xspace}

\newcommand{\ul}[1]{\underline{#1}}

\definecolor{Gray}{gray}{0.9}

\definecolor{text01purple}{RGB}{168,119,200}

\definecolor{text01green}{RGB}{82,208,83}

\definecolor{text02red}{RGB}{211,41,15}
\newcommand{\textred}[1]{\textcolor{text02red}{#1}}

\definecolor{text02yellow}{RGB}{230,119,11}

\newcommand{\gain}[1]{\textbf{#1}}

\newcommand{\dtplus}[1]{\fontsize{6pt}{0.1em}\selectfont \textbf{\textred{#1}}}

\newcommand{\paragraphstart}[1]{\vspace{1.25mm} \noindent \textbf{#1}}

\begin{document}

\title{Generalized Robust Test-Time Adaptation in Continuous Dynamic Scenarios}

\author{Shuang Li, Longhui Yuan, Binhui Xie and Tao Yang
\IEEEcompsocitemizethanks{ 
\IEEEcompsocthanksitem S. Li, L. Yuan and B. Xie are with the School of Computer Science and Technology, Beijing Institute of Technology, Beijing, China. Email: \{shuangli, longhuiyuan, binhuixie\}@bit.edu.cn. \protect
\IEEEcompsocthanksitem
Tao Yang is with the College of Artificial Intelligence, Xi'an
Jiaotong University, Xi'an, China. Email: yt14212@stu.xjtu.edu.cn \protect
\IEEEcompsocthanksitem Corresponding author: Shuang Li. \protect
}
}

\IEEEtitleabstractindextext{%
\begin{abstract}
Test-time adaptation (TTA) adapts the pre-trained models to test distributions during the inference phase exclusively employing unlabeled test data streams, which holds great value for the deployment of models in real-world applications.
Numerous studies have achieved promising performance on simplistic test streams, characterized by independently and uniformly sampled test data originating from a fixed target data distribution.
However, these methods frequently prove ineffective in practical scenarios, where both continual covariate shift and continual label shift occur simultaneously, i.e., data and label distributions change concurrently and continually over time. 
In this study, a more challenging {\it \fullsetting (\setting)} setup is introduced, which takes into account the concurrent presence of continual covariate shift and continual label shift, and we propose a \fullmethod (\method) method to effectively address the difficult problem. 
We start by steadily adapting the model through {\it Robust Parameter Adaptation} to make balanced predictions for test samples.
To be specific, firstly, the effects of continual label shift are eliminated by enforcing the model to learn from a uniform label distribution and introducing recalibration of batch normalization to ensure stability.
Secondly, the continual covariate shift is alleviated by employing a source knowledge regularization with the teacher-student model to update parameters.
Considering the potential information in the test stream, we further refine the balanced predictions by {\it Bias-Guided Output Adaptation}, which exploits latent structure in the feature space and is adaptive to the imbalanced label distribution.
Extensive experiments demonstrate \method outperforms the existing competitors by a large margin under \setting setting, rendering it highly conducive for adoption in real-world applications.
\end{abstract}

}

\maketitle

\section{Introduction}
\label{sec:intro}

\begin{table*}[t]
  \centering
  \caption{Comparison between our proposed \fullsetting (\setting) and related adaptation settings. 
  }
  \label{table:settings}
  \VspaceBefore
  \resizebox{0.86\textwidth}{!}{
  \begin{tabular}{l | c c | c c | c c}
  \toprule[1.2pt]
  \multirow{2}{*}{Setting} & \multicolumn{2}{c |}{Adaptation Stage} & \multicolumn{2}{c |}{Available Data} & \multicolumn{2}{c}{Test Data Stream} \\
  
  \cline{2-7}
  
   & Train & Test & Source & Target & Data distribution & Label distribution \\

  \hline

  Domain Adaptation~\cite{GaninUAGLLML16,GDCAN} & \Checkmark & \XSolidBrush & \Checkmark & \Checkmark & - & - \\

  Domain Generalization~\cite{ZhouY0X21, LvLLZL23} & \Checkmark & \XSolidBrush & \Checkmark & \XSolidBrush & - & - \\
  
  Test-Time Training~\cite{ttt_sun2020} & \XSolidBrush & \Checkmark & \Checkmark & \Checkmark & $p(\bm{x})$ & Uniform $p(y|t) = 1/{\mathcal{C}}$ \\

  \hline

  Source Free Domain Adaptation~\cite{Liang22ShotPlus} & \multirow{4}{*}{\XSolidBrush} & \multirow{4}{*}{\Checkmark} & \multirow{4}{*}{\XSolidBrush} & \multirow{4}{*}{\Checkmark} & $p(\bm{x})$ & Uniform $p(y|t) = 1/{\mathcal{C}}$ \\

  Fully Test-Time Adaptation~\cite{tent_wang2020} &  &  &  &  & $p(\bm{x})$ & Uniform $p(y|t) = 1/{\mathcal{C}}$ \\

  Continual Test-Time Adaptation~\cite{cotta}  &  &  &  &  & $p(\bm{x}|t)$ & Uniform $p(y|t) = 1/{\mathcal{C}}$ \\

  Non-i.i.d. Test-Time Adaptation~\cite{niid_boudiaf2022parameter,note}  &  &  &  &  & $p(\bm{x})$ & One-hot $p(y|t) = [{\bm e}_{c(t)}]_y$ \\

  \hline

  \multirow{2}{*}{\bf \fullsetting (Ours)}& \multirow{2}{*}{\XSolidBrush} & \multirow{2}{*}{\Checkmark} & \multirow{2}{*}{\XSolidBrush} & \multirow{2}{*}{\Checkmark} & $p(\bm{x}|t)$ & One-hot $p(y|t) = [{\bm e}_{c(t)}]_y$ \\

  &  &  &  &  & $p(\bm{x}|t)$ & No Assumption $p(y|t)$ \\
  
  \bottomrule[1.2pt]
  \end{tabular}
  }
  \vspace{-4mm}
\end{table*}

\begin{figure*}[h]
    \centering
    \includegraphics[width=0.95\linewidth]{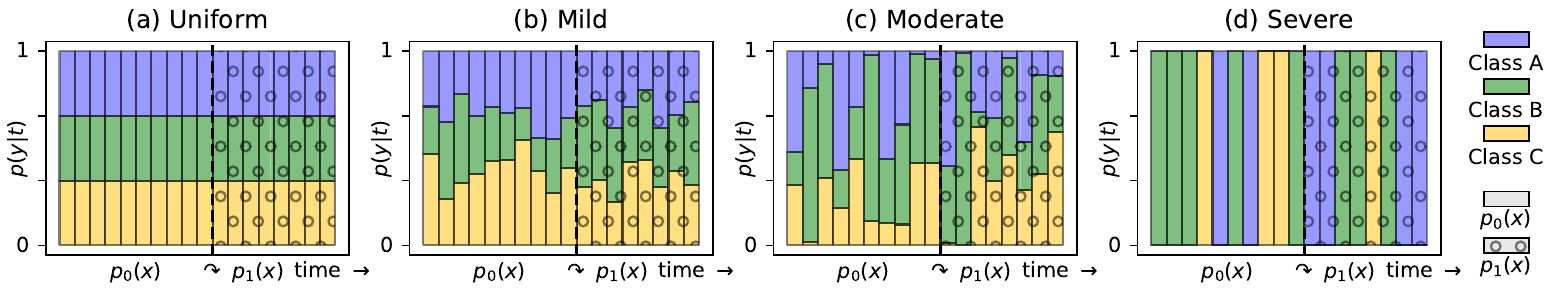}
    \VspaceBefore
    \caption{We consider the \fullsetting, where both continual covariate shift and continual label shift occur simultaneously. Notably, we make no assumptions on the label distribution so that \setting encompasses all of the following scenarios. (a) The most trivial situation is the continual TTA, where only data distribution changes with time while label distribution remains uniform and stationary. (b)/(c) Continual covariate shift with {\bf mild/moderate} continual label shift, corresponding to {\bf mild/moderate} imbalance in and changes of the label distribution. (d) Continual covariate shift with label temporal correlation, which is an extreme case of continual label shift, \ie, label distribution is $p(y|t) = [{\bm e}_{c(t)}]_y$.}
    \label{fig:setting}
    \VspaceAfter
\end{figure*} 

\IEEEPARstart{D}{eep} Neural Networks (DNNs) have achieved remarkable performance in various computer vision
tasks, such as classification~\cite{resnet, DosovitskiyB0WZ21}, object detection~\cite{ren2015faster, ZhuSLLWD21DETR}, and segmentation~\cite{chen2018deeplab, XieWYAAL21SegFormer}.
A crucial precondition for attaining such great achievements lies in the underlying assumption that both training and test data emanate from identical or similar data distributions.
Nonetheless, in practical scenarios, the occurrence of distribution shift~\cite{quinonero2008dataset} is unavoidable, primarily caused by factors like dissimilarities among sensors or alterations in the surrounding environment. Consequently, the effectiveness of deployed models diminishes considerably when confronted with such domain shifts, thereby restricting their applicability~\cite{MDD}.

To mitigate the distribution shift, an extensive body of research has concentrated on the field of transfer learning~\cite{survey}, specifically in the realms of domain adaptation (DA)~\cite{wang2018deep,GaninUAGLLML16,DAN-PAMI,GDCAN,LiangHF20}, and domain generalization (DG)~\cite{zhou2022domain,wang2022generalizing,MuandetBS13}. 
While these methods have exhibited notable advancements in performance, some non-negligible limitations remain, impacting the applicability of DA/DG in real-world scenarios.
Firstly, domain adaptation requires the use of test data during training, but it is often impractical to obtain test samples in advance.
Secondly, given the unpredictable distribution shift in real-life situations, there are considerable challenges in utilizing limited source domains to address the problem within the DG setup.
More importantly, these approaches not only impose substantial resource requirements but also present challenges in applications involving sensitive or proprietary source data, mainly due to concerns related to privacy and security~\cite{LiangHF20}.

In a more pragmatic context, Test-Time Adaptation (TTA) attempts to address the distribution shift at test time with only unlabeled test data, encompassing both offline~\cite{YangWWHJ21, NelakurthiMH18,Liang22ShotPlus} and online test streams~\cite{ttt_sun2020,NiuW0CZZT22,huang2022extrapolative,IwasawaM21,gandelsman2022testtime,zhang2022memo,shu2022testtime}.
Prior TTA studies~\cite{tent_wang2020,ttt_sun2020,chen2022_contrastivetta,goyal2022test} mostly concentrate on a simple adaptation scenario, where test samples are independently and uniformly sampled from a fixed target data distribution, which is denoted as $p(\bm{x})$. 
However, such an assumption is frequently violated in real-life applications.
Here, we take the environmental perception in autonomous driving as an example. 
On one hand, the distribution of perception data undergoes continuous changes as the surrounding environment of the car alters due to factors such as weather, location, and other variables.
To address this issue, Wang~\etal~\cite{cotta} propose the continual test-time adaptation setup, where the test data stream originates from a continually changing data distribution, \ie, $p(\bm{x}|t)$\footnote{In this paper, we follow the same description as~\cite{cotta}, in which $p(\bm{x}|t)$ is represented as a sequence of changing distributions $p_0(\bm{x}), p_1(\bm{x}), \cdots$. That means in a specific period, the test data are sampled from an unchanging data distribution. A similar interpretation can also be utilized for understanding the following $p(y|t)$.}.
On the other hand, the label distribution encountered by the car may also change continually over time, symbolized as $p(y|t)$.
For instance, the car tends to follow a greater number of vehicles on the highway and comes across an increased frequency of pedestrians on urban streets.
Inspired by the challenge, Boudiaf \etal~\cite{niid_boudiaf2022parameter} and Gong \etal~\cite{note} consider a special case of continually changing label distribution, \ie, the label temporal correlation among test samples.
In this scenario, test samples from the same category consistently appear over a period, resulting in the label distribution presenting as a changing one-hot encoding ${\bm e}_{c(t)}$, where $c(t)$ represents the only appearing category at time $t$.

Although these efforts aim to deal with the original simplistic scenario from two perspectives, they often narrow their focus solely on addressing either the issue of continual covariate shift (\ie, continual changes in the data distribution, $p({\bm x}|t)$) or continual label shift (\ie, continual changes in the label distribution, $p(y|t)$), while disregarding the other aspect.
However, from the aforementioned example in autonomous driving, even in a broader range of practical applications, it is evident that continual covariate shift and continual label shift often occur simultaneously when the models are deployed in reality. 

Building on the previous discussion, we introduce a more demanding and realistic adaptation setting called {\it \fullsetting (\setting)} in~\cite{rotta}. 
In PTTA, both continual covariate shift and label temporal correlation are present, making it more challenging.
However, the label temporal correlation assumes that the label distribution is a changing one-hot encoding, which only represents a special case of continual label shift.
To be more general, this work removes this assumption and extends \setting to performing adaptation on a more generalized test data stream where both continual covariate shift and continual label shift are taken into consideration.
To better illustrate our \setting setting, we have provided a concise and visual representation in \Figure~\ref{fig:setting}.
As we can see, the most trivial scenario of continual TTA~\cite{cotta} is shown in \Figure~\ref{fig:setting}(a), where only the data distribution changes continually while the label distribution remains uniform and stationary. 
In contrast, in \Figure~\ref{fig:setting}(b), (c), and (d), both the data distribution and the label distribution change over time, aligning more suitably with realistic requirements.
Furthermore, moving from (b) to (c), and then to (d), we distinctly observe that $p(y|t)$ becomes increasingly imbalanced and changes more dramatically, corresponding to different potential real-world scenarios.
And the most extreme case (d) represents the aforementioned specific instance in~\cite{rotta}, where the continual label shift turns to be label temporal correlation.
Meanwhile, detailed distinctions between \setting and previous setups are summarized in Table~\ref{table:settings}.
From the comparison, we can obtain that \setting is a more practical and valuable problem.

Actually, in the context of this practical test stream, we will face the following challenges when performing test-time adaptation. 
(1) Local overfitting. 
It arises from overfitting to the locally biased label distribution $p(y|t)$.
More concretely, categories that are dominant in earlier test periods might become minority classes in later times, and overfitting to it will inevitably hinder subsequent adaptation. 
To surmount the problem, we force the model to learn a uniform label distribution.
(2) Ineffectiveness of Batch Normalization (BN) layer. 
In particular, the biased label distribution $p(y|t)$ leads to inaccuracies in the statistics computed from test batches.
We design robust global statistics to conquer this issue.
(3) Error accumulation.
In long-term adaptation like \setting, if adopting entropy minimization or pseudo labeling, errors of the model will accumulate continually until the model collapses, leading to invalid adaptation.
To handle it, we resort to the more reliable teacher-student model for a steady parameter updating process.
(4) Catastrophic forgetting. 
In more detail, the model learns well on the current data distribution but sacrifices its generalization ability from source model on other data distributions.
We propose to distill knowledge from the source pre-trained model to effectively address this challenge.

Based on the aforementioned analysis, we propose the {\it \fullmethod} ({\bf \method}), which mainly consists of two parts: {\it Robust Parameter Adaptation} and {\it Bias-Guided Output Adaptation}.
Firstly, to guarantee the reliability of adaptation, we update the parameters of the model robustly and steadily to obtain balanced predictions of test samples.
More concretely, to prevent local overfitting, we establish a memory bank by {\it Category-Balanced Sampling} to simulate a uniform label distribution.
Meanwhile, to recover BN layers,  
we design the {\it Gradient-Preserving Robust Batch Normalization}, where the erroneous statistics of the current batch are replaced by the global ones maintained through exponential moving average and a gradient-preserving approach is adopted to ensure stability.
Subsequently, we introduce the {\it Robust Training without Forgetting}, which leverages the regularization based on source knowledge to address catastrophic forgetting and the teacher-student model to mitigate the accumulation of errors.
Secondly, to further exploit the potential of the test stream, we propose a {\it Bias-Guided Output Adaptation} module to align the balanced predictions with the current biased label distribution, which integrates latent structure information in feature space and batch-level bias reweighting strategy, further boosting the performance. 
With extensive experiments, we demonstrate the effectiveness of \method against the practical test streams in \setting.
 
In short, our contributions can be summarized as:
\begin{itemize}
    \item We introduce a new test-time adaptation setup that aligns better with real-world applications, namely \fullsetting (\setting). \setting considers both continual covariate shift and continual label shift, making it more practical and challenging.
    \item We first propose Robust Parameter Adaptation, which achieves robust and balanced adaptation on dynamic test streams and addresses the key challenges of \setting comprehensively and effectively.
    \item The Bias-Guided Output Adaptation module is presented, which further improves the performance by integrating latent structure information. Furthermore, the module is hot-swappable, providing our method with significant flexibility.
    \item Extensive experiments demonstrate the effectiveness of \method, which encompass three common TTA benchmarks, \ie, CIFAR-10-C, CIFAR-100-C, and ImageNet-C and three domain generalization dataset, \ie, PACS, OfficeHome, and DomainNet. 
    \method consistently achieves state-of-the-art results and obtains 14.6\%, 17.5\%, 10.5\%, 6.1\%, 4.5\%, and 9.9\% performance gain in the respective order.
\end{itemize}

A preliminary version of this work was presented in the conference paper~\cite{rotta}. 
In this extension we mainly make the following improvements: 
(1) \fullsetting (\setting) is extended to a more general setup, where continual covariate shift and continual label shift occur simultaneously. The concept of correlation among test samples mentioned in~\cite{rotta} only represents a specific instance of continual label shift.
(2) We advance each component in~\cite{rotta} into more comprehensive counterparts, including (\rmnum{1}) the category-balanced sampling with timeliness and uncertainty in~\cite{rotta} is simplified into category-Balanced Sampling to obtain a more balanced estimation of test distribution, (\rmnum{2}) we equip robust batch normalization with newly introduced gradient-preserving approach, and (\rmnum{3}) an additional regularization based on source knowledge is employed to mitigate the catastrophic forgetting problem.
(3) A novel bias-guided output adaptation is proposed to make full use of test streams. This module explores the latent structure information to refine the balanced predictions to align them with the biased label distribution, further boosting the performance.
(4) We further enlarge the experimental parts by evaluating \method on ImageNet-C, PACS and OfficeHome, and design comprehensive analysis to carefully verify the superiority of \method.

\section{Related Work}
\label{sec:related}
Test-Time Adaptation concentrates on adapting a source pre-trained model to the target domain during the inference phase~\cite{LiangHF20,ttt_sun2020,tent_wang2020, IwasawaM21}.
According to the training-test paradigm, a multitude of works generally fall into three categories: {\it source-free domain adaptation}, {\it test-time training} and {\it test-time adaptation}.

{\bf Source-Free Domain Adaptation  (SFDA)} intends to transfer the source pre-trained model to the target domain at test time, by leveraging all the test data in an offline manner.
A prominent technique within SFDA is pseudo-labeling. Take~\cite{LiangHF20} as an example, where Liang~\etal fit the source hypothesis by exploiting the information maximization and self-supervised pseudo-labeling. 
Distinctively, data generation plays a pivotal role in some other approaches.
Nayak \etal~\cite{NayakMJC22} perform Dirichlet modeling to estimate the source distribution and optimize the noisy input to generate synthetic data.
Meanwhile, consistency training also serves as a robust method in SFDA. In the work of~\cite{YangW0HJ21}, it maintains a memory bank to discover neighbors and minimize their inner product distances over predictions.
Some studies also leverage different techniques via clustering~\cite{Li2021ISFDA, LiuCDGHX22} and self-supervision~\cite{Liang22ShotPlus, KunduBKSJB22}.
With sufficient adaptation to the target domain, these approaches always perform impressively.
Nevertheless, offline learning entails significant computational resources and results in longer latency during inference, which is intolerable when requires immediate prediction.
In this work, we focus on performing adaptation with only access to test streams in an online manner, which is more efficient and flexible.

{\bf Test-Time Training (TTT)} introduces self-supervised auxiliary tasks during the training stage and optimizes them at test time to improve the performance of the source model~\cite{ttt_sun2020, GandelsmanSCE22, LiHDGW21, SuXJ22}.
In an example, Sun~\etal~\cite{ttt_sun2020} propose the pioneering work that adopts the rotation prediction~\cite{GidarisSK18} as the auxiliary task at both training and test phases to accomplish implicit alignment.
Later on, the work conducted by Liu~\etal~\cite{LiuKDBMA21} suggests that performing alignment to mitigate the domain gap promotes the effect of auxiliary tasks.
Considering the limitation of rotation prediction that is invalid for top-down views, many follow-up works make efforts to explore other self-supervision tasks. 
To name a few, Gandelsman~\etal~\cite{GandelsmanSCE22} exploit masked autoencoders (MAE)~\cite{HeCXLDG22} to conduct self-supervision, utilizing vision transformer backbones.
Osowiechi~\etal~\cite{OsowiechiHNCAD23} employ the unsupervised normalizing flows as an alternative auxiliary task.
On account of altering the training process with auxiliary tasks, these approaches always fail to be employed in privacy-restricted scenarios, where the source domain is unavailable and the model is training-agnostic.

{\bf Test-Time Adaptation (TTA)} strives to take advantage of the online unlabeled test streams to enhance the performance of the pre-trained model online, which has received widespread focus in recent years~\cite{tent_wang2020, tta_iclr1, LiuKDBMA21, ChiWYT21, KunduVVB20, royer2015classifier, Niu00WCZT23, IwasawaM21, EATA, song2023ecotta}.
A series of early studies are devoted to the recalibration of the statistics in batch normalization (BN)~\cite{BN} layers, which are subsequently adopted by many follow-up methods.
As an example, the prediction-time BN, proposed by Nado~\etal~\cite{BN_Stat}, recomputes the statistics for each test batch during inference.
Another line of effort is entropy minimization. 
Wang~\etal~\cite{tent_wang2020} minimize the entropy of the prediction of each test batch and update the BN layers online.
Afterwards, Niu~\etal~\cite{Niu00WCZT23} adopt reliable and sharpness-aware entropy minimization to obtain a smoother entropy surface.
Meanwhile, some works adopt pseudo-labeling for model updates. 
For instance, Goyal~\etal~\cite{goyal2022test} present the conjugate pseudo labels through the convex conjugate function. 
Jang~\etal~\cite{tta_iclr1} obtain pseudo labels from a prototype-based classifier and a neighbor-based classifier.
Due to the absence of label information, consistency regularization becomes one of the remaining reliable objectives. 
Therefore, Wang~\etal~\cite{cotta} force the consistency between the multi-augmented teacher output and the student output. In the meantime, a stochastic restoration process is designed to prevent catastrophic forgetting. 
Lately, D{\"o}ble~\etal~\cite{RMT} achieve the consistency between the teacher model and the student model through a symmetric cross-entropy loss. 
Moreover, some researches explore parameter-free~\cite{niid_boudiaf2022parameter} or parameter-efficient~\cite{vpt_tta} methods to achieve effective TTA.

While these efforts have made great strides, the considered distribution shift often exhibits a unilateral bias, \ie, only continual covariate shift or continual label shift. 
As an example, the test stream is independently drawn from a fixed data distribution~\cite{tent_wang2020} or a continually changing one~\cite{cotta}, while neglecting the label shift.
Subsequently, a specific instance of the label shift, temporal correlation, is taken into consideration in~\cite{note,niid_boudiaf2022parameter}, while the data distribution is stationary.
To be more realistic, in our \setting, the test stream is generated with both continual covariate shift and continual label shift. 
Therefore, these approaches often encounter significant difficulties due to their lack of comprehensive consideration of the two shifts in \setting.

In the latest research~\cite{ODS}, Zhou~\etal consider a similar setup as \setting, where the label distribution and data distribution only change at the same time.
In a real-world example, label distribution changes between driving on the road and waiting traffic light at the corner of the road, while the data distribution remains the same during this phase.
In short, the label distribution changes more frequently than the data distribution in reality, which makes the label distribution tracker and reweighting strategy in~\cite{ODS} ineffective.
In contrast, our \setting takes a more comprehensive consideration of continual covariate shift and continual label shift and launches \method to tackle them by robust model adaptation and bias-guided output adaptation.

\section{Method}
\label{sec:method}

\subsection{Problem Formulation and Preliminaries}
Given a model $f_{\theta_0}$ with parameter $\theta_0$ pre-trained on source domain $\mathcal{D}_{S} = \{(\bm{x}^S, y^S)\}$, the \fullsetting (\setting) is proposed to adapt $f_{\theta_0}$ on a stream of online unlabeled test samples $\batch_0, \batch_1, \batch_2, \cdots$, where $\batch_t$ is a batch of samples from the test distribution $\mathcal{P}_{test}$ with both continual covariate shift and continual label shift occurring simultaneously.
More specifically, at test time, the data distribution $p(\bm{x}|t)$ changes continually as $p_0(\bm{x}), p_1(\bm{x}), \cdots$.
Meanwhile, the label distribution $p(y|t)$ also changes continuously during the period when the data distribution turns to any $p_i(\bm{x})$.
For a more thorough understanding of PTTA, the detailed illustration is provided in \Figure~\ref{fig:setting}.
At test step $t$, we will receive a batch of unlabeled test samples $\batch_t$ with label distribution $p(y|t)$ from data distribution $p(\bm{x}|t)$.
Subsequently, $\batch_t$ is fed into the model $f_{\theta_t}$ and the model should adapt itself to the test stream and make predictions $f_{\theta_t}(\batch_t)$ on the fly.
Compared to existing TTA setups~\cite{tent_wang2020, cotta, niid_boudiaf2022parameter, note, ODS}, our \setting more comprehensively considers both data distribution and label distribution in continuously changing scenarios, which makes it more aligned with the requirements of real-world applications.

\begin{figure}[t]
    \centering
    \includegraphics[width=0.95\linewidth]{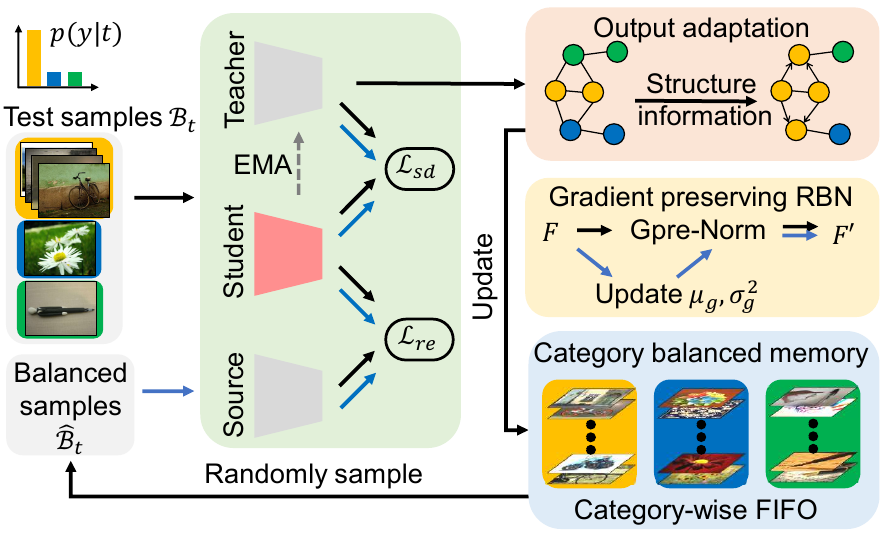}
    \VspaceBefore
    \caption{{\bf Framework overview.} We start by steadily adapting the model through {\it Robust Parameter Adaptation}. Specifically, we first replace the BN layers with GpreRBN to robustly normalize the feature maps. During the online adaptation on the test stream of \setting, a category-balanced memory bank is maintained and utilized to update the model in a robust and balanced manner. Finally, the structure information is leveraged by {\it Bias-guided Output Adaptation} to further refine the balanced predictions.}
    \label{fig:framework}
    \VspaceAfter
\end{figure}

\begin{figure*}[t]
    \centering
    \includegraphics[width=0.99\linewidth]{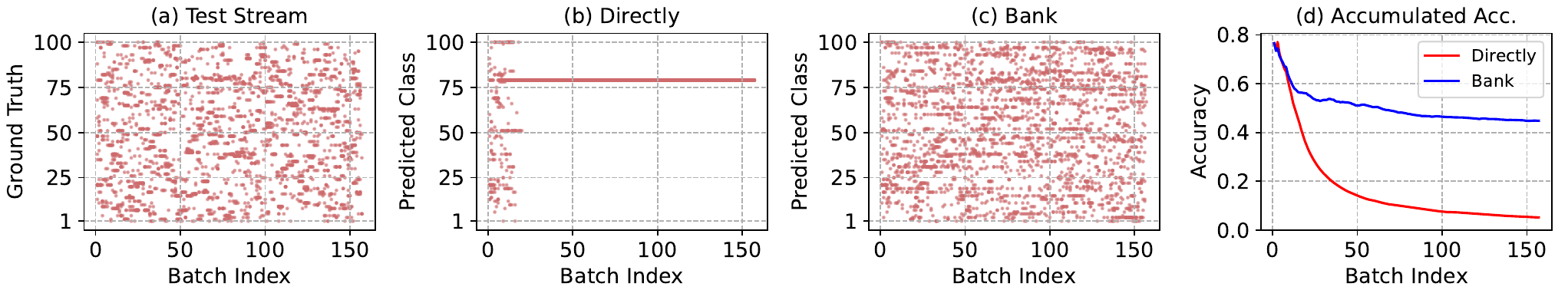}
    \VspaceBefore
    \caption{Experiments are conducted on ``motion blur'' corruption with severity 5 of CIFAR-100-C. (a) The true label distribution of the test stream with the continual label shift. (b) Predicted class from the model directly adapted to current test batches. (c) Predicted class of the test stream from the model updated by random batches from a category-balanced memory bank.
    (d) Accumulated accuracy during the online adaptation procedure of the two different models.}
    \label{fig:bank_analysis}
    \VspaceAfter
\end{figure*}
One of the naive approaches to tackle this complex challenge involves directly adapting the model to $\batch_0, \batch_1, \batch_2, \cdots$.
Nonetheless, with only unsupervised objectives available, attempting to directly adapt to the continually changing $p(\bm{x}|t)$ and $p(y|t)$ within the \setting often results in heightened instability compared to other TTA setups. 
This increased instability stems from the concurrent changes of two dynamic factors over time in \setting.
In light of this, we adopt an alternative approach that achieves robust and effective adaptation in \setting with two steps.
Firstly, the parameters of the model are adapted to the current data distribution $p(\bm{x}|t)$ while maintaining a balanced label distribution, which has been well studied in previous literature where only data distribution changes over time~\cite{cotta,RMT}.
Subsequently, building upon the model refined in the prior step, we proceed to adapt its balanced output to align with the current label distribution $p(y|t)$.
Based on this core concept, as demonstrated in \Figure~\ref{fig:framework}, we introduce \fullmethod (\method), which incorporates both robust parameter adaptation and bias-guided output adaptation.

\subsection{Robust Parameter Adaptation}
\subsubsection{Category-Balanced Sampling}
In the literature, a handful of methods adapt the source pre-trained model to the test stream directly, achieving promising performance when only covariate shift occurs~\cite{tent_wang2020, cotta, EATA, Niu00WCZT23}. 
However, when there exists continual label shift, directly adapting the model will suffer a lot from the local overfitting.
For a clearer understanding,  a toy example is conducted to demonstrate the effect of local overfitting and the results are presented in \Figure~\ref{fig:bank_analysis}.
From \Figure~\ref{fig:bank_analysis}(b), we observe that the model adapted directly eventually collapses. 
For detailed analysis, please refer to Section~\ref{chapter:analysis_bank}.
Some previous methods~\cite{ODS, Zhao0X23} attempt to estimate the current label distribution and perform online category reweighting to address the local overfitting problem.
Unfortunately, when the label distribution changes continually and frequently in \setting, the estimated label distribution becomes highly unreliable, further invalidating the online reweighting.
In a word, direct adaptation on test streams with continual label shift encounters significant difficulties.
Conversely, in \Figure~\ref{fig:bank_analysis}(c), when adapting the model through random batches of a memory bank structured as a category-wise one-instance First-In-First-Out
(FIFO) design, the model remains stable throughout the online adaptation process, without experiencing local overfitting or collapse.

Inspired by the analysis above, this paper proposes a simple but effective category-balanced memory bank $\bank$ with capacity $\banksize$, which compels the model to learn a uniform label distribution. 
More concretely, we maintain a FIFO queue with maximal length $\lceil \frac{\banksize}{\numclass} \rceil $ for each category\footnote{$\lceil \cdot \rceil$ represents the ceiling function, also known as ``round up''.}, \ie, Category-Balanced Sampling (\sampling).
For each update, a batch of samples will be randomly sampled from $\bank$. The sampling algorithm is summarized in Algorithm~\ref{alg:sampling}.

As far as we know, the two existing works~\cite{note, rotta} also solve the problem caused by label shift by maintaining a memory bank.
However, they focus more on meticulously maintaining a very small memory bank and utilizing the entire bank for updates.
In fact, when the number of categories significantly exceeds the size of the memory bank, it introduces another form of label shift, \ie, forcing the model to learn the categories present in the bank.
In contrast, we opt to maintain a larger memory bank with simple structure and update rules and train the model by randomly selecting a batch from it for each optimizing iteration.
This approach strikes a better balance among performance, memory usage, and computational resources.
Furthermore, the memory bank can be complemented with any other instance selection strategies to identify more informative test samples.

\begin{algorithm}[t]
    \caption{Category-Balanced Sampling (\sampling)} 
    \label{alg:sampling}
    \DontPrintSemicolon
    {\bf Input:} a test sample $\bm{x}$ and its predicted category $\hat{y}$. \\
    {\bf Define:} memory bank $\bank = \{\text{\it FIFO}_i\}_{i=1}^{\numclass}$ and its capacity $\banksize$, number of classes $\numclass$. \\
    \If{$|\text{\it FIFO}_{\hat{y}}| \ge \lceil \frac{\banksize}{\numclass} \rceil$}
    {
        Remove the oldest samples in $\text{\it FIFO}_{\hat{y}}$ until $|\text{\it FIFO}_{\hat{y}}| = \lceil \frac{\banksize}{\numclass} \rceil - 1$.
    }
    Add $\bm{x}$ into $\text{\it FIFO}_{\hat{y}}$.
\end{algorithm} 

\subsubsection{Gradient-Preserving Robust Batch Normalization}
Batch Normalization (BN)~\cite{BN} is a popular training technique that speeds up network training, enhances convergence, and stabilizes the process by addressing gradient issues.
Given the feature map $\bm{F} \in \R^{B\times C\times H\times W}$ as the input for a BN layer when training, the channel-wise mean $\bm{\mu} \in \R^C$ and variance $\bm{\sigma}^2 \in \R^C$ are calculated as follows:
\begin{align}
    \label{eq:mu_b}
    \mu_{(c)} &= \frac{1}{BHW}\sum_{b=1}^B \sum_{h=1}^H \sum_{w=1}^W F_{(b,c,h,w)}\,, \\
    \label{eq:sigma_b}
    \sigma^2_{(c)} &= \frac{1}{BHW}\sum_{b=1}^B \sum_{h=1}^H \sum_{w=1}^W (F_{(b,c,h,w)} - \mu_{(c)})^2\,.
\end{align}%
Then the feature map is normalized and refined in a channel-wise manner as 
\begin{align}
    BN(\bm{F};\bm{\mu},\bm{\sigma}^2) = \bm{\gamma}\frac{\bm{F}-\bm{\mu}}{\sqrt{\bm{\sigma}^2+\epsilon}} + \bm{\beta}\,,\label{eq:affine}
\end{align}%
where $\bm{\gamma},\bm{\beta}\in\R^C$ are learnable parameters in the layer and $\epsilon>0$ is a constant for numerical stability. 
Meanwhile, during training, the BN layer maintains a group of global running mean and running variance $(\bm{\mu}_s, \bm{\sigma}^2_s)$ for inference.

\begin{figure}[t]
    \centering
    \includegraphics[width=0.99\linewidth]{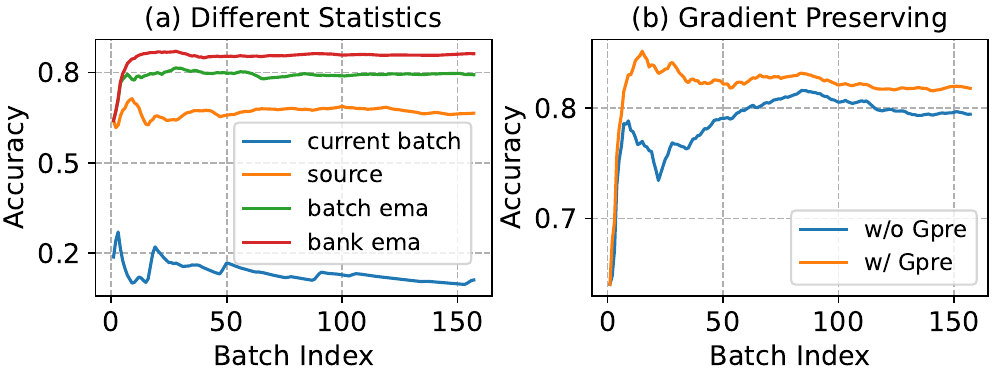}
    \VspaceBefore
    \caption{(a) Accumulated accuracy of different types of normalization statistics during online adaptation, where only statistics recalibration is performed. (b) Accumulated accuracy of models w/ and w/o gradient preserving approach when updating the model by entropy minimization on random batches from the category-balanced memory bank.
    }
    \label{fig:statistics_analysis}
    \VspaceAfter
\end{figure}

However, the statistics $(\bm{\mu}_s, \bm{\sigma}^2_s)$ may result in inaccurate normalization of test features due to the covariate shift during test time, leading to severe performance degradation.
Motivated by this, early works~\cite{BN_Stat,A_BN} try to correct the statistics $(\bm{\mu}_s, \bm{\sigma}^2_s)$ to test distributions at test time, without any updates on parameters. 
In the following works~\cite{tent_wang2020,cotta,EATA}, one of the most common techniques is adopting the statistics of the current batch of data to perform normalization.
Unfortunately, the continual label shift under \setting causes the statistics of the current batch to become untrustworthy as well.
A possible approach to acquire reliable statistics is utilizing the exponential moving average (EMA).
Nonetheless, because of the non-uniform and continually changing label distribution, statistics obtained directly by EMA on the test stream, as employed in~\cite{ManciniK0JC18}, will inevitably exhibit bias.
Fortunately, earlier maintenance of a category-balanced memory bank enables us to obtain highly reliable statistics through a simple EMA approach on it. 
To validate our claim, we compare four types of normalization statistics in \Figure~\ref{fig:statistics_analysis}(a).
The superiority of the statistics obtained by EMA on the memory bank supports our analysis.
We provide detailed analysis in the Section~\ref{chapter:analysis_statistic}.

Driven by the above analysis, we first propose a Robust Batch Normalization (RBN) module, which maintains a group of global statistics $(\bm{\mu}_g,\bm{\sigma}_g^2)$ to normalize the feature map robustly.
In specific, $(\bm{\mu}_g,\bm{\sigma}_g^2)$ is initialized as the running mean and variance $(\bm{\mu}_s,\bm{\sigma}_s^2)$ of the pre-trained model.
During the online adaptation, we update the global statistics by exponential moving average as 
\begin{align}
    \bm{\mu}_g  = (1-\alpha)\bm{\mu}_g + \alpha \bm{\mu}\,,\,\, \bm{\sigma}_g^2  = (1-\alpha)\bm{\sigma}_g^2 + \alpha\bm{\sigma}^2\,,
    \label{eq:global}
\end{align}%
where $\bm{\mu}, \bm{\sigma}^2 \in \R^{C}$ is the channel-wise mean and variance of a random batch from $\bank$, calculated by Eq.~\eqref{eq:mu_b} and~\eqref{eq:sigma_b}.
One thing to emphasize is that the above two formulas are tracked only for random batches from $\bank$, while for other samples, $(\bm{\mu}_g, \bm{\sigma}_g^2)$ are used for normalization directly without updates, which is the key point of RBN and differs from other EMA-based approaches like~\cite{ManciniK0JC18, Zhao0X23}.

While RBN achieves an excellent recalibration of the normalization statistics, it lacks consideration of gradient backpropagation in BN layers.
More specifically, given a random batch of test samples from $\bank$, its concrete forward procedure in the proposed RBN is
\begin{align}
    BN(\bm{F};\bm{\mu}_g,\bm{\sigma}_g^2) = \bm{\gamma}\frac{\bm{F}-((1-\alpha)\bm{\mu}_g + \alpha \bm{\mu})}{\sqrt{(1-\alpha)\bm{\sigma}_g^2 + \alpha\bm{\sigma}^2+\epsilon}} + \bm{\beta}\,.
    \label{eq:no_gpre}
\end{align}%
In the equation above, the gradient of either $BN(\bm{F};\bm{\mu}_g,\bm{\sigma}_g^2)$ w.r.t. $\bm{\gamma}$ or $\bm{F}$ is different from that of $BN(\bm{F};\bm{\mu},\bm{\sigma}^2)$, as encountered during the training process.
However, in test-time adaptation, keeping the gradient backpropagation form consistent with the training procedure is crucial.
To verify this claim, in \Figure~\ref{fig:statistics_analysis}(b), we compare two different methods: one that keeps the gradient the same as during training, denoted as w/ Gpre, and the other that does not, denoted as w/o Gpre.
The better and more stable performance of w/ Gpre conforms to our proposition.
And a detailed description is also provided in Section~\ref{chapter:analysis_statistic}.

Motivated by this, we equip the RBN with the gradient-preserving approach, namely GpreRBN.
Given the statistics $(\bm{\mu},\bm{\sigma}^2)$ of feature $\bm{F}$ with gradient information, we first preserve the gradient backpropagation form by
\begin{align}
    \bm{F}_{Gpre} = \frac{\bm{F} - \bm{\mu} + \sg(\bm{\mu})}{\sqrt{\bm{\sigma}^2 + \epsilon}} {\sqrt{\sg(\bm{\sigma}^2) + \epsilon}}\,,
    \label{eq:gpre}
\end{align}%
where $\sg(\cdot)$ is the stop gradient operation.
Then the final normalized and refined feature $\bm{F}'$ of $\bm{F}$ is calculated by
\begin{align}
    \label{eq:final_norm_refine}
    \bm{F}' = BN(\bm{F}_{Gpre};\bm{\mu}_g,\bm{\sigma}_g^2)\,,
\end{align}%
where $\bm{\mu}_g,\bm{\sigma}_g^2$ is the global statistics maintained by Eq.~\eqref{eq:global} without gradient information.
We summarize the proposed GpreRBN in Algorithm~\ref{alg:GpreRBM}.

\begin{algorithm}[t]
    \caption{Gradient Preserving RBN (GpreRBN)} 
    \label{alg:GpreRBM}
    \DontPrintSemicolon
    {\bf Input:} The forwarding feature map $\bm{F}$. \\
    {\bf Define:} The global mean and variance $(\bm{\mu}_g,\bm{\sigma}_g^2)$, {\it tracking} $\in$ \{{\it enable}, {\it disable}\} indicates whether to update the global statistics or not, respectively.\\
    Calculate $(\bm{\mu},\bm{\sigma}^2)$ of $\bm{F}$ through Eq.~\eqref{eq:mu_b} and~\eqref{eq:sigma_b}. \\

    Calculate $\bm{F}_{Gpre}$ through Eq.~\eqref{eq:gpre} with $(\bm{\mu},\bm{\sigma}^2)$. \\

    \If{tracking is enable}
    {
        Update $(\bm{\mu}_g,\bm{\sigma}_g^2)$ by Eq.~\eqref{eq:global} with $(\sg(\bm{\mu}),\sg(\bm{\sigma}^2))$.
    }
    Calculate output feature $\bm{F}'$ through Eq.~\eqref{eq:final_norm_refine}. \\
    \Return{$\bm{F}'$.}
\end{algorithm} 

\subsubsection{Robust Training without Forgetting}
Actually, after replacing
BN layers with our GpreRBN and obtaining the memory bank via \sampling, the negative effects of continual label shift have been largely resolved.
We can directly adopt widely used techniques like pseudo labeling or entropy minimization to perform test-time adaptation.
Nonetheless, due to the continual covariate shift, these simple unsupervised objectives often face the problem of error accumulation and catastrophic forgetting.
Specifically, due to the absence of label supervision, the model's errors will accumulate continually until the model collapses.
Meanwhile, during the continual learning process, the generalization ability from the source gradually diminishes, which presents a better performance on the current data distribution but extremely poor performance on future data distributions.
In addition, too aggressive updates of the model will make the category balance of $\bank$ and the global statistics maintained by EMA unreliable, resulting in
unstable adaptation.
To address the above concerns, we combine the source knowledge based regularization with the widely employed teacher-student model to update the model.
For the sake of time efficiency and stability, only affine parameters in GpreRBN are trained during adaptation.

At time step $t$, after inferring for the test batch $\batch_t$ with the teacher model $f_{\ptea_t}$ and updating the memory bank $\bank$ with $\batch_t$, we begin to update the student model $f_{\pstu_t}$ and the teacher model $f_{\ptea_t}$. 
As mentioned before, we can optimize the model on the random batches of $\bank$.
However, considering that some samples may not have the opportunity to participate in model updates before being removed from the bank, we update the parameters of student model $\pstu_t\to\pstu_{t+1}$ by minimizing the following loss:
\begin{align}
    \varL_{total} = \frac{1}{|\batch^{\bank}|}\sum_{\bm{x} \in \batch^{\bank}}\ell(\bm{x}) + \lambda_{batch} \frac{1}{|\batch_t|}\sum_{\bm{x} \in \batch_t}\ell(\bm{x})\,,
    \label{eq:total_loss}
\end{align}%
where $\batch^{\bank}$ is a random batch from $\bank$, $\batch_t$ is the current test batch and $\lambda_{batch}$ is a tradeoff parameter between them.
It's worth noting that, only one gradient descent step is performed for the loss at each time step $t$, which is consistent with previous methods.
Subsequently, the teacher model is updated by exponential moving average as
\begin{align}
    \ptea_{t+1} = (1-\nu)\ptea_{t} + \nu \pstu_{t+1}\,.
    \label{eq:teacher_eam}
\end{align}
To obtain the loss value of an instance $\bm{x} \in \batch_t \cup \batch^{\bank}$, we first calculate the self-distillation loss as following:
\begin{align} 
    \varL_{sd}(\bm{x}) = -\frac{1}{\numclass}\sum_{c=1}^{\numclass}p_T(c|\bm{x}')\log p_S(c|\bm{x}'') \,, \label{eq:self_distill}
\end{align}%
where $p_S(y|\bm{x}'')$ is the soft-max prediction of the strongly augmented view $\bm{x}''$ from the student model and $p_T(y|\bm{x}')$ is that of the weakly augmented view\footnote{Weak augmentation is ReSize+CenterCrop. Strong augmentation consists of
nine operations like ColorJitter, RandomAffine and so on.} $\bm{x}'$ from the teacher model. Then the regularization term based on source knowledge is computed as 
\begin{align}
    \varL_{re}(\bm{x}) = -\frac{1}{\numclass}\sum_{c=1}^{\numclass}p_A(c|\bm{x}')\log p_S(c|\bm{x}'') \,, \label{eq:regulariztion}
\end{align}
where $p_S(y|\bm{x}'')$ is the soft-max prediction of the strongly augmented view $\bm{x}''$ from the source pre-trained model.
Equipped with Eq.~\eqref{eq:self_distill} and~\eqref{eq:regulariztion}, the loss value of a single instance on the right-hand side of Eq.~\eqref{eq:total_loss} is calculated as 
\begin{align}
    \ell(\bm{x}) = \varL_{sd}(\bm{x}) + \lambda_{re}\varL_{re}(\bm{x})\,,
\end{align}%
where $\lambda_{re}$ makes a tradeoff between the mentioned terms.
To sum up,  the combination of CBS, GpreRBN, and robust training without forgetting enables us to adapt the model in a robust and balanced manner on the test stream of \setting.

\subsection{Bias-guided Output Adaptation}
\subsubsection{Latent Structure Information Exploration}
Through the robust parameter adaptation, we get a model with balanced output during the inference phase.
However, as the occurrence of continual label shift, there is still a deviation between the learned conditional probability and the true one of the test sample, which is formulated as 
\begin{equation}
    \label{eq:optimal_p}
    q(y|\bm{x}) = \frac{q(y)}{p(y)} p(y|\bm{x})\,,
\end{equation}%
where $p(y|\bm{x})$ and $p(y)$ are the learnt conditional probability and label distribution, while $q(y|\bm{x})$ and $q(y)$ are the true ones, respectively.
One of the most straightforward approaches is directly estimating the true label distribution $q(y)$ and adapting the balanced output according to Eq.~\eqref{eq:optimal_p}.
Unfortunately, the continually changing label distribution makes this approach unreliable.

Based on the fundamental premise of semi-supervised and unsupervised learning, which posits that samples within the same category should have similar features, we propose to explore the latent structure information to promote the refinements of outputs. 
More concretely, given a batch of test samples $\batch = [\bm{x}_1,\cdots,\bm{x}_B]^\top \in \R^{B\times d_x}$, where $B$ is the batch size and $d_x$ is the input dimension, and its balanced predictions $\bm{P} = [\bm{p}_1,\cdots,\bm{p}_B]^\top \in \R^{B\times \numclass}$ from the teacher model $f_{\ptea}$ at any time step, we attempt to optimize latent variables $\bm{Z} = [\bm{z}_1,\cdots,\bm{z}_B]^\top \in \R^{B\times \numclass}$ to obtain the refined predictions according to the following objective:
\vspace{-2mm}
\begin{align}
     \min_{\bm{z}_i,\cdots,\bm{z}_B} & \underset{\text{value consistency}}{\underbrace{(1 - \lambda)\sum_{i=1}^B D_1(\bm{z}_i, \bm{p}_i)}} + \underset{\text{similarity consistency}}{\underbrace{\lambda \sum_{i=1}^B D_2(\bm{w}_i, \bm{s}_i)}} \nonumber \\
     \st & ~~ {\bm z}_i^{\top}\mathbbm{1}_{\mathcal{C}} = 1, i=1, \cdots,B\,,
\end{align}
where $\lambda \in (0, 1)$ makes tradeoff between two dissimilarity functions $D_1(\cdot,\cdot),D_2(\cdot, \cdot)$, $\bm{w}_i$ is the prediction similarity vector of $\bm{x}_i$, calculated by $w_{ij} = \text{sim}(\bm{z}_i, \bm{z}_j)$, and $\bm{s}_i$ is the affinity vector of $\bm{x}_i$, formulated as $s_{ij} = \text{aff}(\bm{x}_i, \bm{x}_j)$. 

For the sake of simplicity and effectiveness, we ensure that when constructing $\bm{s}_i$, it satisfies $\bm{s}_i^{\top}\oneb = 1$, and the following specific objective are adopted for optimization:
\vspace{-2mm}
\begin{align}
    \min_{\bm{z}_i,\cdots,\bm{z}_B} & \underset{L_2 \text{ regularization}}{\underbrace{(1 - \lambda)\sum_{i=1}^B \|{\bm z}_i - {\bm p}_i\|^2}}  + \underset{\text{graph regularization}}{\underbrace{\lambda \sum_{i=1}^B \sum_{j=1}^B {s}_{ij}\|{\bm z}_i - {\bm z}_j \|^2 \nonumber}} \\
     &\text{s.t.}~~ {\bm z}_i^{\top}\one_{\mathcal{C}} = 1, i=1, \cdots,B\,.
     \label{eq:lsie}
\end{align}%
In Eq.~\eqref{eq:lsie}, the first term encourages the optimized predictions close to the original ones so that they will not collapse. 
The second term encourages the relationship between the optimized predictions to be consistent with the relationship between the features of the test samples.
A larger value of $\lambda$ implies a stronger degree of graph regularity, which forces similar samples to get consistent predictions. 
We adopt the Lagrange multiplier method to solve this convex optimization problem as
\begin{align}
    \ell({\bm z}_{1:B},{\bm \gamma}) = & (1 - \lambda)\sum_{i=1}^B \|{\bm z}_i - {\bm p}_i\|^2 + \lambda \sum_{i=1}^B \sum_{j=1}^B s_{ij}\|{\bm z}_i - {\bm z}_j \|^2 \nonumber \\
    + & \sum_{i=1}^B \gamma_i({\bm z}_i^{\top}\onec - 1)\,.
\end{align}
Considering ${\bm s}_i^\top \oneb = 1$, we calculate the gradient by
\begin{align}
    \nabla_{z_i} \ell = 2(1-\lambda)({\bm z}_i - {\bm p}_i) + \lambda\sum_{j=1}^B s_{ij}2({\bm z}_i - {\bm z}_j) + \gamma_i \onec \,.
\end{align}
The optimal solution is achieved when the gradient is zero
\begin{align}
    {\bm z}_i = (1-\lambda){\bm p}_i+\lambda\sum_{j=1}^B s_{ij}{\bm z}_j - \frac{\gamma_i}{2} \onec\,.
    \label{eq:solution_with_gamma}
\end{align}
Take the constrain ${\bm z}_i^{\top}\mathbbm{1}_{\mathcal{C}}=1$ (derived from $\frac{\partial \ell}{\partial \gamma_i} = 0$) into consideration, we have
\begin{align}
    {\bm z}_i^\top \onec  =& (1-\lambda){\bm p}_i^\top \onec + \lambda\sum_{j=1}^B s_{ij}{\bm z}_j^{\top}\onec - \frac{\gamma_i}{2} \onec^\top\onec. 
    \label{eq:intermediate}
\end{align}
Because ${\bm p}_i$ is the precondition, it satisfies ${\bm p}_i^\top \onec = 1$. 
Simplifying the Eq.~\eqref{eq:intermediate}, we have $\gamma_i = 0$. 
Subsequently, from Eq.~\eqref{eq:solution_with_gamma}, We get the following formula ${\bm z}_i = \lambda\sum_{j=1}^B s_{i,j}{\bm z}_j + (1 - \lambda){\bm p}_i$.
For the whole test batch, we have ${\bm Z} = \lambda {\bm S}{\bm Z} + (1 - \lambda){\bm P}$, 
where ${\bm S} = [{\bm s}_1,\cdots,{\bm s}_B]^\top\in \R^{B\times B}$ is the affinity matrix.
Since the spectral radius of any matrix are less than any of its norms, we have $\rho ({\bm S}) \le \|{\bm S}\|_\infty = \max_i \sum_j s_{ij} = 1$. Meanwhile, due to $\lambda \in (0,1)$, so $I-\lambda {\bm S}$ is invertible. 
By applying simple algebraic operations, we obtain the optimal solution, which is given by
\begin{align}
    {\bm Z}^* = (1-\lambda)(I - \lambda {\bm S})^{-1}{\bm P}\,.
    \label{eq:optimal_solution}
\end{align}%
With features ${\bm F} = [{\bm f}_1, \cdots, {\bm f}_B] \in \mathbb{R}^{B \times d_f}$ of these samples in batch $\batch$, where $d_f$ represents the feature dimension, we explore two types of affinity matrices: $k$ Nearest Neighbor ($k$NN) and Radial Basis Function (RBF) affinity.
We compute the $k$NN affinity matrix as follows: $s_{ij} = \mathbb{I}\{{\bm f}_j \text{ is }k\text{NN of } {\bm f}_i\} / k$, where $\mathbb{I}\{\cdot\}$ is the indicator function.
For the RBF affinity, we first compute
\begin{align}
    \hat{s}_{ij} = \mathbb{I}\{i\neq j\}\exp\left(-\frac{\| {\bm f}_i - {\bm f}_j \|^2}{2\sigma^2}\right)\,, 
\end{align}%
where $\sigma$ is the variance parameter.
Subsequently, we normalize the matrix by $s_{ij} = \hat{s}_{ij} / \sum_j \hat{s}_{ij}$. 
For the final optimized output, we adopt the one-hot form of Eq.~\eqref{eq:optimal_solution}, which is formulated as ${\bm Z} = \text{onehot}({\bm Z}^*)$.

\subsubsection{Batch-level Bias Reweighting}
Ideally, $\lambda$ should be proportional to the imbalance of the true class distribution $q(y)$, which means the more imbalanced $q(y)$ is, the stronger strength it will explore the latent structure information.
Motivated by this, we propose an imbalanced score measured at the batch level. 
Firstly, the class distribution $\hat{q}(y)$ is estimated by the balanced model prediction $p(y|{\bm x})$, where $\hat{q}(y) = \frac{1}{B}\sum_{i=1}^B\mathbb{I}\{\argmax_j p(j|{\bm x}_i)=y\}$. 
Since the imbalance is only measured within a batch of test data, we focus on the top $\mathcal{R}=\min(\mathcal{C},B)$ largest values of $\hat{q}(y)$ and normalize them by dividing their sum. 
The normalized values are denoted as $\hat{q}(y_i), i=1,\cdots,\mathcal{R}$. Then the imbalance score is formulated as 
\begin{equation}
    \zeta = \left[ \sum_{i=1}^\mathcal{R}\left( \hat{q}(y_i) - \frac{1}{\mathcal{R}} \right)^2 \right]^{\frac{1}{2}} \,.
    \label{eq:zeta}
\end{equation}
One thing to be noticed is that as the number of classes increases, the sensitivity of $\zeta$ decreases. 
To provide a more concrete example, if observes the top $\mathcal{R} = 3$ values as $[0.5, 0.5, 0.0]$ for both 3 classes and 10 classes, Eq~\eqref{eq:zeta} assigns the same score to both scenarios.
However, as we can see, the scenario with 10 classes is more biased than that with 3 classes.
Thereby, a gamma transformation is adopted to recover the sensitivity of the imbalance score, which is given by $\lambda = \zeta^{\frac{1}{\ln \mathcal{C}}}$.
In the end, we combine ${\bm Z}$ and ${\bm P}$ again to ensure the robustness of the final prediction, expressed as 
\begin{equation}
    {\bm Z} = \zeta {\bm Z} + (1 - \zeta) {\bm P}\,.
\end{equation} 
To better understand the proposed \method, the entire process of it is illustrated in Algorithm~\ref{alg:total}.

\begin{algorithm}[t]
    \caption{\method} 
    \label{alg:total}
    \DontPrintSemicolon
    {\bf Input:} The source pre-trained model $f_{\theta_0}$ and the online test stream $\batch_0,\batch_1,\cdots$ of \setting. \\
    {\bf Define:} Teacher model $f_{\ptea}$, student model $f_{\pstu}$, and the memory bank $\bank$. \\

    Initialize $f_{\ptea}, f_{\pstu}$ with $f_{\theta_0}$, and replace all BN layers in $f_{\ptea}, f_{\pstu}$ and $f_{\theta_0}$ with GpreRBN.\\

    \For{$t \in 0,1,\cdots$}{

        Calculate preconditions ${\bm P}$ and features ${\bm F}$ of $\batch_t$ by $f_{\ptea_t}$, with the GpreRBN disabling tracking. \\

        Update $\bank$ by performing CBS with $\batch_t$ and ${\bm P}$.

        Compute the final prediction ${\bm Z}$ by bias-guided output adaptation with ${\bm P}$ and ${\bm F}$. \\

        Sample a random batch $\batch^{\bank}$ from $\bank$. \\

        Calculate $\varL_{total}$ in Eq.~\eqref{eq:total_loss} by $\batch^{\bank}$ with GpreRBN enable tracking and $\batch_t$ with disabling tracking. \\

        Optimize $f_{\pstu_t}\to f_{\pstu_{t+1}}$ with  $\varL_{total}$. \\

        Update $f_{\ptea_t}\to f_{\ptea_{t+1}}$ with Eq.~\eqref{eq:teacher_eam}.
    }
\end{algorithm} 

\section{Experiments}
\label{sec:Experiment}
\subsection{Experiment Setups}
\label{sec:implementation_details}
\textbf{Datasets}.
CIFAR-10-C, CIFAR-100-C, and ImageNet-C\footnote{Different from previous methods~\cite{tent_wang2020, cotta, EATA} that only 5,000 images of each corruption in ImageNet-C are used for evaluation, our test streams are generated with all the 50,000 images of each corruption.}~\cite{corruptions} are commonly used TTA benchmarks employed to evaluate the robustness of image corruptions.
All of them are obtained by applying 15 kinds of corruption with 5 different degrees of severity on their clean test images of original datasets CIFAR10, CIFAR100 and ImageNet respectively. CIFAR10/CIFAR100~\cite{cifar} have 50,000/10,000 training/test images, all of which fall into 10/100 categories, while ImageNet~\cite{deng2009imagenet} has 1,281,167/50,000 training/test images and they are categorized into 1,000 classes.
In addition, we also evaluate \method on three widely used domain adaptation benchmarks including PACS, OfficeHome, and DomainNet.
PACS~\cite{PACS} contains images across four distinct domains: {\it Photo}, {\it Art painting}, {\it Cartoon}, and {\it Sketch}. 
It consists of approximately 7,000 images and all these domains contribute images to the shared 7 classes.
OfficeHome~\cite{Office-Home} comprises images from four different domains: {\it Art}, {\it Clipart}, {\it Product}, and {\it Real}, representing various office and home environments. 
It contains a total of 15,500 images spread across 65 different categories.
DomainNet~\cite{DomainNet} is the largest and hardest dataset to date for domain adaptation and consists of about 0.6 million images with 345 classes. It consists of six different domains including {\it Clipart (clp)}, {\it Infograph (inf)}, {\it Painting (pnt)}, {\it Quickdraw (qdr)}, {\it Real (rel)}, and {\it Sketch (skt)}. 

\paragraphstart{Comparisons Methods}
\method is compared with multiple baselines include Source, BN~\cite{BN_Stat}, PL~\cite{PL}, TENT~\cite{tent_wang2020}, LAME~\cite{niid_boudiaf2022parameter}, CoTTA~\cite{cotta}, NOTE~\cite{note}, ODS~\cite{ODS} and RoTTA~\cite{rotta}. For more details, please refer to the Appendix.
 
\paragraphstart{Implementation Details}.
All of our experiments are conducted with PyTorch~\cite{paszke2019pytorch} framework. 

{\it Source Pre-trained Model:} In the case of robustness to corruption, following the previous methods~\cite{tent_wang2020, EATA, cotta}, we obtain the pre-trained model from RobustBench benchmark~\cite{RobustBench}, including the WildResNet-28~\cite{wildresnet} for CIFAR10 $\to$ CIFAR10-C, the ResNeXt-29~\cite{ResNext} for CIFAR100 $\to$ CIFAR100-C and ResNet-50~\cite{resnet} for ImageNet $\to$ ImageNet-C.
In the case of generalization under the huge domain gap, we train a ResNet-50~\cite{resnet} by standard classification loss for each domain in PACS, OfficeHome and DomainNet. 

{\it Test Stream:} For experiments under corruptions, we change the test corruption at the highest severity 5 one by one, which is {\it motion blur $\to$ snow $\to$ fog $\to$ shot noise $\to$ defocus blur $\to$ contrast $\to$ zoom blur $\to$ brightness $\to$ frost $\to$ elastic transform $\to$ glass blur $\to$ gaussian noise $\to$ pixelate $\to$ jpeg compression $\to$ impulse noise}, to simulate that the data distribution continually changes with time.
For experiments under huge domain gaps, for each source pre-trained model, we adapt it to the rest domains according to alphabetical order.
For example, the model pre-trained on {\it Cartoon} of PACS is adapted to {\it Art painting} $\to$ {\it Photo} $\to$ {\it Sketch} continually.
To obtain the continual label shift, each corruption or domain is separated into several periods. 
For each period, we sample test data with the label distribution generated by Dirichlet Distribution with the concentration parameter $\gamma$ to control the imbalance degree. 
The closer the value of $\gamma$ to 0, the more imbalanced the generated label distribution is and the more dramatically it changes. 

{\it Setup:} For optimization, we adopt Adam~\cite{Adam} optimizer with learning rate $1.0\times 10^{-3}$, $\beta=0.9$. 
For all methods, we set the batch size as 64.
Concerning the hyperparameters, we adopt a unified set of values for \method across all experiments without additional claims, including $\alpha=0.05$, $\nu=0.001$, $\lambda_{batch}=0.01$, $\lambda_{re}=0.1$, $\banksize=1024$ and we adopt the kNN affinity matrix with $k=5$. 

\vspace{-0.5mm}
\begin{table*}[t]
  \centering
  \caption{Classification error of continually adapting the source pre-trained model to test streams under PTTA.}
  \label{table:main_results}
  \VspaceBefore
  \resizebox{\textwidth}{!}{
  \renewcommand{\arraystretch}{0.8}
  {
  \begin{tabular}{l|ccccc|ccccc|ccccc}
      \toprule[1.2pt]

      & \multicolumn{5}{c|}{\bf CIFAR-10-C} & \multicolumn{5}{c|}{\bf CIFAR-100-C} & \multicolumn{5}{c}{\bf ImageNet-C} \\

      \multicolumn{1}{r|}{$\imb$} & $10^{-1}$ & $10^{-2}$ & $10^{-3}$ & $10^{-4}$ & Avg. ($\downarrow$) & $10^{-1}$ & $10^{-2}$ & $10^{-3}$ & $10^{-4}$ & Avg.($\downarrow$) & $10^{-1}$ & $10^{-2}$ & $10^{-3}$ & $10^{-4}$ & Avg.($\downarrow$) \\
      
      \midrule
      Source & 43.6 & 43.9 & 43.6 & 44.2 & 43.8 & 46.0 & 46.7 & 46.7 & 47.1 & 46.6 & 82.1 & 82.2 & 82.4 & 81.5 & 82.1 \\
      BN~\cite{BN} & 52.2 & 70.8 & 74.0 & 74.6 & 67.9 & 47.1 & 76.5 & 90.4 & 93.3 & 76.8 & \ul{72.8} & 78.1 & 91.4 & 97.2 & 84.9 \\
      PL~\cite{PL} & 65.8 & 77.8 & 82.4 & 82.2 & 77.1 & 88.2 & 96.6 & 98.0 & 98.3 & 95.3 & 98.9 & 99.4 & 99.9 & 99.9 & 99.5 \\
      TENT~\cite{tent_wang2020} & 68.3 & 82.9 & 84.4 & 84.1 & 79.9 & 89.6 & 97.2 & 98.6 & 98.5 & 95.9 & 99.1 & 99.6 & 99.9 & 99.9 & 99.6 \\
      LAME~\cite{niid_boudiaf2022parameter} & 41.4 & 40.0 & 39.4 & 40.3 & 40.3 & 42.1 & \ul{35.4} & \ul{33.4} & \ul{33.5} & \ul{36.1} & 82.5 & 79.9 & \ul{76.5} & \ul{74.7} & 78.4 \\
      CoTTA~\cite{cotta} & 59.7 & 79.0 & 79.9 & 80.7 & 74.8 & 47.8 & 79.4 & 93.0 & 95.5 & 78.9 & 87.3 & 94.2 & 99.1 & 99.7 & 95.1 \\
      ODS~\cite{ODS} & 66.7 & 78.8 & 80.7 & 82.4 & 77.1 & 90.6 & 96.8 & 98.1 & 98.3 & 95.9 & 99.5 & 99.6 & 99.8 & 99.9 & 99.7 \\
      NOTE~\cite{note} & 27.3 & 31.1 & 32.2 & 32.8 & 30.8 & 75.1 & 78.7 & 82.5 & 83.9 & 80.0 & 98.8 & 98.8 & 98.9 & 99.0 & 98.9 \\
      \midrule
      RoTTA~\cite{rotta} & \ul{20.0} & \ul{22.1} & \ul{22.9} & \ul{24.1} & \ul{22.3} & \ul{35.1} & 39.2 & 42.9 & 45.0 & 40.6 & \bf 71.0 & \bf 72.3 & 80.5 & 86.2 & \ul{77.5} \\
      \bf \method & 
      \bf 13.1 & \bf 7.0 & \bf 5.3 & \bf 5.5 & \bf 7.7\dtplus{$\downarrow$14.6} & \bf 31.0 & \bf 20.6 & \bf 12.4 & \bf 10.2 & \bf 18.6\dtplus{$\downarrow$17.5} & 78.9 & \ul{74.3} & \bf 61.2 & \bf 53.7 & \bf 67.0\dtplus{$\downarrow$10.5} \\

      \midrule
      \midrule

      & \multicolumn{5}{c|}{\bf PACS} & \multicolumn{5}{c|}{\bf OfficeHome} & \multicolumn{5}{c}{\bf DomainNet} \\

      \multicolumn{1}{r|}{$\imb$} & $10^{-1}$ & $10^{-2}$ & $10^{-3}$ & $10^{-4}$ & Avg. ($\downarrow$) & $10^{-1}$ & $10^{-2}$ & $10^{-3}$ & $10^{-4}$ & Avg.($\downarrow$) & $10^{-1}$ & $10^{-2}$ & $10^{-3}$ & $10^{-4}$ & Avg.($\downarrow$) \\
      
      \midrule
      Source & 45.5 & 47.1 & 48.3 & 45.0 & 46.5 & 47.9 & 48.6 & 45.8 & 48.3 & 47.6 & 77.5 & 77.4 & 77.3 & 77.3 & 77.4 \\
      BN~\cite{BN} & 57.1 & 68.4 & 72.1 & 73.5 & 67.8 & 60.2 & 83.4 & 91.2 & 91.9 & 81.7 & 77.2 & 84.8 & 94.0 & 96.4 & 88.1 \\
      PL~\cite{PL} & 58.0 & 69.8 & 73.2 & 75.4 & 69.1 & 69.8 & 88.8 & 94.1 & 94.5 & 86.8 & 96.6 & 98.4 & 99.3 & 99.4 & 98.4 \\
      TENT~\cite{tent_wang2020} & 59.7 & 70.7 & 73.4 & 75.6 & 69.8 & 75.0 & 91.3 & 95.7 & 95.4 & 89.3 & 97.7 & 98.9 & 99.4 & 99.4 & 98.8 \\
      LAME~\cite{niid_boudiaf2022parameter} & 43.4 & 44.7 & 45.7 & 41.0 & 43.7 & \ul{43.1} & \ul{38.8} & \ul{35.8} & \ul{38.1} & \ul{39.0} & 77.4 & \ul{73.5} & \ul{71.3} & \ul{71.1} & \ul{73.3} \\
      CoTTA~\cite{cotta} & 57.1 & 68.5 & 72.1 & 73.8 & 67.9 & 70.5 & 87.6 & 93.0 & 93.1 & 86.0 & 86.5 & 94.1 & 98.4 & 99.1 & 94.5 \\
      ODS~\cite{ODS} & 53.6 & 63.7 & 67.5 & 70.6 & 63.9 & 79.8 & 90.3 & 94.8 & 94.6 & 89.9 & 98.6 & 98.9 & 99.3 & 99.3 & 99.0 \\
      NOTE~\cite{note} & 30.5 & \ul{29.0} & \ul{31.0} & 34.4 & \ul{31.2} & 62.1 & 64.3 & 64.5 & 63.5 & 63.6 & 94.1 & 94.6 & 94.8 & 94.8 & 94.6 \\
      \midrule
      RoTTA~\cite{rotta} & \ul{29.7} & 30.1 & 33.3 & \ul{32.5} & 31.4 & 47.0 & 49.1 & 46.9 & 49.2 & 48.0 & \ul{74.5} & 76.2 & 77.9 & 78.6 & 76.8 \\
      \bf \method & \bf 25.3 & \bf 23.6 & \bf 28.6 & \bf 23.0 & \bf 25.1\dtplus{$\downarrow$6.1} & \bf 42.8 & \bf 35.0 & \bf 28.6 & \bf 31.5 & \bf 34.5\dtplus{$\downarrow$4.5} & \bf 73.2 & \bf 67.4 & \bf 58.5 & \bf 55.9 & \bf 63.8\dtplus{$\downarrow$9.9} \\

  \bottomrule[1.2pt]
  \end{tabular}
  }
  }
\end{table*}

\begin{table*}[t]
  \centering
  \caption{Detailed classification error when adapting the model to test streams of \setting on {\bf ImageNet-C} with $\bm{\gamma = 10^{-4}}$.}
  \label{table:imagenet}
  \VspaceBefore
  \resizebox{\linewidth}{!}{
  \renewcommand{\arraystretch}{0.8}
  {
  \begin{tabular}{l|ccccccccccccccc|c}
      \toprule[1.2pt]
      Time & \multicolumn{15}{l|}{$t\xrightarrow{\hspace*{18.5cm}}$}& \\ \hline
      Method & \rotatebox[origin=c]{45}{motion} & \rotatebox[origin=c]{45}{snow} & \rotatebox[origin=c]{45}{fog} & \rotatebox[origin=c]{45}{shot} & \rotatebox[origin=c]{45}{defocus} & \rotatebox[origin=c]{45}{contrast} & \rotatebox[origin=c]{45}{zoom} & \rotatebox[origin=c]{45}{brightness} & \rotatebox[origin=c]{45}{frost} & \rotatebox[origin=c]{45}{elastic} & \rotatebox[origin=c]{45}{glass} & \rotatebox[origin=c]{45}{gaussian} & \rotatebox[origin=c]{45}{pixelate} & \rotatebox[origin=c]{45}{jpeg} & \rotatebox[origin=c]{45}{impulse}
      & Avg. ($\downarrow$) \\ 
      
      \midrule
      Source & 85.1 & 83.0 & 79.0 & 92.5 & 84.9 & 96.6 & 75.9 & 43.2 & 80.7 & 84.5 & 91.7 & \ul{94.6} & 75.7 & 63.7 & \ul{91.7} & 81.5 \\

      BN~\cite{BN_Stat} & 97.9 & 96.5 & 95.6 & 98.3 & 98.6 & 98.8 & 96.4 & 94.3 & 96.2 & 96.5 & 98.9 & 98.2 & 96.7 & 96.7 & 98.0 & 97.2 \\

      PL~\cite{PL} & 99.5 & 100.0 & 100.0 & 100.0 & 100.0 & 99.7 & 99.8 & 99.9 & 99.9 & 99.8 & 100.0 & 100.0 & 99.9 & 100.0 & 100.0 & 99.9 \\

      TENT~\cite{tent_wang2020} & 99.5 & 99.9 & 99.8 & 100.0 & 100.0 & 100.0 & 100.0 & 99.9 & 100.0 & 99.7 & 100.0 & 100.0 & 100.0 & 100.0 & 100.0 & 99.9 \\
      
      LAME~\cite{niid_boudiaf2022parameter} & 79.3 & 76.5 & 70.7 & \ul{90.4} & \ul{77.2} & 95.8 & \ul{65.9} & \ul{23.7} & \ul{73.6} & \ul{79.9} & \ul{88.9} & \bf 93.6 & \ul{67.1} & \ul{48.6} & \bf 89.2 & \ul{74.7} \\

      CoTTA~\cite{cotta} & 99.2 & 99.7 & 99.8 & 99.8 & 99.8 & 99.7 & 99.8 & 99.8 & 99.8 & 99.8 & 99.9 & 99.7 & 99.8 & 99.9 & 99.7 & 99.7 \\

      ODS~\cite{ODS} & 99.4 & 100.0 & 99.9 & 100.0 & 99.8 & 99.9 & 100.0 & 99.9 & 100.0 & 99.8 & 100.0 & 100.0 & 100.0 & 100.0 & 99.9 & 99.9 \\
      
      NOTE~\cite{note} & 89.5 & 96.5 & 100.0 & 100.0 & 100.0 & 100.0 & 99.9 & 99.8 & 99.7 & 100.0 & 99.5 & 99.4 & 100.0 & 100.0 & 100.0 & 99.0 \\
      
      \midrule

      RoTTA~\cite{rotta} & \ul{77.6} & \ul{73.2} & \ul{63.6} & 93.8 & 89.0 & \ul{95.3} & 81.0 & 69.5 & 88.6 & 85.7 & 92.4 & 98.0 & 93.8 & 93.0 & 98.3 & 86.2 \\

      \bf \method & \bf 38.0 & \bf 35.4 & \bf 21.4 & \bf 78.2 & \bf 74.9 & \bf 66.4 & \bf 37.6 & \bf 21.0 & \bf 42.0 & \bf 32.4 & \bf 70.7 & 99.0 & \bf 59.3 & \bf 35.2 & 93.8 & \bf 53.7\dtplus{$\downarrow$21.0} \\ 
  
  \bottomrule[1.2pt]
  \end{tabular}
  }
  }
\end{table*}

\begin{table*}[!htbp]
  \caption{Detailed classification error when adapting the model to test streams of \setting on {\bf DomainNet} with $\bm{\gamma = 10^{-4}}$.}
  \VspaceBefore
  \label{table:domainnet}
 \setlength{\tabcolsep}{0.085em}
 \resizebox{\linewidth}{!}{
 \renewcommand{\arraystretch}{0.8}
 {
 \centering
 
 \begin{tabular}{c|c c c c c c c || c | c c c c c c c || c |c c c c c c c}
  \toprule[1.2pt]

  Time & \multicolumn{6}{c}{$t\xrightarrow{\hspace*{3.5cm}}$} & & Time & \multicolumn{6}{c}{$t\xrightarrow{\hspace*{3.5cm}}$} & & Time & \multicolumn{6}{c}{$t\xrightarrow{\hspace*{3.5cm}}$} &  \\

  \hline

  Source & \it{clp} & \it{inf} & \it{pnt} & \it{qdr} & \it{rel} & \it{skt} & Avg. ($\downarrow$) & BN & \it{clp} & \it{inf} & \it{pnt} & \it{qdr} & \it{rel} & \it{skt} & Avg. ($\downarrow$) & TENT & \it{clp} & \it{inf} & \it{pnt} & \it{qdr} & \it{rel} & \it{skt} & Avg. ($\downarrow$) \\
  \midrule
  \it{clp} & N/A & 87.8 & 70.2 & 89.4 & 55.7 & 63.7 & 73.3 & \it{clp} & N/A & 97.2 & 95.6 & 98.1 & 94.3 & 94.3 & 95.9 & \it{clp} & N/A & 99.4 & 99.6 & 99.5 & 99.5 & 99.4 & 99.5 \\
  \it{inf} & 73.3 & N/A & 77.9 & 94.8 & 64.7 & 79.7 & 78.1 & \it{inf} & 97.3 & N/A & 97.5 & 99.3 & 97.0 & 97.5 & 97.7 & \it{inf} & 99.2 & N/A & 99.4 & 99.4 & 99.6 & 98.7 & 99.2 \\
  \it{pnt} & 65.2 & 84.9 & N/A & 96.7 & 55.5 & 69.0 & 74.3 & \it{pnt} & 95.6 & 96.8 & N/A & 99.0 & 94.5 & 95.9 & 96.3 & \it{pnt} & 98.8 & 99.4 & N/A & 99.5 & 99.3 & 99.9 & 99.4 \\
  \it{qdr} & 88.5 & 99.6 & 98.6 & N/A & 95.5 & 92.7 & 95.0 & \it{qdr} & 93.9 & 99.3 & 97.7 & N/A & 96.5 & 96.1 & 96.7 & \it{qdr} & 98.6 & 99.9 & 99.8 & N/A & 100.0 & 99.5 & 99.6 \\
  \it{rel} & 55.3 & 82.8 & 58.7 & 94.0 & N/A & 67.3 & 71.6 & \it{rel} & 94.6 & 96.5 & 94.2 & 98.9 & N/A & 95.1 & 95.9 & \it{rel} & 98.5 & 99.7 & 99.7 & 100.0 & N/A & 98.7 & 99.3 \\
  \it{skt} & 52.3 & 87.6 & 70.4 & 89.4 & 59.3 & N/A & 71.8 & \it{skt} & 94.5 & 96.7 & 95.0 & 98.3 & 94.4 & N/A & 95.8 & \it{skt} & 98.6 & 99.4 & 99.9 & 99.8 & 99.9 & N/A & 99.5 \\
  Avg. ($\downarrow$) & 66.9 & 88.5 & 75.1 & 92.9 & 66.1 & 74.5 & 77.3 & Avg. ($\downarrow$) & 95.2 & 97.3 & 96.0 & 98.7 & 95.3 & 95.8 & 96.4 & Avg. ($\downarrow$) & 98.7 & 99.6 & 99.7 & 99.6 & 99.6 & 99.2 & 99.4 \\

  \hline\hline

 Time & \multicolumn{6}{c}{$t\xrightarrow{\hspace*{3.5cm}}$} & & Time & \multicolumn{6}{c}{$t\xrightarrow{\hspace*{3.5cm}}$} & & Time & \multicolumn{6}{c}{$t\xrightarrow{\hspace*{3.5cm}}$} &  \\

  \hline
 
  LAME & \it{clp} & \it{inf} & \it{pnt} & \it{qdr} & \it{rel} & \it{skt} & Avg. ($\downarrow$) & COTTA & \it{clp} & \it{inf} & \it{pnt} & \it{qdr} & \it{rel} & \it{skt} & Avg. ($\downarrow$) & ODS & \it{clp} & \it{inf} & \it{pnt} & \it{qdr} & \it{rel} & \it{skt} & Avg. ($\downarrow$) \\
  \midrule
  \it{clp} & N/A & 84.9 & 61.8 & 86.6 & 42.3 & 53.3 & 65.8 & \it{clp} & N/A & 98.9 & 99.6 & 99.7 & 99.5 & 99.6 & 99.5 & \it{clp} & N/A & 99.2 & 99.7 & 99.2 & 99.5 & 99.5 & 99.4 \\
  \it{inf} & 65.4 & N/A & 69.5 & 93.4 & 52.4 & 72.3 & 70.6 & \it{inf} & 98.8 & N/A & 99.5 & 99.5 & 99.3 & 99.3 & 99.3 & \it{inf} & 98.9 & N/A & 99.5 & 99.8 & 99.7 & 99.3 & 99.4 \\
  \it{pnt} & 56.9 & 81.0 & N/A & 96.7 & 46.1 & 60.6 & 68.3 & \it{pnt} & 97.8 & 99.5 & N/A & 99.7 & 99.3 & 99.5 & 99.2 & \it{pnt} & 98.3 & 99.3 & N/A & 99.3 & 99.4 & 100.0 & 99.3 \\
  \it{qdr} & 86.2 & 99.6 & 98.7 & N/A & 94.7 & 91.5 & 94.1 & \it{qdr} & 96.2 & 99.6 & 99.1 & N/A & 98.8 & 98.4 & 98.4 & \it{qdr} & 97.0 & 99.9 & 99.5 & N/A & 99.9 & 99.4 & 99.2 \\
  \it{rel} & 44.4 & 77.9 & 46.3 & 92.4 & N/A & 56.9 & 91.5 & \it{rel} & 97.5 & 99.5 & 99.5 & 99.8 & N/A & 99.3 & 99.1 & \it{rel} & 97.8 & 99.6 & 99.4 & 100.0 & N/A & 99.2 & 99.2 \\
  \it{skt} & 40.5 & 85.0 & 62.8 & 86.3 & 47.9 & N/A & 64.5 & \it{skt} & 97.6 & 99.3 & 99.4 & 99.8 & 99.7 & N/A & 99.1 & \it{skt} & 97.8 & 99.5 & 99.9 & 99.9 & 99.8 & N/A & 99.4 \\
  Avg. ($\downarrow$) & 58.7 & 85.7 & 67.8 & 91.1 & 56.7 & 66.9 & \ul{71.1} & Avg. ($\downarrow$) & 97.6 & 99.4 & 99.4 & 99.7 & 99.3 & 99.2 & 99.1 & Avg. ($\downarrow$) & 98.0 & 99.5 & 99.6 & 99.6 & 99.7 & 99.5 & 99.3 \\

  \hline\hline

 Time & \multicolumn{6}{c}{$t\xrightarrow{\hspace*{3.5cm}}$} & & Time & \multicolumn{6}{c}{$t\xrightarrow{\hspace*{3.5cm}}$} & & Time & \multicolumn{6}{c}{$t\xrightarrow{\hspace*{3.5cm}}$} &  \\

  \hline
 
  NOTE & \it{clp} & \it{inf} & \it{pnt} & \it{qdr} & \it{rel} & \it{skt} & Avg. ($\downarrow$) & RoTTA & \it{clp} & \it{inf} & \it{pnt} & \it{qdr} & \it{rel} & \it{skt} & Avg. ($\downarrow$) & \bf \method & \it{clp} & \it{inf} & \it{pnt} & \it{qdr} & \it{rel} & \it{skt} & Avg. ($\downarrow$) \\
  \midrule
  \it{clp} & N/A & 92.8 & 94.8 & 98.8 & 99.5 & 99.8 & 97.1 & \it{clp} & N/A & 90.3 & 71.0 & 87.7 & 63.2 & 69.4 & 76.3 & \it{clp} & N/A & 79.0 & 43.4 & 69.3 & 26.2 & 32.1 & 50.0 \\
  \it{inf} & 89.6 & N/A & 97.0 & 99.7 & 99.3 & 97.8 & 96.7 & \it{inf} & 76.2 & N/A & 79.2 & 95.8 & 74.3 & 84.8 & 82.0 & \it{inf} & 55.3 & N/A & 53.2 & 88.1 & 43.1 & 54.3 & 58.8 \\
  \it{pnt} & 76.9 & 97.2 & N/A & 99.4 & 97.8 & 99.8 & 94.2 & \it{pnt} & 63.8 & 87.3 & N/A & 93.3 & 63.5 & 73.0 & 76.2 & \it{pnt} & 40.0 & 72.7 & N/A & 83.7 & 33.6 & 38.9 & 53.8 \\
  \it{qdr} & 75.2 & 99.5 & 99.9 & N/A & 99.6 & 99.9 & 94.8 & \it{qdr} & 75.7 & 98.6 & 93.4 & N/A & 89.8 & 88.6 & 89.2 & \it{qdr} & 55.4 & 98.1 & 86.2 & N/A & 71.6 & 70.6 & 76.4 \\
  \it{rel} & 73.1 & 97.9 & 98.7 & 99.7 & N/A & 99.0 & 93.7 & \it{rel} & 58.2 & 85.7 & 64.0 & 93.4 & N/A & 74.2 & 75.1 & \it{rel} & 27.9 & 68.9 & 31.6 & 82.5 & N/A & 34.1 & 49.0 \\
  \it{skt} & 69.8 & 95.3 & 97.5 & 99.5 & 99.3 & N/A & 92.3 & \it{skt} & 51.9 & 87.5 & 69.2 & 90.2 & 63.5 & N/A & 72.5 & \it{skt} & 23.0 & 74.0 & 38.5 & 74.3 & 28.8 & N/A & 47.7 \\
  Avg. ($\downarrow$) & 76.9 & 96.5 & 97.6 & 99.4 & 99.1 & 99.3 & 94.8 & Avg. ($\downarrow$) & 65.2 & 89.9 & 75.3 & 92.1 & 70.9 & 78.0 & 78.6 & Avg. ($\downarrow$) & 40.3 & 78.5 & 50.6 & 79.6 & 40.7 & 46.0 & \bf 55.9\dtplus{$\downarrow$15.2} \\
 \bottomrule[1.2pt]
 
 \end{tabular}
 }
 }
 \VspaceAfter
\end{table*}

\subsection{Empirical Observation}
\subsubsection{Analysis on Local Overfitting}
\label{chapter:analysis_bank}
We perform a toy experiment to demonstrate the negative effects of directly adapting to continually changing label distribution $p(y|t)$. 
In this experiment, the test stream originates from the {\it motion blur} corruption with a severity level of 5 on CIFAR-100-C according to a continually changing label distribution.
For comparison, the model is optimized by entropy minimization on the current test batch (denoted as Directly) and the random batch from a memory bank (denoted as Bank) respectively. 
The memory bank is structured as a category-wise  First-In-First-Out (FIFO) design with one sample per class.
Simultaneously, only the affine parameters within the Batch Normalization (BN) layers are updated, and the source statistics are employed for feature normalization.
The results are shown in \Figure~\ref{fig:bank_analysis}.

Firstly, from \Figure~\ref{fig:bank_analysis}a, it's evident that the label distribution is consistently and frequently changing within the test stream when continual label shift occurs. Secondly, the model that is directly adapted to the test stream where exists continual label shift eventually collapses. This is because when the label distribution becomes imbalanced, using unsupervised objective functions like entropy minimization can easily lead the model to bias toward dominant categories, namely local overfitting.
This is beneficial at the current time step, but when the label distribution changes and these dominant categories become minorities, the model's bias will lead to a dramatic drop in performance. 
Further unsupervised updates at this point will lead to catastrophic consequences, ultimately causing the model to collapse, \ie, all test samples are predicted as the same category.
From \Figure~\ref{fig:bank_analysis}b, we can clearly see the process of model collapse, where the model becomes biased towards fewer and fewer categories, eventually converging to only one category.

As the memory bank is constructed as category-balanced, learning with the random batch from it will force the model to learn a balanced label distribution.
As depicted in \Figure~\ref{fig:bank_analysis}c, the model's output remains stable, and no collapse is observed during the online updating.
In addition, \Figure~\ref{fig:bank_analysis}d illustrates the real-time accumulated accuracy of updating the model using two different methods during online adaptation.
We can observe that due to the occurrence of model collapse, the accuracy of direct updates on the test batch keeps decreasing, while using the bank updates can maintain a stable accuracy.

In conclusion, when encountering continual label shift, directly adapting the model to the continually changing $p(y|t)$ suffers severely from local overfitting while forcing the model to learn a balanced label distribution achieves stable adaptation.

\subsubsection{Analysis on GpreRBN}
\label{chapter:analysis_statistic}
\noindent {\bf Statistics Selection.} In this part, we compare four types of normalization statistics including (1) the running mean and running variance of the source domain, (2) statistics of the current batch, (3) statistics obtained by EMA on the current batch and (4) statistics obtained by EMA on the random batch from the category-balanced memory bank. 
The test stream is generated from the {\it motion blur} corruption with a severity level of 5 on CIFAR-10-C according to a continually changing label distribution. 
We only perform statistics recalibration with these statistics, without updating any learnable parameters in the model.
The accumulated accuracy during online inference is shown in \Figure~\ref{fig:statistics_analysis}a.

The statistics from the current batch are notably poorer than those from the source domain. This is mainly due to the imbalanced label distribution introduces significant bias into the statistics. 
In contrast, applying EMA to the test data stream has led to improved performance compared to the source statistics, demonstrating the effectiveness of statistic recalibration in test-time adaptation. 
Nevertheless, due to the continual label shift, some bias remains in the statistics, resulting in suboptimal performance compared to the latest statistics. 
In the end, we achieve a set of robust and reliable statistics through EMA on category-balanced samples, surpassing other methods in performance.

\begin{figure*}[t]
  \centering
  \includegraphics[width=0.99\linewidth]{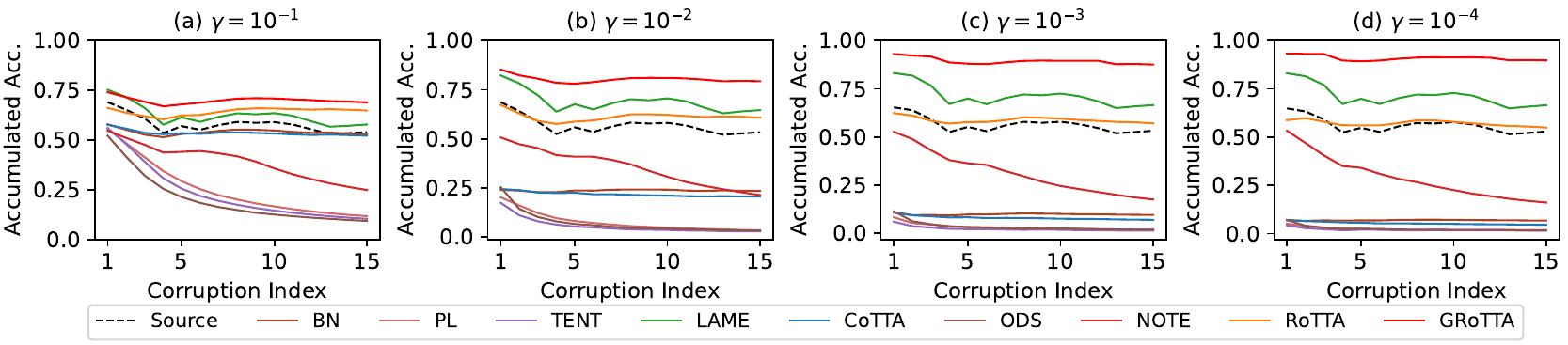}
  \VspaceBefore
  \caption{Accumulated classification accuracy when continually adapting the model to test streams of \setting on CIFAR-100-C.}
  \label{fig:process}
  \VspaceAfter
\end{figure*}

\noindent {\bf Gradient Preserving.}
Following the same test stream and maintaining the same category-balanced memory bank above, we perform entropy minimization on the random batch from the memory bank to update the affine parameters in RBN. 
The accumulated accuracies of w/ and w/o gradient preserving during online adaptation are demonstrated in \Figure~\ref{fig:statistics_analysis}b.
We can see that after adopting this technique, the results have become more stable and improved significantly, verifying our claims.
It's worth noting that in the late stages, both results show a slight decline, which could potentially be attributed to error accumulation with only the entropy minimization objective.

\subsection{Comparisons with the State-of-the-arts}
The overall classification error rates on CIFAR-10-C, CIFAR-100-C, ImageNet-C, PACS, OfficeHome and DomainNet are demonstrated in Table~\ref{table:main_results}.
The test streams are generated as described in Implementation Details with both continual covariate shift and continual label shift.
For each dataset, we evaluate \method across four different values of the label shift parameter, \ie, $\gamma \in \{10^{-1},10^{-2},10^{-3},10^{-4}\}$, representing various potential scenarios that can be encountered in the real world.
The same test stream is shared across all compared methods within each respective scenario.
Meanwhile, the detailed results of the two most difficult tasks, \ie, test streams originate from large-scale datasets ImageNet-C and DomainNet with $\gamma = 10^{-4}$, are shown in Table~\ref{table:imagenet} and~\ref{table:domainnet}.
More detailed results are provided in the Appendix.

\subsubsection{Robustness under Corruptions}
From the top part of Table~\ref{table:main_results}, we can see that \method
achieves the best performance compared to previous methods.
More specifically, \method has a significant performance gain over the second-best method that \gain{14.6\%}, \gain{17.5\%} and \gain{10.5\%} average improvement on CIFAR-10-C, CIFAR-100-C and  ImageNet-C respectively, verifying the effectiveness of \method under \setting.

In more detail, we can observe that BN~\cite{BN_Stat}, PL~\cite{PL}, TENT~\cite{tent_wang2020}, and CoTTA~\cite{cotta} consistently lead to negative adaptation of the model in nearly all scenarios when compared to Source, for example, TENT~\cite{tent_wang2020} has a 36.1\%, 49.3\% and 17.5\% average performance drop on CIFAR-10-C, CIFAR-100-C and ImageNet-C respectively.
This is attributed to the fact that these
methods overlook the issues posed by the continual label shift.
For one aspect, these methods rely on statistics calculated from the current test batch, which are incorrect and biased due to the imbalanced label distribution.
For the other aspect, TENT~\cite{tent_wang2020} and CoTTA~\cite{cotta} directly adapt models to the complex test streams, suffering severely from local overfitting.
Equipped with CBS and GpreRBN, \method is no longer affected by this issue. 
Meanwhile, as continual covariate shift is not taken into account, the performance of NOTE~\cite{note} declines significantly on the harder two datasets CIFAR-100-C and ImageNet-C, which is 33.4\% and 16.8\% lower than Source respectively.
After initially addressing both continual covariate shift and continual label shift simultaneously, our preliminary version RoTTA~\cite{rotta} easily surpasses these single-sided methods.

In addition, we find that LAME~\cite{niid_boudiaf2022parameter} remains a competitive baseline although it never updates the pre-trained model during the entire inference procedure.
However, due to the poor performance of the source model, the improvement that LAME~\cite{niid_boudiaf2022parameter} can bring is limited.
Moreover, ODS~\cite{ODS} tries to perform online category reweighting by estimating the label distribution to address the local overfitting.
However, the continuously and frequently changing label distribution renders the estimation unreliable, resulting in ineffective reweighting.
Meanwhile, neglecting other challenges, it performs poorly in all evaluation scenarios. 
On the contrary, our \method addresses \setting through robust parameter adaptation and further refining the balanced predictions by bias-guided output adaptation, achieving the best performance across \gain{10/12} evaluation scenarios on corruption datasets by a large margin.

\subsubsection{Generalization under Domain Shift}
To further validate the effectiveness of \method, we also evaluate it under huge domain gaps.
For each source domain, the pre-trained model is adapted to the rest domains continually, and the average classification error of all source domains is reported at the bottom of Table~\ref{table:main_results}. 
As we can see, most of the methods, including BN~\cite{BN_Stat}, PL~\cite{PL}, TENT~\cite{tent_wang2020} and CoTTA~\cite{cotta}, perform worse than Source, suggesting that neglecting continual label shift can have disastrous effects once more.
Consistent with the previous analysis, ODS~\cite{ODS} exhibits subpar performance across all datasets, confirming our claims again.
With a comprehensive consideration of both continual covariate shift and continual label shift, \method outperforms the seconde-best baseline by a significant improvement that \gain{6.1\%}, \gain{4.5\%} and \gain{9.9\%} on PACS, OfficeHome and DomainNet respectively, further proving the superiority of \method under \setting.

\begin{table*}[t]
  \begin{minipage}{0.63\textwidth}
      \centering
      \caption{Classification error of different variants of \method on different test streams of CIFAR-100-C.}
      \label{table:ablation}
      \VspaceBefore
      \resizebox{\linewidth}{!}{
      \renewcommand{\arraystretch}{0.8}
      {
      \begin{tabular}{l|cc|cc|cc|cccc|c}
          \toprule[1.2pt]
          & GpreRBN & CBS & $\varL_{sd}$ & $\varL_{re}$ & LSIE & BBR & $10^{-1}$ & $10^{-2}$ & $10^{-3}$ & $10^{-4}$ & Avg. \\
  
          \midrule
  
          BN~\cite{BN_Stat} &   &   &   &   &   &   & 47.1 & 76.5 & 90.4 & 93.3 & 76.8     \\
  
          (a) & \Checkmark & &&&&& 37.7 & 49.5 & 58.4 & 60.7 & 51.6 \\
  
          (b) & \Checkmark & \Checkmark & &&&& 37.0 & 38.1 & 38.9 & 39.3 & 38.3 \\
  
          (c) & \Checkmark & \Checkmark & \Checkmark & &&& 34.4 & 36.5 & 38.4 & 39.0 & 37.1 \\

          (d) & \Checkmark & \Checkmark  & \Checkmark & \Checkmark & && 34.4 & 36.4 & 38.3 & 38.9 & 37.0 \\
  
          (e) & \Checkmark & \Checkmark  & \Checkmark & \Checkmark & \Checkmark & & \bf 29.5 & \bf 19.9 & 16.8 & 16.3 & 20.6 \\
  
          \method & \Checkmark & \Checkmark  & \Checkmark & \Checkmark & \Checkmark & \Checkmark & 31.0 & 20.6 & \bf 12.4 & \bf 10.2 & \bf 18.6 \\
  
          \bottomrule[1.2pt]
      \end{tabular}
      }
      }
      \VspaceAfter
  \end{minipage}%
  \begin{minipage}{0.02\textwidth}
  \end{minipage}
  \begin{minipage}{0.35\textwidth}
      \centering
      \caption{Classification error of different affinity matrixes on CIFAR-100-C.}
      \label{table:diff_aff}
      \VspaceBefore
      \resizebox{\linewidth}{!}{
      \renewcommand{\arraystretch}{0.8}
      {
      \begin{tabular}{lcccccc}
          \toprule[1.2pt]
          & & $10^{-1}$ & $10^{-2}$ & $10^{-3}$ & $10^{-4}$ & Avg. \\
  
          \midrule

          \multirow{4}{*}{$\sigma$} & 0.1 & 32.1 & 26.2 & 25.3 & 25.5 & 27.3 \\
  
          & 1.0 & 33.1 & 28.2 & 15.7 & 11.8 & 22.2 \\
          
          & 2.5 & 33.1 & 28.5 & 15.8 & 11.9 & 22.3 \\
  
          & 5.0 & 33.1 & 28.6 & 15.7 & 11.8 & 22.3 \\
          
          \hline
  
          \multirow{4}{*}{$k$} & 1 & 83.8 & 98.0 & 98.5 & 98.6 & 94.7 \\
  
          & 5 & \bf 31.0 & \bf 20.6 & 12.4 & 10.2 & \bf 18.6 \\
          
          & 10 & 31.2 & 22.1 & \bf 12.1 & \bf 9.2 & \bf 18.6 \\
  
          & 15 & 31.5 & 23.7 & 12.7 & 9.8 & 19.4 \\
  
          \bottomrule[1.2pt]
      \end{tabular}
      }
      }
      \VspaceAfter
  \end{minipage}
\end{table*}

\subsubsection{Adaptation Procedure}
To demonstrate the model performance during the adaptation process, we show the accumulated accuracy of compared methods and \method in Fig.~\ref{fig:process} when continually adapting the model to test streams with $\gamma \in \{10^{-1},10^{-2},10^{-3},10^{-4}\}$ on CIFAR-100-C. 
Firstly, we can observe that those methods with the same batch normalization recalibration as BN~\cite{BN_Stat}, including PL~\cite{PL}, TENT~\cite{tent_wang2020}, CoTTA~\cite{cotta} and ODS~\cite{ODS}, have a poor initial classification accuracy than Source. 
Meanwhile, as the value of $\gamma$ decreases, the initial accuracy also decreases.
This is because, with the label distribution becoming more imbalanced, the statistics calculated from the current batch, which are employed by these methods, become less accurate.
Thanks to GpreRBN, \method is not affected by this issue and maintains a higher initial accuracy across different values of $\gamma$.

Secondly, as we can see, the accuracy of PL~\cite{PL}, TENT~\cite{tent_wang2020}, ODS~\cite{ODS} and NOTE~\cite{note} drops continually when performing adaptation.
Nonetheless, various factors contribute to this phenomenon.
Due to local overfitting, the model may collapse into a single class, leading to an accumulated accuracy curve that resembles an inverse proportional function.
For error accumulation, the model's error rate will gradually increase, leading to a roughly linear decrease in accumulated accuracy.
In other words, PL~\cite{PL}, TENT~\cite{tent_wang2020}, and ODS~\cite{ODS} are more affected by local overfitting, whereas NOTE~\cite{note} suffers more from error accumulation, which further validates our earlier analysis.
Both LAME~\cite{niid_boudiaf2022parameter} and RoTTA~\cite{rotta} exhibit stable adaptive processes, but the performance improvements achieved by both methods are limited. Because LAME~\cite{niid_boudiaf2022parameter} does not consider adaptive model parameters, while RoTTA~\cite{rotta} does not account for the prior information conveyed by the imbalanced label distribution.
After thoroughly considering both continual covariate shift and continual label shift, \method achieves the most stable and best-performing adaptation.

\subsection{Ablation Study}
\textbf{Effect of Each Component.}
We further investigate the efficacy of each component in \method, and the results on CIFAR-100-C are shown in Table~\ref{table:ablation}.

Firstly, we examined the influence of batch normalization recalibration.
As observed, the commonly employed approach of recalculating normalization statistics for each batch of test data exhibits deteriorating performance as the parameter $\gamma$ decreases in magnitude.
When we adopt the approach of maintaining normalization statistics through Exponential Moving Average (EMA) on test streams, denoted as variant (a), the model's performance notably improves. 
However, significant performance degradation still occurs as $\gamma$ decreases.
That is because when the label distribution becomes imbalanced, biases still exist in statistics calculated through EMA directly on the test stream. 
In contrast, variant (b) chooses to compute the normalization statistics with EMA on randomly selected batches from memory bank $\bank$, which further enhances and stabilizes the performance across various values of $\gamma$, proving the effectiveness of the recalibration we adopted.

Secondly, we investigate the losses $\varL_{sd}$ and $\varL_{re}$ to assess their impact.
The results clearly show that both variants (c) and (d) outperform variant (b), demonstrating that updating parameters at test time indeed leads to improved performance.
But, with the limited difficulty of CIFAR-100-C, there is no serious catastrophic forgetting  during adaptation, resulting in only marginal progress from variant (c) to (d).

Finally, we study the effect of output adaptation including {\it Latent Structure Information Exploration} (LSIE) and {\it Batch-level Bias Reweighting} (BBR).
In variant (e), we utilize a fixed value of $\lambda = 0.6$ in Eq.~\eqref{eq:optimal_solution}.
The incorporation of latent structure information to refine predictions has resulted in a substantial reduction in classification errors, indicating that there is indeed information lurking in the data stream.
However, fixed parameters make it inflexible to handle all potential scenarios.
In the end, after integrating the BBR, our \method boosts the performance again, achieving the lowest average classification error.
In short, every component of \method behaves valid to enable effective adaptation under the more challenging \setting.

\paragraphstart{Different Affinity Matrix.}
We have introduced two types of affinity matrices for performing output adaptation, including RBF affinity with parameter $\sigma$ and $k$NN affinity with parameter $k$.
To compare the differences between the two matrices, we employ RBF with $\sigma \in \{0.1, 1.0, 2.5, 5.0\}$ and $k$NN with $k \in \{1, 5, 10, 15\}$ to refine the balanced predictions on CIFAR-100-C, respectively. And the results are demonstrated in Table~\ref{table:diff_aff}.

We observe that, for RBF affinity, the results remain largely consistent when $\sigma \ge 1.0$. This stability arises because, with a large $\sigma$ value, the RBF matrix behaves similarly to a uniform dense matrix, leading to similar results of Eq.~\eqref{eq:optimal_solution}.
Meanwhile, when we set $\sigma = 0.1$, the RBF matrix becomes notably sparse, making it less suitable for leveraging latent structure information. 
A similar phenomenon occurs when adopting $k$NN affinity, where the output nearly collapses when dealing with an excessively sparse matrix, \ie, $k = 1$.
At the same time, when the affinity matrix becomes too dense, i.e., $k = 15$, we notice a slight drop in results.
Conversely, selecting $k$ from the set $\{5, 10\}$ yields an appropriate affinity matrix that leads to the best performance.
In general, if the affinity matrix is either too sparse or too dense, the performance always falls into suboptimal.
For the sake of simplicity, consistency and effectiveness, without additional clarification, we choose the kNN affinity with fixed $k = 5$ for all of our experiments.

\begin{figure*}[t]
  \centering
  \subfloat[]{
      \begin{minipage}[b]{0.24\linewidth} 
        \centering  
        \includegraphics[width=0.99\linewidth]{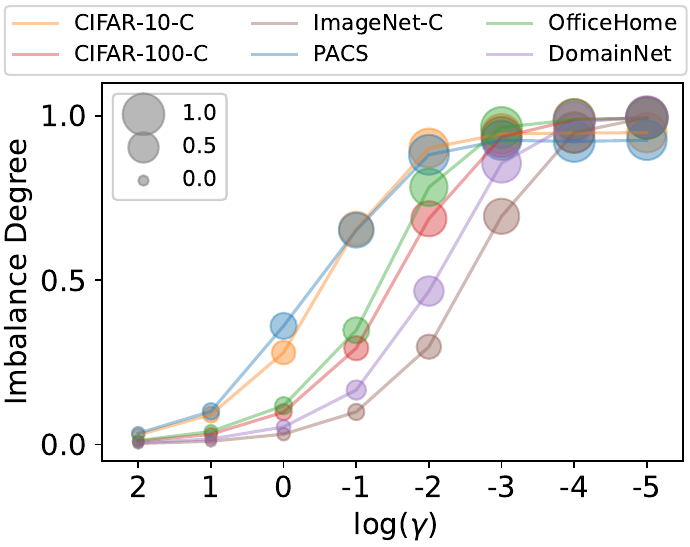}
        \label{fig:gamma_imbalance}
        \VspaceAfter
      \end{minipage}
    }
    \subfloat[]{
      \begin{minipage}[b]{0.24\linewidth} 
        \centering  
        \includegraphics[width=0.99\linewidth]{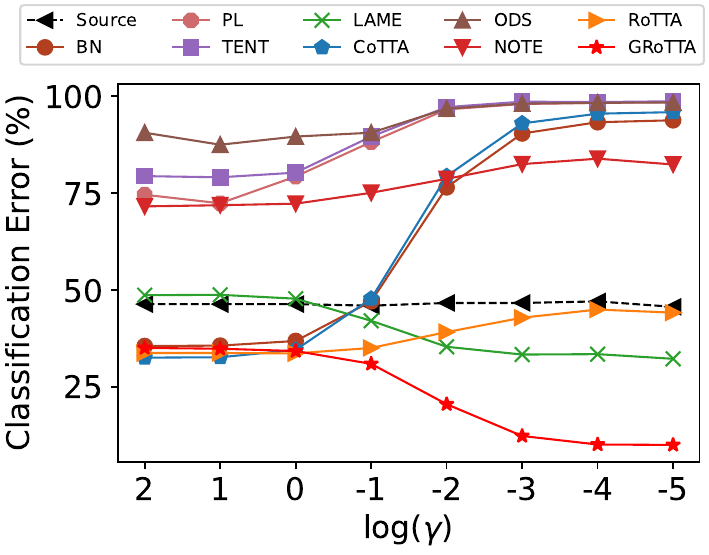}
        \label{fig:different_gamma}
        \VspaceAfter
      \end{minipage}
    }
  \subfloat[]{
    \begin{minipage}[b]{0.24\linewidth} 
      \centering  
      \includegraphics[width=0.99\linewidth]{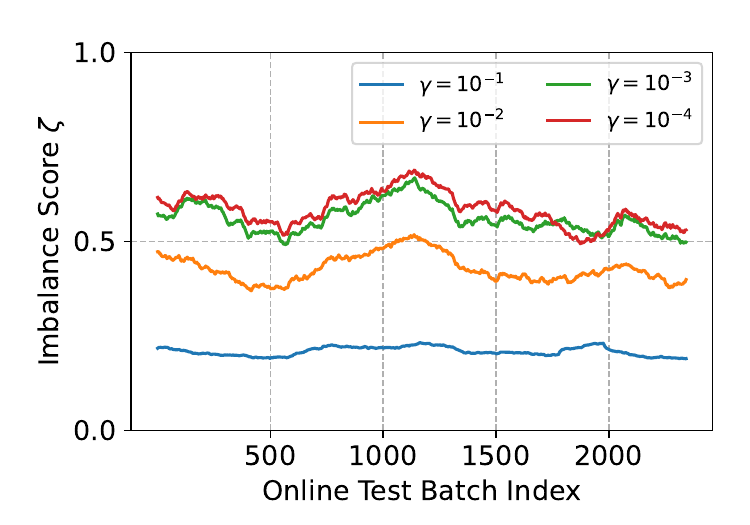}
      \label{fig:visual_zeta}
      \VspaceAfter
    \end{minipage}
  }
  \subfloat[]{
    \begin{minipage}[b]{0.24\linewidth} 
      \centering
      \includegraphics[width=0.99\linewidth]{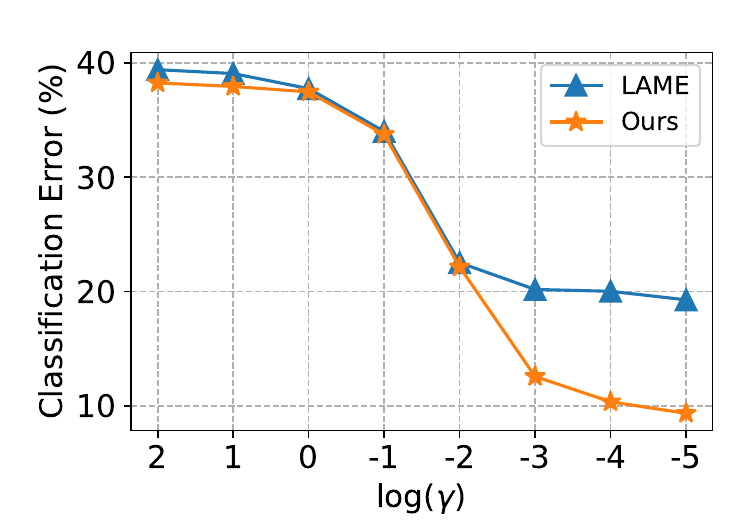}
      \label{fig:compare_oa}
      \VspaceAfter
    \end{minipage}%
  }
  \VspaceBefore
  \caption{(a) the imbalance degree (ID), represented by the y-axis, and change degree (CD), represented by the size of the marker, of the generated label distribution with various values of $\gamma$ on different datasets. A higher ID and CD value indicates the generated label distribution is more imbalanced and changes more dramatically, respectively. (b) We verify the effect of $\gamma$ on different methods on CIFAR-100-C. (c) Visualization of the batch-level imbalance score $\zeta$. (d) We compare our Bias-guided Output Adaptation with LAME~\cite{niid_boudiaf2022parameter} based on our proposed CBS and GpreRBN.}
  \label{fig:gamma_zeta}
  \VspaceAfter
\end{figure*} 

\paragraphstart{Effect of Dirichlet Parameter $\gamma$.}
In the proposed \setting, with the appearance of continual label shift, the label distribution $p(y|t)$ is represented as a sequence of changing distributions $p_0(y), p_1(y), \cdots, p_n(y)$. In our experiments, $\forall i, p_i(y)$ is generated by Dirichlet Distribution with parameter $\gamma$.
As mentioned earlier, as the value of $\gamma$ approaches 0, the generated label distribution becomes increasingly imbalanced and undergoes more dramatic changes.
To validate this assertion, we introduce two metrics to quantify the characteristics of $p_0(y), p_1(y), \cdots, p_n(y)$, including imbalance degree (ID) and change degree (CD), which are calculated by $\text{ID} = \frac{1}{n+1}\sum_{i=0}^n[\sum_{c=1}^{\numclass}(p_i(c) - \frac{1}{\numclass})^2]^{\frac{1}{2}}$ and $\text{CD} = \frac{\sqrt{2}}{2n}\sum_{i=1}^n[\sum_{c=1}^{\numclass}(p_i(c) - p_{i-1}(c))^2]^{\frac{1}{2}}$. 
As we can readily discern, their respective ranges are $\text{ID}\in[0, 1-\frac{1}{\numclass}]$ and $\text{CD}\in[0, 1]$.
A higher ID value indicates a greater degree of label distribution imbalance, while a larger CD value signifies a more significant fluctuation in the label distribution.

Fig.~\ref{fig:gamma_zeta}(a) illustrates the ID and CD values for six datasets as we manipulate $\gamma$.
The first thing we can observe is that as the value of $\gamma$ becomes closer to 0, both ID and CD values tend to increase, confirming our earlier claims on the generated label distribution.
Simultaneously, as $\gamma$ varies, PTTA encompasses a broad spectrum of potential test streams, implying that the generation approach is reasonable.
In addition, it's worth noting that, for a given $\gamma$ value, as the number of categories decreases, the generated label distribution becomes increasingly imbalanced and changes more dramatically, which results from the inherent properties of Dirichlet Distribution.
Finally, to ensure that the test streams cover a wider range of possibilities and the tasks are sufficiently challenging, we adopted $\gamma\in\{10^{-1},10^{-2},10^{-3},10^{-4}\}$ in the main experiment.

Furthermore, we explore a broader range of $\gamma$ values on CIFAR-100-C and conduct comparisons between \method and other methods, as illustrated in Figure~\ref{fig:gamma_zeta}(b).
Consistent with the previous analysis, due to a lack of consideration of continual label shift, the performance of BN~\cite{BN_Stat}, PL~\cite{PL}, TENT~\cite{tent_wang2020} and CoTTA~\cite{cotta} drops quickly as the value of $\gamma$ decreases.
Because of neglecting the continual covariate shift, NOTE~\cite{note} and ODS~\cite{ODS} exhibit poor performance across a wide range of gamma values.
Meanwhile, although RoTTA~\cite{rotta} achieves robust adaptation on dynamic test streams, it doesn't fully leverage the potential information of the label distribution of the test stream. Consequently, it experiences a slight performance drop as $\gamma$ decreases.
Additionally, it's worth noting that LAME~\cite{niid_boudiaf2022parameter} surpasses Source when $\gamma\le 10^{-1}$ but performs worse when $\gamma\ge 10^{0}$.
This suggests that the performance of LAME~\cite{niid_boudiaf2022parameter} is limited to scenarios with an imbalanced label distribution.

Last but not least, when the label distribution approximates a stable uniform distribution, \ie, $\gamma\ge 10^{0}$, our \method can achieve competitive results.
Simultaneously, as the label distribution becomes imbalanced and changes more dramatically, \ie, $\gamma\le 10^{-1}$, the latent structure information in the feature space is more and more apparent, leading to significant performance improvements as $\gamma$ decreases.
The consistently superior performance across scenarios highlights the effectiveness of \method.

\paragraphstart{Visualization of Imbalance Score $\zeta$.} 
We incorporate Batch-level Bias Reweighting into the output adaptation phase of \method to enhance its adaptability to diverse test streams. This involves the introduction of a batch-level imbalance score $\zeta$ that determines the value of $\lambda$ in Eq.~\eqref{eq:optimal_solution}. To assess its effectiveness, in Fig.~\ref{fig:gamma_zeta}(c), we visualize the changes in $\zeta$ during the test-time adaptation on CIFAR-100-C across various $\gamma$ values\footnote{For clarity, the results are processed using mean filtering.}.
As we can see, as $\gamma$ decreases, $\zeta$ increases, indicating that $\zeta$ exactly works as the more imbalanced the label distribution is, the stronger strength \method will explore the latent structure information.
Equipped with it, \method is empowered to effectively handle a wider range of potential dynamic scenarios.

\begin{table}[t]
  \centering
  \caption{Average classification error of 10 different data distribution changing orders.}
  \label{table:different_order}
  \VspaceBefore
  \resizebox{0.85\linewidth}{!}{
  \renewcommand{\arraystretch}{0.8}
  {
  \begin{tabular}{l|cccc|c}
      \toprule[1.2pt]

      \multicolumn{1}{r|}{$\imb$} & $10^{-1}$ & $10^{-2}$ & $10^{-3}$ & $10^{-4}$ & Avg. ($\downarrow$) \\
      
      \midrule
      Source & 46.1 & 46.3 & 46.3 & 46.5 & 46.3 \\

      BN~\cite{BN_Stat} & 47.1 & 76.5 & 90.5 & 93.3 & 76.9 \\

      PL~\cite{PL} & 86.9 & 96.5 & 98.1 & 98.3 & 94.9 \\

      TENT~\cite{tent_wang2020} & 90.2 & 97.3 & 98.4 & 98.5 & 96.1 \\
      
      LAME~\cite{niid_boudiaf2022parameter} & 42.4 & \underline{35.2} & \underline{32.9} & \underline{32.9} & \underline{35.8} \\

      CoTTA~\cite{cotta} & 48.4 & 79.9 & 93.1 & 95.5 & 79.2 \\

      ODS~\cite{ODS} & 76.8 & 80.8 & 83.3 & 83.4 & 81.1 \\
      
      NOTE~\cite{note} & 91.6 & 96.9 & 98.1 & 98.3 & 96.2 \\

      \midrule

      RoTTA~\cite{rotta} & \underline{36.1} & 40.2 & 43.9 & 45.2 & 41.3 \\

      \method & \bf 31.2 & \bf 20.4 & \bf 11.6 & \bf 9.9 & \bf 18.3\dtplus{$\downarrow$17.5} \\ 

      \bottomrule[1.2pt]
  \end{tabular}
  }
  }
  \VspaceAfter
\end{table}

\paragraphstart{Compare to Other Output Adaptation Method.}
In addition, we compare our bias-guided output adaptation with LAME on CIFAR-100-C, and the results are shown in \Figure~\ref{fig:gamma_zeta}(d).
For a fair comparison, we combine the variant (b) of \method in ablation study with LAME~\cite{niid_boudiaf2022parameter} and our bias-guided output adaptation, respectively.
As demonstrated in \Figure~\ref{fig:gamma_zeta}(d), when the task is less difficult, \ie, $\gamma \ge 10^{-2}$, our bias-guided output adaptation only outperform LAME~\cite{niid_boudiaf2022parameter} marginally.
However, with the help of Batch-level Bias Reweighting, our bias-guided output adaptation achieves a significant performance gain than LAME~\cite{niid_boudiaf2022parameter} on harder tasks $\gamma \le 10^{-3}$, verifying the superiority of bias-guided output adaptation.

\paragraphstart{Effect of The Distribution Changing Order.}
To exclude the effect of a fixed changing order of data distribution $p(\bm{x}|t)$, we conducted experiments on ten different sequences of data distribution with four degrees of continual label shift on CIFAR-100-C.
As shown in Table~\ref{table:different_order}, no matter what the value of $\gamma$ is, \method achieves impressive performance improvement.
More specifically, \method obtains a \gain{17.5\%} average performance gain against the second-best baseline.
This demonstrates that \method can robustly and effectively adapt the model in diverse scenarios, establishing it as a reliable option for model deployment.

\begin{figure}[t]
  \centering
  \subfloat[]{
      \begin{minipage}[b]{0.49\linewidth} 
        \centering  
        \includegraphics[width=0.99\linewidth]{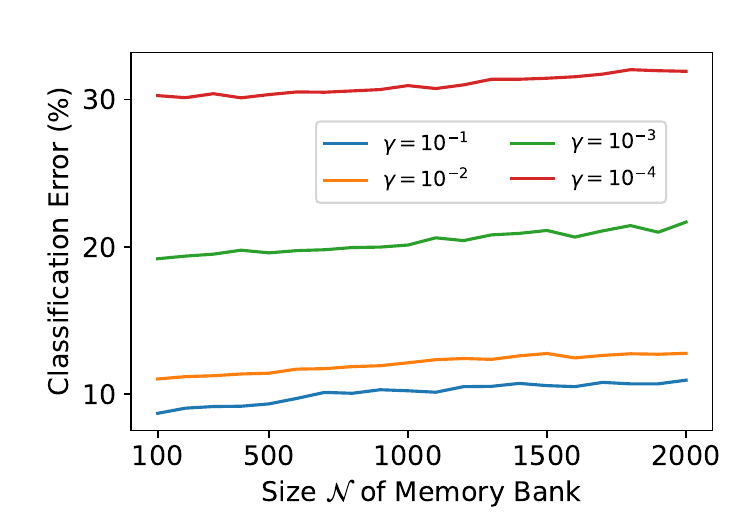}
        \label{fig:bank_size}
        \VspaceAfter
      \end{minipage}
    }
  \subfloat[]{
      \begin{minipage}[b]{0.49\linewidth} 
        \centering  
        \includegraphics[width=0.99\linewidth]{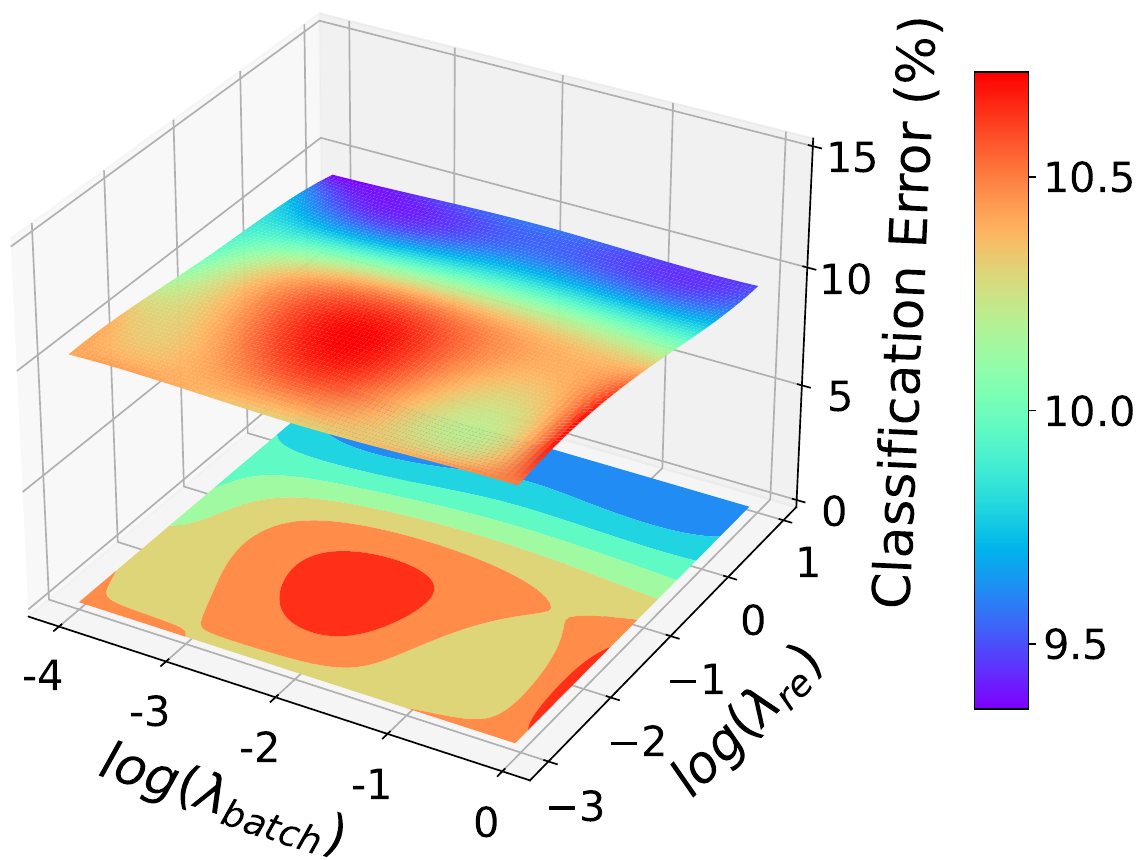}
        \label{fig:sensitivity}
        \VspaceAfter
      \end{minipage}
    }
  \VspaceBefore
  \caption{(a) We verify the effect of the size $\banksize$ of the memory bank on \method on CIFAR-100-C. (b) The sensitivity of \method to parameters $\lambda_{batch}$ and $\lambda_{re}$.}
  \label{fig:size_lambdas}
  \VspaceAfter
\end{figure} 

\paragraphstart{Effect of Memory Bank Size.}
Since we employ a memory bank $\bank$ in \method, its size $\banksize$ plays a crucial role in determining the performance. 
In \Figure~\ref{fig:size_lambdas}(a), we vary $\banksize$ from 100 to 2000 on CIFAR-100-C and demonstrate the classification error of \method.
We can observe that as the $\banksize$ becomes larger and larger, the classification error rate of the model increases slightly.
This is because, as $\banksize$ increases, some older samples begin to be stored in the bank, and they would have negative impacts on current model adaptation.
For consistency, without additional clarification, we simply chose $\banksize=1024$ for all of our experiments.

\paragraphstart{Parameter Sensitivity.}
For the parameters $\lambda_{batch}$ and $\lambda_{re}$, \Figure~\ref{fig:size_lambdas}(b) shows the sensitivity of them on CIFAR-100-C with $\gamma = 10^{-4}$.
While the values of $\lambda_{batch}$ and $\lambda_{re}$ vary across multiple orders of magnitude, only minor performance fluctuations within a small range (less than 2\%) occur, verifying the robustness of \method. For simplicity and consistency, we set $\lambda_{batch}=0.01$ and $\lambda_{re}=0.1$ for all experiments.

\section{Conclusion}
\label{sec:conclusion}
In this paper, we introduce a more realistic Test-Time Adaptation setting called \fullsetting (\setting), where both continual covariate shift and continual label shift occur simultaneously during the test phase. 
To address these challenges, we present the \fullmethod (\method) approach, designed to handle the complexities of dynamic data streams.
More specifically, we begin by updating the model with {\it Robust Parameter Adaptation} for balanced predictions on test samples, which involves eliminating the effects of continual label shift and addressing continual covariate shift.
Since the potential test stream information is overlooked by the balanced predictions, we refine them further through {\it Bias-Guided Output Adaptation}, which improves the performance significantly.
We conduct extensive experiments and perform ablation studies to validate the robustness and effectiveness of our proposed \method. 
We believe that this work opens up new ways for adapting models in real-world applications.

{\small
\bibliographystyle{IEEEtran}
\bibliography{reference}
}

\setcounter{table}{0} \renewcommand{\thetable}{\Alph{section}\arabic{table}}
\setcounter{figure}{0} \renewcommand{\thefigure}{\Alph{section}\arabic{figure}}
\setcounter{section}{0}\renewcommand\thesection{\Alph{section}}

\section{Additional experiment details and results}
\subsection{Comparison Methods}
\paragraphstart{Source} directly employs the source pre-trained model for inference without any adaptation.

\paragraphstart{BN}~\cite{BN_Stat} utilizes statistics of the current batch of data to normalize their feature maps without tuning any parameters.

\paragraphstart{PL}~\cite{PL} is based on BN~\cite{BN_Stat}, and adopts pseudo labels to train the affine parameters in BN layers.

\paragraphstart{TENT}~\cite{tent_wang2020} is the first to propose fully test-time adaptation. It adopts test-time batch normalization and utilizes entropy minimization to train the affine parameters of BN layers. We reimplement it following the released code \url{https://github.com/DequanWang/tent}.

\paragraphstart{LAME}~\cite{niid_boudiaf2022parameter} adapts the output of the pre-trained model by optimizing a group of latent variables without tuning any inner parts of the model. We reimplement it following the released code \url{https://github.com/fiveai/LAME}.

\paragraphstart{CoTTA}~\cite{cotta} considers performing test-time adaptation on continually changing distributions and proposes augmentation-averaged pseudo-labels and stochastic restoration to address error accumulation and catastrophic forgetting. We reimplement it following the released code \url{https://github.com/qinenergy/cotta}.

\paragraphstart{NOTE}~\cite{note} proposes instance-aware normalization and prediction-balanced reservoir sampling to stabilize the adaptation on temporally correlated test streams. We reimplement it following the released code \url{https://github.com/TaesikGong/NOTE}.

\paragraphstart{ODS}~\cite{ODS} considers the most similar scenarios as \setting. It tracks the label distributions and performs online category-wise reweighting to prevent local overfitting. Meanwhile, a similar technique as LAME~\cite{niid_boudiaf2022parameter} is employed to perform output adaptation.

\paragraphstart{RoTTA}~\cite{rotta} is the preliminary version of \method. It is the first to take both continual covariate shift and label temporal correlation into consideration simultaneously. It achieves robust model adaptation on complex test streams but neglects their potential.
Official implementation of RoTTA is released at \url{https://github.com/BIT-DA/RoTTA}.

\subsection{Additional Results}
To demonstrate the effectiveness of our \method, we conducted extensive experiments on 6 datasets, including CIFAR-10-C, CIFAR-100-C, ImageNet-C, PACS, OfficeHome and DomainNet, with different values of $\gamma\in\{10^{-1},10^{-2},10^{-3},10^{-4}\}$, and the results are reported in Table~2 of the main paper.
Then we present the specific results here, including Table~A, B, C and D for CIFAR-10-C, Table~E, F, G and H for CIFAR-100-C, Table~I, J, K and L for ImageNet-C, Table~M, N, O and P for PACS~\footnote{There is a marginal performance gap between PL~\cite{PL} and TENT~\cite{tent_wang2020}, and for better show the results, we chose TENT as the representative of the two. The same selection is adopted for OfficeHome and DomainNet}, Table~Q, R, S and T for OfficeHome and Table~U, V, W and X for DomainNet, respectively.

\begin{table*}[!htbp]
  \vspace{-2mm}
  \centering
  \caption{Detailed classification error when adapting the model to test streams of \setting on {\bf CIFAR-10-C} with $\bm{\gamma = 10^{-1}}$.}
  
  \VspaceBefore
  \resizebox{\linewidth}{!}{
  \renewcommand{\arraystretch}{0.9}
  {
\begin{tabular}{l|ccccccccccccccc|c}
      \toprule[1.2pt]
      Time & \multicolumn{15}{l|}{$t\xrightarrow{\hspace*{18.5cm}}$}& \\ \hline
      Method & \rotatebox[origin=c]{45}{motion} & \rotatebox[origin=c]{45}{snow} & \rotatebox[origin=c]{45}{fog} & \rotatebox[origin=c]{45}{shot} & \rotatebox[origin=c]{45}{defocus} & \rotatebox[origin=c]{45}{contrast} & \rotatebox[origin=c]{45}{zoom} & \rotatebox[origin=c]{45}{brightness} & \rotatebox[origin=c]{45}{frost} & \rotatebox[origin=c]{45}{elastic} & \rotatebox[origin=c]{45}{glass} & \rotatebox[origin=c]{45}{gaussian} & \rotatebox[origin=c]{45}{pixelate} & \rotatebox[origin=c]{45}{jpeg} & \rotatebox[origin=c]{45}{impulse}
      & Avg. (($\downarrow$)) \\ 
      
      \midrule
      Source & 33.5 & 24.5 & 25.4 & 70.1 & 46.9 & 43.3 & 42.3 & 10.8 & 39.9 & 26.9 & 50.1 & 69.9 & 62.9 & 33.1 & 74.5 & 43.6  \\
      BN~\cite{BN} & 48.3 & 51.4 & 47.8 & 55.5 & 48.1 & 43.9 & 48.5 & 42.6 & 51.0 & 54.2 & 62.2 & 54.9 & 54.3 & 58.9 & 61.4 & 52.2  \\
      PL~\cite{PL} & 48.8 & 56.5 & 53.0 & 63.6 & 57.1 & 59.5 & 60.0 & 63.0 & 63.6 & 73.2 & 79.0 & 70.6 & 78.1 & 80.4 & 81.1 & 65.8  \\
      TENT~\cite{tent_wang2020} & 49.3 & 55.0 & 54.1 & 61.8 & 58.1 & 63.3 & 65.5 & 69.7 & 69.6 & 77.1 & 80.4 & 73.7 & 79.7 & 83.3 & 84.4 & 68.3  \\
      LAME~\cite{niid_boudiaf2022parameter} & 30.0 & 19.3 & 21.6 & 71.6 & 44.3 & 42.6 & 39.2 & 7.8 & 37.9 & 21.4 & 47.9 & 71.8 & 63.0 & 27.3 & 75.2 & 41.4  \\
      CoTTA~\cite{cotta} & 49.6 & 58.8 & 57.9 & 63.2 & 58.3 & 57.0 & 59.3 & 56.6 & 61.1 & 61.4 & 66.1 & 59.2 & 61.0 & 63.7 & 63.2 & 59.7  \\
\end{tabular}
  }
  }
\end{table*}

\begin{table*}[!htbp]
  \vspace{-2mm}
  \centering
  \caption{Detailed classification error when adapting the model to test streams of \setting on {\bf CIFAR-10-C} with $\bm{\gamma = 10^{-2}}$.}
  
  \VspaceBefore
  \resizebox{\linewidth}{!}{
  \renewcommand{\arraystretch}{0.9}
  {
  \begin{tabular}{l|ccccccccccccccc|c}
      \toprule[1.2pt]
      Time & \multicolumn{15}{l|}{$t\xrightarrow{\hspace*{18.5cm}}$}& \\ \hline
      Method & \rotatebox[origin=c]{45}{motion} & \rotatebox[origin=c]{45}{snow} & \rotatebox[origin=c]{45}{fog} & \rotatebox[origin=c]{45}{shot} & \rotatebox[origin=c]{45}{defocus} & \rotatebox[origin=c]{45}{contrast} & \rotatebox[origin=c]{45}{zoom} & \rotatebox[origin=c]{45}{brightness} & \rotatebox[origin=c]{45}{frost} & \rotatebox[origin=c]{45}{elastic} & \rotatebox[origin=c]{45}{glass} & \rotatebox[origin=c]{45}{gaussian} & \rotatebox[origin=c]{45}{pixelate} & \rotatebox[origin=c]{45}{jpeg} & \rotatebox[origin=c]{45}{impulse}
      & Avg. (($\downarrow$)) \\ 
      
      \midrule
      Source & 33.9 & 23.7 & 26.0 & 66.4 & 44.3 & 46.6 & 42.2 & 9.3 & 46.0 & 26.9 & 57.4 & 77.8 & 56.4 & 29.2 & 72.2 & 43.9  \\
      BN~\cite{BN} & 68.0 & 70.7 & 67.8 & 72.8 & 68.2 & 65.4 & 67.4 & 65.1 & 70.4 & 73.5 & 78.3 & 72.9 & 71.0 & 73.0 & 77.3 & 70.8  \\
      PL~\cite{PL} & 68.8 & 72.9 & 71.0 & 77.6 & 73.4 & 72.0 & 75.1 & 73.3 & 79.6 & 83.5 & 83.4 & 83.5 & 82.0 & 84.2 & 86.4 & 77.8  \\
      TENT~\cite{tent_wang2020} & 69.1 & 73.1 & 72.0 & 78.9 & 79.8 & 83.4 & 83.7 & 85.3 & 83.7 & 88.1 & 90.7 & 89.1 & 85.7 & 88.2 & 92.2 & 82.9  \\
      LAME~\cite{niid_boudiaf2022parameter} & 29.0 & 17.0 & 21.2 & 66.6 & 39.3 & 43.7 & 37.7 & 5.6 & 42.5 & 18.6 & 54.3 & 78.6 & 55.2 & 21.4 & 70.0 & 40.0  \\
      CoTTA~\cite{cotta} & 69.3 & 76.2 & 78.3 & 79.0 & 79.0 & 78.6 & 77.9 & 78.2 & 78.9 & 80.1 & 84.3 & 82.2 & 79.5 & 79.7 & 83.5 & 79.0  \\
\end{tabular}
  }
  }
\end{table*}

\begin{table*}[!htbp]
  \vspace{-2mm}
  \centering
  \caption{Detailed classification error when adapting the model to test streams of \setting on {\bf CIFAR-10-C} with $\bm{\gamma = 10^{-3}}$.}
  
  \VspaceBefore
  \resizebox{\linewidth}{!}{
  \renewcommand{\arraystretch}{0.9}
  {
  \begin{tabular}{l|ccccccccccccccc|c}
      \toprule[1.2pt]
      Time & \multicolumn{15}{l|}{$t\xrightarrow{\hspace*{18.5cm}}$}& \\ \hline
      Method & \rotatebox[origin=c]{45}{motion} & \rotatebox[origin=c]{45}{snow} & \rotatebox[origin=c]{45}{fog} & \rotatebox[origin=c]{45}{shot} & \rotatebox[origin=c]{45}{defocus} & \rotatebox[origin=c]{45}{contrast} & \rotatebox[origin=c]{45}{zoom} & \rotatebox[origin=c]{45}{brightness} & \rotatebox[origin=c]{45}{frost} & \rotatebox[origin=c]{45}{elastic} & \rotatebox[origin=c]{45}{glass} & \rotatebox[origin=c]{45}{gaussian} & \rotatebox[origin=c]{45}{pixelate} & \rotatebox[origin=c]{45}{jpeg} & \rotatebox[origin=c]{45}{impulse}
      & Avg. (($\downarrow$)) \\ 
      
      \midrule
      Source & 35.1 & 23.7 & 26.7 & 66.4 & 46.3 & 44.2 & 42.7 & 10.7 & 41.3 & 26.9 & 44.1 & 77.5 & 64.4 & 29.3 & 74.2 & 43.6  \\
      BN~\cite{BN} & 71.8 & 73.9 & 71.5 & 76.4 & 71.1 & 71.4 & 72.3 & 70.6 & 71.5 & 75.7 & 78.4 & 75.7 & 74.4 & 76.8 & 78.7 & 74.0  \\
      PL~\cite{PL} & 72.2 & 75.9 & 74.7 & 77.5 & 77.5 & 80.7 & 81.1 & 81.7 & 82.7 & 86.9 & 88.5 & 87.4 & 89.0 & 89.2 & 90.6 & 82.4  \\
      TENT~\cite{tent_wang2020} & 72.4 & 77.8 & 78.3 & 79.8 & 84.1 & 86.2 & 86.7 & 84.5 & 84.2 & 89.0 & 85.5 & 89.7 & 91.3 & 87.1 & 89.3 & 84.4  \\
      LAME~\cite{niid_boudiaf2022parameter} & 30.0 & 16.7 & 21.7 & 66.8 & 41.2 & 41.0 & 37.0 & 6.6 & 37.7 & 18.6 & 39.4 & 78.2 & 63.0 & 20.7 & 72.7 & 39.4  \\
      CoTTA~\cite{cotta} & 72.9 & 78.4 & 78.8 & 80.7 & 78.7 & 80.3 & 78.3 & 79.3 & 79.5 & 80.9 & 83.8 & 80.9 & 80.9 & 80.5 & 84.0 & 79.9  \\
\end{tabular}
  }
  }
\end{table*}

\begin{table*}[!htbp]
  \vspace{-2mm}
  \centering
  \caption{Detailed classification error when adapting the model to test streams of \setting on {\bf CIFAR-10-C} with $\bm{\gamma = 10^{-4}}$.}
  
  \VspaceBefore
  \resizebox{\linewidth}{!}{
  \renewcommand{\arraystretch}{0.9}
  {
  \begin{tabular}{l|ccccccccccccccc|c}
      \toprule[1.2pt]
      Time & \multicolumn{15}{l|}{$t\xrightarrow{\hspace*{18.5cm}}$}& \\ \hline
      Method & \rotatebox[origin=c]{45}{motion} & \rotatebox[origin=c]{45}{snow} & \rotatebox[origin=c]{45}{fog} & \rotatebox[origin=c]{45}{shot} & \rotatebox[origin=c]{45}{defocus} & \rotatebox[origin=c]{45}{contrast} & \rotatebox[origin=c]{45}{zoom} & \rotatebox[origin=c]{45}{brightness} & \rotatebox[origin=c]{45}{frost} & \rotatebox[origin=c]{45}{elastic} & \rotatebox[origin=c]{45}{glass} & \rotatebox[origin=c]{45}{gaussian} & \rotatebox[origin=c]{45}{pixelate} & \rotatebox[origin=c]{45}{jpeg} & \rotatebox[origin=c]{45}{impulse}
      & Avg. (($\downarrow$)) \\ 
      
      \midrule
      Source & 37.7 & 26.2 & 26.2 & 61.7 & 49.2 & 51.4 & 43.1 & 9.7 & 42.5 & 26.1 & 57.2 & 72.9 & 52.9 & 32.6 & 74.5 & 44.2  \\
      BN~\cite{BN} & 73.0 & 74.1 & 71.6 & 75.6 & 73.2 & 71.0 & 74.0 & 70.7 & 72.3 & 75.9 & 78.8 & 76.7 & 75.3 & 77.4 & 79.4 & 74.6  \\
      PL~\cite{PL} & 73.3 & 75.2 & 74.3 & 79.2 & 76.9 & 78.9 & 79.5 & 80.8 & 85.6 & 88.2 & 89.2 & 88.7 & 88.0 & 87.0 & 88.6 & 82.2  \\
      TENT~\cite{tent_wang2020} & 73.4 & 76.1 & 75.2 & 81.7 & 78.2 & 82.4 & 84.1 & 84.6 & 89.7 & 90.4 & 90.9 & 90.4 & 87.9 & 87.0 & 89.1 & 84.1  \\
      LAME~\cite{niid_boudiaf2022parameter} & 32.0 & 20.4 & 21.5 & 61.2 & 45.0 & 48.6 & 37.3 & 5.7 & 38.5 & 18.5 & 53.8 & 72.9 & 51.5 & 25.3 & 72.0 & 40.3  \\
      CoTTA~\cite{cotta} & 73.3 & 79.4 & 79.7 & 82.8 & 82.0 & 80.9 & 82.2 & 78.4 & 80.3 & 82.0 & 83.6 & 80.4 & 79.6 & 82.8 & 83.2 & 80.7  \\
\end{tabular}
  }
  }
\end{table*}

\clearpage

\begin{table*}[!htbp]
  \vspace{-2mm}
  \centering
  \caption{Detailed classification error when adapting the model to test streams of \setting on {\bf CIFAR-100-C} with $\bm{\gamma = 10^{-1}}$.}
  
  \VspaceBefore
  \resizebox{\linewidth}{!}{
  \renewcommand{\arraystretch}{0.9}
  {
  \begin{tabular}{l|ccccccccccccccc|c}
      \toprule[1.2pt]
      Time & \multicolumn{15}{l|}{$t\xrightarrow{\hspace*{18.5cm}}$}& \\ \hline
      Method & \rotatebox[origin=c]{45}{motion} & \rotatebox[origin=c]{45}{snow} & \rotatebox[origin=c]{45}{fog} & \rotatebox[origin=c]{45}{shot} & \rotatebox[origin=c]{45}{defocus} & \rotatebox[origin=c]{45}{contrast} & \rotatebox[origin=c]{45}{zoom} & \rotatebox[origin=c]{45}{brightness} & \rotatebox[origin=c]{45}{frost} & \rotatebox[origin=c]{45}{elastic} & \rotatebox[origin=c]{45}{glass} & \rotatebox[origin=c]{45}{gaussian} & \rotatebox[origin=c]{45}{pixelate} & \rotatebox[origin=c]{45}{jpeg} & \rotatebox[origin=c]{45}{impulse}
      & Avg. (($\downarrow$)) \\ 
      
      \midrule
      Source & 30.9 & 38.6 & 47.9 & 68.5 & 29.1 & 53.9 & 28.3 & 29.6 & 44.9 & 38.0 & 54.1 & 71.5 & 74.2 & 41.4 & 38.7 & 46.0  \\
      BN~\cite{BN} & 42.1 & 47.5 & 52.3 & 52.2 & 40.6 & 43.1 & 39.8 & 40.0 & 45.9 & 48.2 & 52.8 & 51.8 & 44.9 & 52.7 & 53.1 & 47.1  \\
      PL~\cite{PL} & 44.4 & 58.8 & 72.7 & 86.6 & 91.0 & 94.5 & 96.0 & 95.2 & 97.3 & 97.1 & 98.2 & 98.3 & 98.2 & 98.2 & 97.0 & 88.2  \\
      TENT~\cite{tent_wang2020} & 43.9 & 60.0 & 78.7 & 94.0 & 95.3 & 96.4 & 95.7 & 96.1 & 97.2 & 97.1 & 97.6 & 97.8 & 98.3 & 97.9 & 97.7 & 89.6  \\
      LAME~\cite{niid_boudiaf2022parameter} & 24.7 & 31.9 & 44.7 & 67.7 & 23.5 & 52.4 & 23.0 & 24.3 & 40.9 & 31.3 & 50.8 & 71.3 & 75.5 & 36.7 & 33.2 & 42.1  \\
      CoTTA~\cite{cotta} & 42.3 & 46.2 & 50.6 & 49.0 & 44.5 & 48.5 & 42.9 & 44.9 & 48.5 & 49.4 & 51.5 & 50.7 & 45.1 & 50.6 & 52.1 & 47.8  \\
\end{tabular}
  }
  }
\end{table*}

\begin{table*}[!htbp]
  \vspace{-2mm}
  \centering
  \caption{Detailed classification error when adapting the model to test streams of \setting on {\bf CIFAR-100-C} with $\bm{\gamma = 10^{-2}}$.}
  
  \VspaceBefore
  \resizebox{\linewidth}{!}{
  \renewcommand{\arraystretch}{0.9}
  {
  \begin{tabular}{l|ccccccccccccccc|c}
      \toprule[1.2pt]
      Time & \multicolumn{15}{l|}{$t\xrightarrow{\hspace*{18.5cm}}$}& \\ \hline
      Method & \rotatebox[origin=c]{45}{motion} & \rotatebox[origin=c]{45}{snow} & \rotatebox[origin=c]{45}{fog} & \rotatebox[origin=c]{45}{shot} & \rotatebox[origin=c]{45}{defocus} & \rotatebox[origin=c]{45}{contrast} & \rotatebox[origin=c]{45}{zoom} & \rotatebox[origin=c]{45}{brightness} & \rotatebox[origin=c]{45}{frost} & \rotatebox[origin=c]{45}{elastic} & \rotatebox[origin=c]{45}{glass} & \rotatebox[origin=c]{45}{gaussian} & \rotatebox[origin=c]{45}{pixelate} & \rotatebox[origin=c]{45}{jpeg} & \rotatebox[origin=c]{45}{impulse}
      & Avg. (($\downarrow$)) \\ 
      
      \midrule
      Source & 31.3 & 40.5 & 52.2 & 66.3 & 30.3 & 58.3 & 27.0 & 28.0 & 46.0 & 38.7 & 56.5 & 73.8 & 74.0 & 38.4 & 38.8 & 46.7  \\
      BN~\cite{BN} & 76.0 & 76.7 & 78.8 & 77.3 & 72.7 & 76.8 & 73.3 & 74.8 & 76.3 & 76.2 & 80.9 & 78.3 & 73.8 & 78.6 & 77.7 & 76.5  \\
      PL~\cite{PL} & 79.8 & 88.1 & 95.5 & 98.4 & 97.7 & 97.8 & 98.5 & 99.7 & 98.4 & 98.7 & 99.6 & 98.7 & 99.2 & 99.9 & 98.8 & 96.6  \\
      TENT~\cite{tent_wang2020} & 82.6 & 95.3 & 98.1 & 98.7 & 98.9 & 97.0 & 99.2 & 99.5 & 97.4 & 97.6 & 99.6 & 97.7 & 99.6 & 98.7 & 97.6 & 97.2  \\
      LAME~\cite{niid_boudiaf2022parameter} & 17.6 & 25.8 & 39.7 & 61.4 & 17.3 & 48.3 & 13.2 & 15.1 & 34.7 & 20.4 & 45.6 & 70.6 & 71.0 & 23.5 & 26.3 & \ul{35.4} \\
      CoTTA~\cite{cotta} & 75.6 & 76.7 & 79.8 & 78.4 & 77.0 & 81.9 & 78.1 & 80.9 & 80.7 & 80.0 & 82.4 & 80.6 & 77.8 & 80.7 & 80.8 & 79.4  \\
\end{tabular}
  }
  }
\end{table*}

\begin{table*}[!htbp]
  \vspace{-2mm}
  \centering
  \caption{Detailed classification error when adapting the model to test streams of \setting on {\bf CIFAR-100-C} with $\bm{\gamma = 10^{-3}}$.}
  
  \VspaceBefore
  \resizebox{\linewidth}{!}{
  \renewcommand{\arraystretch}{0.9}
  {
  \begin{tabular}{l|ccccccccccccccc|c}
      \toprule[1.2pt]
      Time & \multicolumn{15}{l|}{$t\xrightarrow{\hspace*{18.5cm}}$}& \\ \hline
      Method & \rotatebox[origin=c]{45}{motion} & \rotatebox[origin=c]{45}{snow} & \rotatebox[origin=c]{45}{fog} & \rotatebox[origin=c]{45}{shot} & \rotatebox[origin=c]{45}{defocus} & \rotatebox[origin=c]{45}{contrast} & \rotatebox[origin=c]{45}{zoom} & \rotatebox[origin=c]{45}{brightness} & \rotatebox[origin=c]{45}{frost} & \rotatebox[origin=c]{45}{elastic} & \rotatebox[origin=c]{45}{glass} & \rotatebox[origin=c]{45}{gaussian} & \rotatebox[origin=c]{45}{pixelate} & \rotatebox[origin=c]{45}{jpeg} & \rotatebox[origin=c]{45}{impulse}
      & Avg. (($\downarrow$)) \\ 
      
      \midrule
      Source & 34.6 & 37.6 & 50.1 & 67.3 & 34.2 & 58.2 & 26.0 & 29.0 & 46.7 & 37.1 & 55.9 & 67.0 & 81.0 & 39.4 & 36.9 & 46.7  \\
      BN~\cite{BN} & 89.2 & 92.0 & 90.2 & 91.4 & 88.5 & 89.9 & 88.3 & 88.2 & 90.5 & 91.0 & 91.8 & 90.9 & 92.0 & 91.7 & 91.2 & 90.4  \\
      PL~\cite{PL} & 91.7 & 97.7 & 97.0 & 99.1 & 98.7 & 97.3 & 98.7 & 99.6 & 95.9 & 99.6 & 99.6 & 99.5 & 100.0 & 99.3 & 96.6 & 98.0  \\
      TENT~\cite{tent_wang2020} & 94.0 & 98.6 & 98.5 & 99.7 & 98.5 & 97.7 & 99.4 & 99.3 & 97.0 & 99.5 & 99.8 & 99.6 & 99.6 & 99.8 & 97.9 & 98.6  \\
      LAME~\cite{niid_boudiaf2022parameter} & 16.8 & 19.5 & 33.5 & 61.9 & 18.2 & 48.0 & 10.8 & 14.6 & 31.8 & 19.4 & 42.4 & 60.2 & 77.6 & 22.6 & 24.3 & \ul{33.4}  \\
      CoTTA~\cite{cotta} & 89.0 & 92.3 & 92.2 & 93.5 & 91.3 & 93.9 & 91.6 & 92.2 & 93.9 & 94.1 & 93.8 & 93.5 & 94.8 & 94.5 & 94.1 & 93.0  \\
\end{tabular}
  }
  }
\end{table*}

\begin{table*}[!htbp]
  \vspace{-2mm}
  \centering
  \caption{Detailed classification error when adapting the model to test streams of \setting on {\bf CIFAR-100-C} with $\bm{\gamma = 10^{-4}}$.}
  
  \VspaceBefore
  \resizebox{\linewidth}{!}{
  \renewcommand{\arraystretch}{0.9}
  {
  \begin{tabular}{l|ccccccccccccccc|c}
      \toprule[1.2pt]
      Time & \multicolumn{15}{l|}{$t\xrightarrow{\hspace*{18.5cm}}$}& \\ \hline
      Method & \rotatebox[origin=c]{45}{motion} & \rotatebox[origin=c]{45}{snow} & \rotatebox[origin=c]{45}{fog} & \rotatebox[origin=c]{45}{shot} & \rotatebox[origin=c]{45}{defocus} & \rotatebox[origin=c]{45}{contrast} & \rotatebox[origin=c]{45}{zoom} & \rotatebox[origin=c]{45}{brightness} & \rotatebox[origin=c]{45}{frost} & \rotatebox[origin=c]{45}{elastic} & \rotatebox[origin=c]{45}{glass} & \rotatebox[origin=c]{45}{gaussian} & \rotatebox[origin=c]{45}{pixelate} & \rotatebox[origin=c]{45}{jpeg} & \rotatebox[origin=c]{45}{impulse}
      & Avg. (($\downarrow$)) \\ 
      
      \midrule
      Source & 35.1 & 38.2 & 49.0 & 67.8 & 35.7 & 57.9 & 27.5 & 30.4 & 44.8 & 35.5 & 55.3 & 72.1 & 80.9 & 39.7 & 36.2 & 47.1  \\
      BN~\cite{BN} & 93.1 & 93.8 & 92.7 & 93.7 & 92.7 & 93.4 & 91.4 & 92.1 & 92.7 & 93.7 & 94.3 & 93.6 & 93.8 & 94.4 & 93.9 & 93.3  \\
      PL~\cite{PL} & 94.9 & 97.5 & 98.3 & 99.3 & 97.8 & 98.7 & 99.7 & 99.2 & 97.2 & 99.1 & 98.7 & 97.9 & 99.0 & 99.7 & 97.6 & 98.3  \\
      TENT~\cite{tent_wang2020} & 95.8 & 98.7 & 98.9 & 99.8 & 97.0 & 98.3 & 99.4 & 99.3 & 97.1 & 100.0 & 98.5 & 97.6 & 98.9 & 100.0 & 97.5 & 98.5  \\
      LAME~\cite{niid_boudiaf2022parameter} & 16.9 & 20.0 & 32.2 & 62.4 & 19.0 & 46.5 & 12.0 & 14.6 & 29.8 & 17.8 & 41.5 & 67.0 & 76.7 & 22.4 & 23.3 & \ul{33.5}  \\
      CoTTA~\cite{cotta} & 93.0 & 94.2 & 94.3 & 95.5 & 95.1 & 96.6 & 94.5 & 95.2 & 95.7 & 96.2 & 96.3 & 95.9 & 96.2 & 96.5 & 96.7 & 95.5  \\
\end{tabular}
  }
  }
\end{table*}

\clearpage

\begin{table*}[!htbp]
  \vspace{-2mm}
  \centering
  \caption{Detailed classification error when adapting the model to test streams of \setting on {\bf ImageNet-C} with $\bm{\gamma = 10^{-1}}$.}
  
  \VspaceBefore
  \resizebox{\linewidth}{!}{
  \renewcommand{\arraystretch}{0.9}
  {
  \begin{tabular}{l|ccccccccccccccc|c}
      \toprule[1.2pt]
      Time & \multicolumn{15}{l|}{$t\xrightarrow{\hspace*{18.5cm}}$}& \\ \hline
      Method & \rotatebox[origin=c]{45}{motion} & \rotatebox[origin=c]{45}{snow} & \rotatebox[origin=c]{45}{fog} & \rotatebox[origin=c]{45}{shot} & \rotatebox[origin=c]{45}{defocus} & \rotatebox[origin=c]{45}{contrast} & \rotatebox[origin=c]{45}{zoom} & \rotatebox[origin=c]{45}{brightness} & \rotatebox[origin=c]{45}{frost} & \rotatebox[origin=c]{45}{elastic} & \rotatebox[origin=c]{45}{glass} & \rotatebox[origin=c]{45}{gaussian} & \rotatebox[origin=c]{45}{pixelate} & \rotatebox[origin=c]{45}{jpeg} & \rotatebox[origin=c]{45}{impulse}
      & Avg. (($\downarrow$)) \\ 
      
      \midrule
      Source & 86.8 & 84.5 & 77.6 & 93.3 & 84.9 & 95.7 & 77.2 & 44.5 & 79.6 & 85.7 & 91.6 & 94.5 & 77.0 & 65.2 & 93.7 & 82.1  \\
      BN~\cite{BN} & 79.7 & 69.0 & 56.0 & 86.7 & 88.8 & 89.6 & 65.0 & 37.9 & 70.1 & 59.9 & 88.2 & 88.0 & 58.1 & 67.8 & 87.1 & \ul{72.8}  \\
      PL~\cite{PL} & 86.0 & 99.7 & 99.8 & 99.8 & 99.9 & 99.9 & 99.9 & 99.8 & 99.9 & 99.9 & 99.9 & 99.9 & 99.9 & 99.9 & 99.9 & 98.9  \\
      TENT~\cite{tent_wang2020} & 89.5 & 99.7 & 99.7 & 99.8 & 99.8 & 99.9 & 99.8 & 99.8 & 99.8 & 99.9 & 99.9 & 99.9 & 99.8 & 99.9 & 99.9 & 99.1  \\
      LAME~\cite{niid_boudiaf2022parameter} & 87.3 & 85.7 & 78.8 & 93.6 & 84.9 & 95.8 & 77.5 & 43.5 & 80.0 & 87.3 & 92.1 & 94.7 & 77.2 & 65.0 & 93.9 & 82.5  \\
      CoTTA~\cite{cotta} & 80.5 & 87.0 & 82.6 & 88.8 & 88.1 & 91.8 & 85.6 & 85.1 & 90.9 & 85.7 & 89.0 & 91.2 & 85.3 & 87.6 & 90.8 & 87.3  \\
\end{tabular}
  }
  }
\end{table*}

\begin{table*}[!htbp]
  \vspace{-2mm}
  \centering
  \caption{Detailed classification error when adapting the model to test streams of \setting on {\bf ImageNet-C} with $\bm{\gamma = 10^{-2}}$.}
  
  \VspaceBefore
  \resizebox{\linewidth}{!}{
  \renewcommand{\arraystretch}{0.9}
  {
  \begin{tabular}{l|ccccccccccccccc|c}
      \toprule[1.2pt]
      Time & \multicolumn{15}{l|}{$t\xrightarrow{\hspace*{18.5cm}}$}& \\ \hline
      Method & \rotatebox[origin=c]{45}{motion} & \rotatebox[origin=c]{45}{snow} & \rotatebox[origin=c]{45}{fog} & \rotatebox[origin=c]{45}{shot} & \rotatebox[origin=c]{45}{defocus} & \rotatebox[origin=c]{45}{contrast} & \rotatebox[origin=c]{45}{zoom} & \rotatebox[origin=c]{45}{brightness} & \rotatebox[origin=c]{45}{frost} & \rotatebox[origin=c]{45}{elastic} & \rotatebox[origin=c]{45}{glass} & \rotatebox[origin=c]{45}{gaussian} & \rotatebox[origin=c]{45}{pixelate} & \rotatebox[origin=c]{45}{jpeg} & \rotatebox[origin=c]{45}{impulse}
      & Avg. (($\downarrow$)) \\ 
      
      \midrule
      Source & 87.0 & 84.4 & 77.5 & 93.4 & 84.5 & 95.9 & 77.4 & 44.7 & 79.8 & 85.6 & 92.1 & 94.4 & 76.8 & 65.2 & 93.8 & 82.2  \\
      BN~\cite{BN} & 83.3 & 74.6 & 64.8 & 89.5 & 91.0 & 91.6 & 72.7 & 49.0 & 75.7 & 68.4 & 90.8 & 90.6 & 66.5 & 74.1 & 89.6 & 78.1  \\
      PL~\cite{PL} & 92.6 & 99.7 & 99.9 & 99.9 & 99.9 & 100.0 & 99.8 & 99.9 & 99.8 & 99.9 & 99.9 & 99.9 & 99.9 & 99.9 & 99.9 & 99.4  \\
      TENT~\cite{tent_wang2020} & 95.5 & 99.8 & 99.9 & 99.9 & 99.9 & 99.9 & 99.7 & 99.9 & 99.8 & 99.9 & 99.9 & 99.9 & 100.0 & 99.9 & 100.0 & 99.6  \\
      LAME~\cite{niid_boudiaf2022parameter} & 85.8 & 83.7 & 75.0 & 93.0 & 82.0 & 95.7 & 74.0 & 34.6 & 77.6 & 84.5 & 92.0 & 94.2 & 73.6 & 58.7 & 93.5 & 79.9  \\
      CoTTA~\cite{cotta} & 88.3 & 94.2 & 92.4 & 95.1 & 94.4 & 96.7 & 93.3 & 93.6 & 96.3 & 93.4 & 95.1 & 96.1 & 93.5 & 94.6 & 96.1 & 94.2  \\
\end{tabular}
  }
  }
\end{table*}

\begin{table*}[!htbp]
  \vspace{-2mm}
  \centering
  \caption{Detailed classification error when adapting the model to test streams of \setting on {\bf ImageNet-C} with $\bm{\gamma = 10^{-3}}$.}
  
  \VspaceBefore
  \resizebox{\linewidth}{!}{
  \renewcommand{\arraystretch}{0.9}
  {
  \begin{tabular}{l|ccccccccccccccc|c}
      \toprule[1.2pt]
      Time & \multicolumn{15}{l|}{$t\xrightarrow{\hspace*{18.5cm}}$}& \\ \hline
      Method & \rotatebox[origin=c]{45}{motion} & \rotatebox[origin=c]{45}{snow} & \rotatebox[origin=c]{45}{fog} & \rotatebox[origin=c]{45}{shot} & \rotatebox[origin=c]{45}{defocus} & \rotatebox[origin=c]{45}{contrast} & \rotatebox[origin=c]{45}{zoom} & \rotatebox[origin=c]{45}{brightness} & \rotatebox[origin=c]{45}{frost} & \rotatebox[origin=c]{45}{elastic} & \rotatebox[origin=c]{45}{glass} & \rotatebox[origin=c]{45}{gaussian} & \rotatebox[origin=c]{45}{pixelate} & \rotatebox[origin=c]{45}{jpeg} & \rotatebox[origin=c]{45}{impulse}
      & Avg. (($\downarrow$)) \\ 
      
      \midrule
      Source & 87.0 & 84.1 & 79.4 & 93.7 & 83.7 & 95.5 & 77.6 & 45.7 & 79.9 & 86.5 & 90.7 & 94.3 & 77.6 & 67.6 & 93.3 & 82.4  \\
      BN~\cite{BN} & 93.8 & 89.6 & 86.7 & 95.4 & 96.5 & 96.5 & 89.2 & 79.4 & 90.6 & 88.3 & 96.3 & 96.2 & 87.4 & 89.3 & 95.5 & 91.4  \\
      PL~\cite{PL} & 98.8 & 99.7 & 99.9 & 100.0 & 100.0 & 99.9 & 100.0 & 99.9 & 100.0 & 100.0 & 100.0 & 100.0 & 100.0 & 100.0 & 100.0 & 99.9  \\
      TENT~\cite{tent_wang2020} & 99.3 & 99.8 & 100.0 & 99.9 & 100.0 & 100.0 & 100.0 & 100.0 & 100.0 & 100.0 & 100.0 & 99.9 & 100.0 & 99.8 & 100.0 & 99.9  \\
      LAME~\cite{niid_boudiaf2022parameter} & 82.8 & 79.4 & 72.7 & 92.4 & 76.3 & 94.5 & 69.3 & 26.8 & 72.9 & 83.2 & 87.8 & 93.3 & 69.9 & 54.2 & 91.7 & \ul{76.5}  \\
      CoTTA~\cite{cotta} & 97.4 & 99.0 & 99.0 & 99.2 & 99.1 & 99.5 & 98.9 & 99.1 & 99.4 & 99.2 & 99.3 & 99.6 & 99.0 & 99.1 & 99.5 & 99.1  \\
\end{tabular}
  }
  }
\end{table*}

\begin{table*}[!htbp]
  \vspace{-2mm}
  \centering
  \caption{Detailed classification error when adapting the model to test streams of \setting on {\bf ImageNet-C} with $\bm{\gamma = 10^{-4}}$.}
  
  \VspaceBefore
  \resizebox{\linewidth}{!}{
  \renewcommand{\arraystretch}{0.9}
  {
  \begin{tabular}{l|ccccccccccccccc|c}
      \toprule[1.2pt]
      Time & \multicolumn{15}{l|}{$t\xrightarrow{\hspace*{18.5cm}}$}& \\ \hline
      Method & \rotatebox[origin=c]{45}{motion} & \rotatebox[origin=c]{45}{snow} & \rotatebox[origin=c]{45}{fog} & \rotatebox[origin=c]{45}{shot} & \rotatebox[origin=c]{45}{defocus} & \rotatebox[origin=c]{45}{contrast} & \rotatebox[origin=c]{45}{zoom} & \rotatebox[origin=c]{45}{brightness} & \rotatebox[origin=c]{45}{frost} & \rotatebox[origin=c]{45}{elastic} & \rotatebox[origin=c]{45}{glass} & \rotatebox[origin=c]{45}{gaussian} & \rotatebox[origin=c]{45}{pixelate} & \rotatebox[origin=c]{45}{jpeg} & \rotatebox[origin=c]{45}{impulse}
      & Avg. (($\downarrow$)) \\ 
      
      \midrule
      Source & 85.1 & 83.0 & 79.0 & 92.5 & 84.9 & 96.6 & 75.9 & 43.2 & 80.7 & 84.5 & 91.7 & 94.6 & 75.7 & 63.7 & 91.7 & 81.5  \\
      BN~\cite{BN} & 97.9 & 96.5 & 95.6 & 98.3 & 98.6 & 98.8 & 96.4 & 94.3 & 96.2 & 96.5 & 98.9 & 98.2 & 96.7 & 96.7 & 98.0 & 97.2  \\
      PL~\cite{PL} & 99.5 & 100.0 & 100.0 & 100.0 & 100.0 & 99.7 & 99.8 & 99.9 & 99.9 & 99.8 & 100.0 & 100.0 & 99.9 & 100.0 & 100.0 & 99.9  \\
      TENT~\cite{tent_wang2020} & 99.5 & 99.9 & 99.8 & 100.0 & 100.0 & 100.0 & 100.0 & 99.9 & 100.0 & 99.7 & 100.0 & 100.0 & 100.0 & 100.0 & 100.0 & 99.9  \\
      LAME~\cite{niid_boudiaf2022parameter} & 79.3 & 76.5 & 70.7 & 90.4 & 77.2 & 95.8 & 65.9 & 23.7 & 73.6 & 79.9 & 88.9 & 93.6 & 67.1 & 48.6 & 89.2 & \ul{74.7}  \\
      CoTTA~\cite{cotta} & 99.2 & 99.7 & 99.8 & 99.8 & 99.8 & 99.7 & 99.8 & 99.8 & 99.8 & 99.8 & 99.9 & 99.7 & 99.8 & 99.9 & 99.7 & 99.7  \\
\end{tabular}
  }
  }
\end{table*}

\clearpage

\begin{table*}[!htbp]
  \caption{Detailed classification error when adapting the model to test streams of \setting on {\bf PACS} with $\bm{\gamma = 10^{-1}}$.}
  \VspaceBefore
  
 \setlength{\tabcolsep}{0.085em}
 \resizebox{\linewidth}{!}{
 \renewcommand{\arraystretch}{0.9}
 {
 \centering
 
 \begin{tabular}{c|c c c c c || c | c c c c c || c |c c c c c}
  \toprule[1.2pt]

  Time & \multicolumn{4}{c}{$t\xrightarrow{\hspace*{2.5cm}}$} & & Time & \multicolumn{4}{c}{$t\xrightarrow{\hspace*{2.5cm}}$} & & Time & \multicolumn{4}{c}{$t\xrightarrow{\hspace*{2.5cm}}$} &  \\

  \hline

  Source & A & C & P & S & Avg. ($\downarrow$) & BN & A & C & P & S & Avg. ($\downarrow$) & TENT & A & C & P & S & Avg. ($\downarrow$) \\
  \midrule
  A & N/A & 23.6 & 5.1 & 62.9 & 30.6 & A & N/A & 49.8 & 38.3 & 60.2 & 49.5 & A & N/A & 51.5 & 42.6 & 65.2 & 53.1 \\
  C & 35.7 & N/A & 13.3 & 38.4 & 29.1 & C & 65.2 & N/A & 55.5 & 64.6 & 61.8 & C & 66.8 & N/A & 58.5 & 68.6 & 64.7 \\
  P & 18.1 & 63.2 & N/A & 77.7 & 53.0 & P & 60.6 & 57.0 & N/A & 66.7 & 61.4 & P & 63.2 & 61.9 & N/A & 66.7 & 63.9 \\
  S & 71.3 & 68.3 & 67.7 & N/A & 69.1 & S & 62.1 & 61.1 & 44.4 & N/A & 55.9 & S & 61.9 & 65.5 & 43.4 & N/A & 56.9 \\
  Avg. ($\downarrow$) & 41.7 & 51.7 & 28.7 & 59.6 & 45.5 & Avg. ($\downarrow$) & 62.6 & 56.0 & 46.1 & 63.8 & 57.1 & Avg. ($\downarrow$) & 64.0 & 59.6 & 48.2 & 66.8 & 59.7 \\

  \hline\hline

  Time & \multicolumn{4}{c}{$t\xrightarrow{\hspace*{2.5cm}}$} & & Time & \multicolumn{4}{c}{$t\xrightarrow{\hspace*{2.5cm}}$} & & Time & \multicolumn{4}{c}{$t\xrightarrow{\hspace*{2.5cm}}$} &  \\
\end{tabular}
 }
 }
 \VspaceAfter
\end{table*}

\begin{table*}[!htbp]
  \caption{Detailed classification error when adapting the model to test streams of \setting on {\bf PACS} with $\bm{\gamma = 10^{-2}}$.}
  \VspaceBefore
  
 \setlength{\tabcolsep}{0.085em}
 \resizebox{\linewidth}{!}{
 \renewcommand{\arraystretch}{0.9}
 {
 \centering
 
 \begin{tabular}{c|c c c c c || c | c c c c c || c |c c c c c}
  \toprule[1.2pt]

  Time & \multicolumn{4}{c}{$t\xrightarrow{\hspace*{2.5cm}}$} & & Time & \multicolumn{4}{c}{$t\xrightarrow{\hspace*{2.5cm}}$} & & Time & \multicolumn{4}{c}{$t\xrightarrow{\hspace*{2.5cm}}$} &  \\

  \hline

  Source & A & C & P & S & Avg. ($\downarrow$) & BN & A & C & P & S & Avg. ($\downarrow$) & TENT & A & C & P & S & Avg. ($\downarrow$) \\
  \midrule
  A & N/A & 25.9 & 4.5 & 71.5 & 34.0 & A & N/A & 65.0 & 62.0 & 74.5 & 67.2 & A & N/A & 66.5 & 64.0 & 80.2 & 70.2 \\
  C & 35.8 & N/A & 4.8 & 38.0 & 26.2 & C & 68.2 & N/A & 65.2 & 72.0 & 68.4 & C & 68.5 & N/A & 68.0 & 77.8 & 71.4 \\
  P & 20.9 & 57.4 & N/A & 92.6 & 56.9 & P & 71.2 & 63.2 & N/A & 78.2 & 70.9 & P & 73.0 & 64.2 & N/A & 81.3 & 72.8 \\
  S & 72.5 & 66.8 & 74.2 & N/A & 71.2 & S & 65.0 & 69.6 & 66.2 & N/A & 67.0 & S & 65.3 & 75.3 & 64.1 & N/A & 68.3 \\
  Avg. ($\downarrow$) & 43.1 & 50.0 & 27.8 & 67.4 & 47.1 & Avg. ($\downarrow$) & 68.1 & 66.0 & 64.5 & 74.9 & 68.4 & Avg. ($\downarrow$) & 68.9 & 68.7 & 65.3 & 79.8 & 70.7 \\

  \hline\hline

  Time & \multicolumn{4}{c}{$t\xrightarrow{\hspace*{2.5cm}}$} & & Time & \multicolumn{4}{c}{$t\xrightarrow{\hspace*{2.5cm}}$} & & Time & \multicolumn{4}{c}{$t\xrightarrow{\hspace*{2.5cm}}$} &  \\
\end{tabular}
 }
 }
 \VspaceAfter
\end{table*}

\begin{table*}[!htbp]
  \caption{Detailed classification error when adapting the model to test streams of \setting on {\bf PACS} with $\bm{\gamma = 10^{-3}}$.}
  \VspaceBefore
  
 \setlength{\tabcolsep}{0.085em}
 \resizebox{\linewidth}{!}{
 \renewcommand{\arraystretch}{0.9}
 {
 \centering
 
 \begin{tabular}{c|c c c c c || c | c c c c c || c |c c c c c}
  \toprule[1.2pt]

  Time & \multicolumn{4}{c}{$t\xrightarrow{\hspace*{2.5cm}}$} & & Time & \multicolumn{4}{c}{$t\xrightarrow{\hspace*{2.5cm}}$} & & Time & \multicolumn{4}{c}{$t\xrightarrow{\hspace*{2.5cm}}$} &  \\

  \hline

  Source & A & C & P & S & Avg. ($\downarrow$) & BN & A & C & P & S & Avg. ($\downarrow$) & TENT & A & C & P & S & Avg. ($\downarrow$) \\
  \midrule
  A & N/A & 31.6 & 4.3 & 73.8 & 36.6 & A & N/A & 72.0 & 64.9 & 72.2 & 69.7 & A & N/A & 72.3 & 66.5 & 67.5 & 68.8 \\
  C & 25.0 & N/A & 18.4 & 34.0 & 25.8 & C & 70.8 & N/A & 68.2 & 72.2 & 70.4 & C & 71.8 & N/A & 70.4 & 78.4 & 73.5 \\
  P & 16.3 & 76.8 & N/A & 86.8 & 60.0 & P & 72.5 & 73.9 & N/A & 78.2 & 74.9 & P & 74.3 & 76.1 & N/A & 79.2 & 76.5 \\
  S & 76.2 & 68.9 & 67.9 & N/A & 71.0 & S & 73.8 & 75.2 & 71.8 & N/A & 73.6 & S & 73.5 & 77.2 & 73.7 & N/A & 74.8 \\
  Avg. ($\downarrow$) & 39.2 & 59.1 & 30.2 & 64.9 & 48.3 & Avg. ($\downarrow$) & 72.4 & 73.7 & 68.3 & 74.2 & 72.1 & Avg. ($\downarrow$) & 73.2 & 75.2 & 70.2 & 75.0 & 73.4 \\

  \hline\hline

  Time & \multicolumn{4}{c}{$t\xrightarrow{\hspace*{2.5cm}}$} & & Time & \multicolumn{4}{c}{$t\xrightarrow{\hspace*{2.5cm}}$} & & Time & \multicolumn{4}{c}{$t\xrightarrow{\hspace*{2.5cm}}$} &  \\
\end{tabular}
 }
 }
 \VspaceAfter
\end{table*}

\begin{table*}[!htbp]
  \caption{Detailed classification error when adapting the model to test streams of \setting on {\bf PACS} with $\bm{\gamma = 10^{-4}}$.}
  \VspaceBefore
  
 \setlength{\tabcolsep}{0.085em}
 \resizebox{\linewidth}{!}{
 \renewcommand{\arraystretch}{0.9}
 {
 \centering
 
 \begin{tabular}{c|c c c c c || c | c c c c c || c |c c c c c}
  \toprule[1.2pt]

  Time & \multicolumn{4}{c}{$t\xrightarrow{\hspace*{2.5cm}}$} & & Time & \multicolumn{4}{c}{$t\xrightarrow{\hspace*{2.5cm}}$} & & Time & \multicolumn{4}{c}{$t\xrightarrow{\hspace*{2.5cm}}$} &  \\

  \hline

  Source & A & C & P & S & Avg. ($\downarrow$) & BN & A & C & P & S & Avg. ($\downarrow$) & TENT & A & C & P & S & Avg. ($\downarrow$) \\
  \midrule
  A & N/A & 30.6 & 5.2 & 44.8 & 26.8 & A & N/A & 69.8 & 67.8 & 75.7 & 71.1 & A & N/A & 70.3 & 71.3 & 76.8 & 72.8 \\
  C & 34.1 & N/A & 13.2 & 46.9 & 31.4 & C & 81.6 & N/A & 66.7 & 77.2 & 75.2 & C & 83.2 & N/A & 68.5 & 81.0 & 77.6 \\
  P & 23.6 & 43.9 & N/A & 83.9 & 50.5 & P & 73.1 & 68.7 & N/A & 80.4 & 74.0 & P & 75.0 & 67.8 & N/A & 84.9 & 75.9 \\
  S & 67.5 & 81.5 & 64.8 & N/A & 71.3 & S & 73.6 & 81.0 & 66.5 & N/A & 73.7 & S & 74.5 & 82.3 & 71.7 & N/A & 76.2 \\
  Avg. ($\downarrow$) & 41.7 & 52.0 & 27.7 & 58.5 & 45.0 & Avg. ($\downarrow$) & 76.1 & 73.1 & 67.0 & 77.8 & 73.5 & Avg. ($\downarrow$) & 77.6 & 73.5 & 70.5 & 80.9 & 75.6 \\

  \hline\hline

  Time & \multicolumn{4}{c}{$t\xrightarrow{\hspace*{2.5cm}}$} & & Time & \multicolumn{4}{c}{$t\xrightarrow{\hspace*{2.5cm}}$} & & Time & \multicolumn{4}{c}{$t\xrightarrow{\hspace*{2.5cm}}$} &  \\
\end{tabular}
 }
 }
 \VspaceAfter
\end{table*}

\begin{table*}[!htbp]
  \caption{Detailed classification error when adapting the model to test streams of \setting on {\bf OfficeHome} with $\bm{\gamma = 10^{-1}}$.}
  \VspaceBefore
  
 \setlength{\tabcolsep}{0.085em}
 \resizebox{\linewidth}{!}{
 \renewcommand{\arraystretch}{0.9}
 {
 \centering
 
 \begin{tabular}{c|c c c c c || c | c c c c c || c |c c c c c}
  \toprule[1.2pt]

  Time & \multicolumn{4}{c}{$t\xrightarrow{\hspace*{2.5cm}}$} & & Time & \multicolumn{4}{c}{$t\xrightarrow{\hspace*{2.5cm}}$} & & Time & \multicolumn{4}{c}{$t\xrightarrow{\hspace*{2.5cm}}$} &  \\

  \hline

  Source & A & C & P & R & Avg. ($\downarrow$) & BN & A & C & P & R & Avg. ($\downarrow$) & TENT & A & C & P & R & Avg. ($\downarrow$) \\
  \midrule
  A & N/A & 57.7 & 45.1 & 36.9 & 46.6 & A & N/A & 66.4 & 62.1 & 52.9 & 60.4 & A & N/A & 70.5 & 84.1 & 87.8 & 80.8 \\
  C & 58.0 & N/A & 44.9 & 45.4 & 49.4 & C & 68.5 & N/A & 61.4 & 58.2 & 62.7 & C & 71.9 & N/A & 78.5 & 86.4 & 78.9 \\
  P & 59.7 & 62.7 & N/A & 37.6 & 53.4 & P & 68.3 & 69.5 & N/A & 53.7 & 63.8 & P & 70.9 & 79.0 & N/A & 75.3 & 75.1 \\
  R & 43.0 & 54.3 & 28.9 & N/A & 42.1 & R & 56.8 & 61.8 & 42.7 & N/A & 53.8 & R & 58.7 & 68.8 & 68.5 & N/A & 65.3 \\
  Avg. ($\downarrow$) & 53.6 & 58.2 & 39.7 & 40.0 & 47.9 & Avg. ($\downarrow$) & 64.5 & 65.9 & 55.4 & 54.9 & 60.2 & Avg. ($\downarrow$) & 67.2 & 72.8 & 77.0 & 83.2 & 75.0 \\

  \hline\hline

  Time & \multicolumn{4}{c}{$t\xrightarrow{\hspace*{2.5cm}}$} & & Time & \multicolumn{4}{c}{$t\xrightarrow{\hspace*{2.5cm}}$} & & Time & \multicolumn{4}{c}{$t\xrightarrow{\hspace*{2.5cm}}$} &  \\
\end{tabular}
 }
 }
 \VspaceAfter
\end{table*}

\begin{table*}[!htbp]
  \caption{Detailed classification error when adapting the model to test streams of \setting on {\bf OfficeHome} with $\bm{\gamma = 10^{-2}}$.}
  \VspaceBefore
  
 \setlength{\tabcolsep}{0.085em}
 \resizebox{\linewidth}{!}{
 \renewcommand{\arraystretch}{0.9}
 {
 \centering
 
 \begin{tabular}{c|c c c c c || c | c c c c c || c |c c c c c}
  \toprule[1.2pt]

  Time & \multicolumn{4}{c}{$t\xrightarrow{\hspace*{2.5cm}}$} & & Time & \multicolumn{4}{c}{$t\xrightarrow{\hspace*{2.5cm}}$} & & Time & \multicolumn{4}{c}{$t\xrightarrow{\hspace*{2.5cm}}$} &  \\

  \hline

  Source & A & C & P & R & Avg. ($\downarrow$) & BN & A & C & P & R & Avg. ($\downarrow$) & TENT & A & C & P & R & Avg. ($\downarrow$) \\
  \midrule
  A & N/A & 64.7 & 43.0 & 43.0 & 50.2 & A & N/A & 87.2 & 84.8 & 83.4 & 85.2 & A & N/A & 89.5 & 92.0 & 94.6 & 92.0 \\
  C & 53.3 & N/A & 42.9 & 43.4 & 46.5 & C & 84.2 & N/A & 83.5 & 82.7 & 83.5 & C & 88.4 & N/A & 91.7 & 96.0 & 92.0 \\
  P & 59.7 & 66.4 & N/A & 35.4 & 53.8 & P & 84.2 & 88.8 & N/A & 79.5 & 84.2 & P & 86.8 & 94.7 & N/A & 95.3 & 92.3 \\
  R & 42.5 & 59.7 & 29.7 & N/A & 43.9 & R & 79.7 & 84.9 & 78.0 & N/A & 80.9 & R & 82.7 & 91.4 & 92.0 & N/A & 88.7 \\
  Avg. ($\downarrow$) & 51.8 & 63.6 & 38.5 & 40.6 & 48.6 & Avg. ($\downarrow$) & 82.7 & 87.0 & 82.1 & 81.9 & 83.4 & Avg. ($\downarrow$) & 86.0 & 91.9 & 91.9 & 95.3 & 91.3 \\

  \hline\hline

  Time & \multicolumn{4}{c}{$t\xrightarrow{\hspace*{2.5cm}}$} & & Time & \multicolumn{4}{c}{$t\xrightarrow{\hspace*{2.5cm}}$} & & Time & \multicolumn{4}{c}{$t\xrightarrow{\hspace*{2.5cm}}$} &  \\
\end{tabular}
 }
 }
 \VspaceAfter
\end{table*}

\begin{table*}[!htbp]
  \caption{Detailed classification error when adapting the model to test streams of \setting on {\bf OfficeHome} with $\bm{\gamma = 10^{-3}}$.}
  \VspaceBefore
  
 \setlength{\tabcolsep}{0.085em}
 \resizebox{\linewidth}{!}{
 \renewcommand{\arraystretch}{0.9}
 {
 \centering
 
 \begin{tabular}{c|c c c c c || c | c c c c c || c |c c c c c}
  \toprule[1.2pt]

  Time & \multicolumn{4}{c}{$t\xrightarrow{\hspace*{2.5cm}}$} & & Time & \multicolumn{4}{c}{$t\xrightarrow{\hspace*{2.5cm}}$} & & Time & \multicolumn{4}{c}{$t\xrightarrow{\hspace*{2.5cm}}$} &  \\

  \hline

  Source & A & C & P & R & Avg. ($\downarrow$) & BN & A & C & P & R & Avg. ($\downarrow$) & TENT & A & C & P & R & Avg. ($\downarrow$) \\
  \midrule
  A & N/A & 52.8 & 58.1 & 33.0 & 48.0 & A & N/A & 92.4 & 94.3 & 89.6 & 92.1 & A & N/A & 94.0 & 98.5 & 95.8 & 96.1 \\
  C & 51.5 & N/A & 51.9 & 40.5 & 48.0 & C & 91.9 & N/A & 93.2 & 90.4 & 91.8 & C & 94.4 & N/A & 98.5 & 98.4 & 97.1 \\
  P & 57.4 & 54.7 & N/A & 40.2 & 50.8 & P & 92.5 & 91.4 & N/A & 87.5 & 90.5 & P & 94.2 & 94.3 & N/A & 97.8 & 95.4 \\
  R & 41.4 & 42.8 & 25.1 & N/A & 36.4 & R & 91.6 & 91.7 & 87.4 & N/A & 90.2 & R & 93.5 & 93.9 & 95.7 & N/A & 94.4 \\
  Avg. ($\downarrow$) & 50.1 & 50.1 & 45.0 & 37.9 & 45.8 & Avg. ($\downarrow$) & 92.0 & 91.8 & 91.6 & 89.2 & 91.2 & Avg. ($\downarrow$) & 94.0 & 94.1 & 97.5 & 97.3 & 95.7 \\

  \hline\hline

  Time & \multicolumn{4}{c}{$t\xrightarrow{\hspace*{2.5cm}}$} & & Time & \multicolumn{4}{c}{$t\xrightarrow{\hspace*{2.5cm}}$} & & Time & \multicolumn{4}{c}{$t\xrightarrow{\hspace*{2.5cm}}$} &  \\
\end{tabular}
 }
 }
 \VspaceAfter
\end{table*}

\begin{table*}[!htbp]
  \caption{Detailed classification error when adapting the model to test streams of \setting on {\bf OfficeHome} with $\bm{\gamma = 10^{-4}}$.}
  \VspaceBefore
  
 \setlength{\tabcolsep}{0.085em}
 \resizebox{\linewidth}{!}{
 \renewcommand{\arraystretch}{0.9}
 {
 \centering
 
 \begin{tabular}{c|c c c c c || c | c c c c c || c |c c c c c}
  \toprule[1.2pt]

  Time & \multicolumn{4}{c}{$t\xrightarrow{\hspace*{2.5cm}}$} & & Time & \multicolumn{4}{c}{$t\xrightarrow{\hspace*{2.5cm}}$} & & Time & \multicolumn{4}{c}{$t\xrightarrow{\hspace*{2.5cm}}$} &  \\

  \hline

  Source & A & C & P & R & Avg. ($\downarrow$) & BN & A & C & P & R & Avg. ($\downarrow$) & TENT & A & C & P & R & Avg. ($\downarrow$) \\
  \midrule
  A & N/A & 56.9 & 35.4 & 32.6 & 41.6 & A & N/A & 94.0 & 91.4 & 90.5 & 92.0 & A & N/A & 95.2 & 96.2 & 99.2 & 96.9 \\
  C & 68.9 & N/A & 47.6 & 46.3 & 54.3 & C & 93.7 & N/A & 91.0 & 91.4 & 92.0 & C & 94.1 & N/A & 92.6 & 97.7 & 94.8 \\
  P & 67.7 & 64.2 & N/A & 32.0 & 54.7 & P & 94.0 & 93.4 & N/A & 90.3 & 92.6 & P & 95.2 & 96.5 & N/A & 96.0 & 95.9 \\
  R & 48.1 & 54.8 & 25.4 & N/A & 42.8 & R & 91.5 & 91.0 & 90.4 & N/A & 91.0 & R & 93.2 & 94.2 & 95.3 & N/A & 94.2 \\
  Avg. ($\downarrow$) & 61.6 & 58.6 & 36.1 & 37.0 & 48.3 & Avg. ($\downarrow$) & 93.1 & 92.8 & 91.0 & 90.8 & 91.9 & Avg. ($\downarrow$) & 94.1 & 95.3 & 94.7 & 97.6 & 95.4 \\

  \hline\hline

  Time & \multicolumn{4}{c}{$t\xrightarrow{\hspace*{2.5cm}}$} & & Time & \multicolumn{4}{c}{$t\xrightarrow{\hspace*{2.5cm}}$} & & Time & \multicolumn{4}{c}{$t\xrightarrow{\hspace*{2.5cm}}$} &  \\
\end{tabular}
 }
 }
 \VspaceAfter
\end{table*}

\begin{table*}[!htbp]
  \caption{Detailed classification error when adapting the model to test streams of \setting on {\bf DomainNet} with $\bm{\gamma = 10^{-1}}$.}
  \VspaceBefore
  
 \setlength{\tabcolsep}{0.085em}
 \resizebox{\linewidth}{!}{
 \renewcommand{\arraystretch}{0.9}
 {
 \centering
 
 \begin{tabular}{c|c c c c c c c || c | c c c c c c c || c |c c c c c c c}
  \toprule[1.2pt]

  Time & \multicolumn{6}{c}{$t\xrightarrow{\hspace*{3.5cm}}$} & & Time & \multicolumn{6}{c}{$t\xrightarrow{\hspace*{3.5cm}}$} & & Time & \multicolumn{6}{c}{$t\xrightarrow{\hspace*{3.5cm}}$} &  \\

  \hline

  Source & \it{clp} & \it{inf} & \it{pnt} & \it{qdr} & \it{rel} & \it{skt} & Avg. ($\downarrow$) & BN & \it{clp} & \it{inf} & \it{pnt} & \it{qdr} & \it{rel} & \it{skt} & Avg. ($\downarrow$) & TENT & \it{clp} & \it{inf} & \it{pnt} & \it{qdr} & \it{rel} & \it{skt} & Avg. ($\downarrow$) \\
  \midrule
  \it{clp} & N/A & 87.5 & 70.6 & 90.7 & 57.3 & 61.7 & 73.6 & \it{clp} & N/A & 90.4 & 70.5 & 88.4 & 58.7 & 62.6 & 74.1 & \it{clp} & N/A & 97.3 & 99.1 & 99.4 & 99.6 & 99.6 & 99.0 \\
  \it{inf} & 74.1 & N/A & 78.7 & 96.9 & 64.9 & 77.3 & 78.4 & \it{inf} & 78.2 & N/A & 80.3 & 97.2 & 69.2 & 79.3 & 80.8 & \it{inf} & 94.5 & N/A & 99.1 & 99.6 & 99.7 & 99.5 & 98.5 \\
  \it{pnt} & 65.0 & 85.9 & N/A & 96.9 & 52.2 & 65.8 & 73.2 & \it{pnt} & 65.1 & 88.6 & N/A & 94.5 & 53.8 & 66.6 & 73.7 & \it{pnt} & 88.0 & 99.7 & N/A & 99.7 & 99.8 & 99.7 & 97.4 \\
  \it{qdr} & 89.6 & 99.3 & 98.3 & N/A & 95.2 & 91.5 & 94.8 & \it{qdr} & 77.3 & 98.5 & 93.5 & N/A & 88.0 & 86.3 & 88.7 & \it{qdr} & 91.9 & 99.6 & 99.7 & N/A & 99.7 & 99.8 & 98.1 \\
  \it{rel} & 56.8 & 82.5 & 59.8 & 94.8 & N/A & 67.5 & 72.3 & \it{rel} & 59.5 & 87.0 & 62.0 & 94.1 & N/A & 66.8 & 73.9 & \it{rel} & 87.1 & 99.5 & 99.7 & 99.7 & N/A & 99.6 & 97.1 \\
  \it{skt} & 55.3 & 89.0 & 69.7 & 89.0 & 60.1 & N/A & 72.6 & \it{skt} & 55.0 & 90.0 & 67.4 & 89.8 & 57.2 & N/A & 71.9 & \it{skt} & 82.6 & 99.5 & 99.7 & 99.9 & 99.6 & N/A & 96.3 \\
\end{tabular}
 }
 }
 \VspaceAfter
\end{table*}

\begin{table*}[!htbp]
  \caption{Detailed classification error when adapting the model to test streams of \setting on {\bf DomainNet} with $\bm{\gamma = 10^{-2}}$.}
  \VspaceBefore
  
 \setlength{\tabcolsep}{0.085em}
 \resizebox{\linewidth}{!}{
 \renewcommand{\arraystretch}{0.9}
 {
 \centering
 
 \begin{tabular}{c|c c c c c c c || c | c c c c c c c || c |c c c c c c c}
  \toprule[1.2pt]

  Time & \multicolumn{6}{c}{$t\xrightarrow{\hspace*{3.5cm}}$} & & Time & \multicolumn{6}{c}{$t\xrightarrow{\hspace*{3.5cm}}$} & & Time & \multicolumn{6}{c}{$t\xrightarrow{\hspace*{3.5cm}}$} &  \\

  \hline

  Source & \it{clp} & \it{inf} & \it{pnt} & \it{qdr} & \it{rel} & \it{skt} & Avg. ($\downarrow$) & BN & \it{clp} & \it{inf} & \it{pnt} & \it{qdr} & \it{rel} & \it{skt} & Avg. ($\downarrow$) & TENT & \it{clp} & \it{inf} & \it{pnt} & \it{qdr} & \it{rel} & \it{skt} & Avg. ($\downarrow$) \\
  \midrule
  \it{clp} & N/A & 87.6 & 69.1 & 90.3 & 57.5 & 60.5 & 73.0 & \it{clp} & N/A & 92.6 & 79.5 & 92.5 & 73.7 & 74.7 & 82.6 & \it{clp} & N/A & 98.2 & 99.2 & 99.6 & 99.5 & 99.8 & 99.2 \\
  \it{inf} & 76.2 & N/A & 77.9 & 96.9 & 63.8 & 77.1 & 78.4 & \it{inf} & 86.9 & N/A & 87.0 & 98.1 & 82.0 & 86.8 & 88.2 & \it{inf} & 97.3 & N/A & 99.4 & 99.8 & 99.7 & 99.5 & 99.1 \\
  \it{pnt} & 66.6 & 85.2 & N/A & 97.1 & 51.9 & 66.9 & 73.5 & \it{pnt} & 77.9 & 91.1 & N/A & 96.5 & 70.4 & 78.9 & 83.0 & \it{pnt} & 95.1 & 99.6 & N/A & 99.7 & 99.8 & 99.4 & 98.7 \\
  \it{qdr} & 89.7 & 99.3 & 98.3 & N/A & 95.1 & 91.8 & 94.9 & \it{qdr} & 82.8 & 98.6 & 94.9 & N/A & 91.0 & 90.3 & 91.5 & \it{qdr} & 95.6 & 99.7 & 99.8 & N/A & 99.7 & 99.4 & 98.8 \\
  \it{rel} & 58.0 & 81.6 & 60.0 & 94.3 & N/A & 66.9 & 72.2 & \it{rel} & 73.8 & 90.2 & 75.0 & 95.8 & N/A & 77.5 & 82.5 & \it{rel} & 95.4 & 99.8 & 99.8 & 99.7 & N/A & 99.8 & 98.9 \\
  \it{skt} & 56.6 & 88.1 & 70.3 & 88.0 & 60.0 & N/A & 72.6 & \it{skt} & 70.9 & 91.9 & 78.1 & 93.3 & 71.5 & N/A & 81.1 & \it{skt} & 93.4 & 99.8 & 99.6 & 99.8 & 99.8 & N/A & 98.5 \\
\end{tabular}
 }
 }
 \VspaceAfter
\end{table*}

\begin{table*}[!htbp]
  \caption{Detailed classification error when adapting the model to test streams of \setting on {\bf DomainNet} with $\bm{\gamma = 10^{-3}}$.}
  \VspaceBefore
  
 \setlength{\tabcolsep}{0.085em}
 \resizebox{\linewidth}{!}{
 \renewcommand{\arraystretch}{0.9}
 {
 \centering
 
 \begin{tabular}{c|c c c c c c c || c | c c c c c c c || c |c c c c c c c}
  \toprule[1.2pt]

  Time & \multicolumn{6}{c}{$t\xrightarrow{\hspace*{3.5cm}}$} & & Time & \multicolumn{6}{c}{$t\xrightarrow{\hspace*{3.5cm}}$} & & Time & \multicolumn{6}{c}{$t\xrightarrow{\hspace*{3.5cm}}$} &  \\

  \hline

  Source & \it{clp} & \it{inf} & \it{pnt} & \it{qdr} & \it{rel} & \it{skt} & Avg. ($\downarrow$) & BN & \it{clp} & \it{inf} & \it{pnt} & \it{qdr} & \it{rel} & \it{skt} & Avg. ($\downarrow$) & TENT & \it{clp} & \it{inf} & \it{pnt} & \it{qdr} & \it{rel} & \it{skt} & Avg. ($\downarrow$) \\
  \midrule
  \it{clp} & N/A & 86.8 & 69.6 & 92.5 & 58.0 & 61.9 & 73.8 & \it{clp} & N/A & 96.1 & 91.7 & 97.2 & 90.3 & 90.0 & 93.1 & \it{clp} & N/A & 99.2 & 99.3 & 99.8 & 99.9 & 99.4 & 99.5 \\
  \it{inf} & 70.6 & N/A & 81.7 & 96.3 & 63.2 & 76.1 & 77.6 & \it{inf} & 95.0 & N/A & 95.5 & 98.9 & 93.8 & 95.1 & 95.6 & \it{inf} & 99.1 & N/A & 99.1 & 99.7 & 99.7 & 99.6 & 99.5 \\
  \it{pnt} & 64.2 & 86.3 & N/A & 97.5 & 51.7 & 67.1 & 73.3 & \it{pnt} & 91.8 & 95.8 & N/A & 98.6 & 89.7 & 91.9 & 93.6 & \it{pnt} & 98.8 & 99.4 & N/A & 99.5 & 99.0 & 98.7 & 99.1 \\
  \it{qdr} & 89.3 & 99.3 & 98.6 & N/A & 94.7 & 92.1 & 94.8 & \it{qdr} & 91.6 & 99.0 & 97.1 & N/A & 95.4 & 94.1 & 95.4 & \it{qdr} & 98.5 & 99.7 & 100.0 & N/A & 99.6 & 99.9 & 99.5 \\
  \it{rel} & 56.0 & 83.7 & 60.2 & 93.6 & N/A & 66.4 & 72.0 & \it{rel} & 90.5 & 95.4 & 90.5 & 98.2 & N/A & 91.8 & 93.3 & \it{rel} & 98.6 & 99.9 & 99.3 & 100.0 & N/A & 99.5 & 99.5 \\
  \it{skt} & 53.7 & 88.7 & 72.7 & 89.0 & 57.2 & N/A & 72.3 & \it{skt} & 89.9 & 95.6 & 92.0 & 97.3 & 89.4 & N/A & 92.8 & \it{skt} & 98.3 & 99.8 & 99.6 & 99.9 & 100.0 & N/A & 99.5 \\
\end{tabular}
 }
 }
 \VspaceAfter
\end{table*}

\begin{table*}[!htbp]
  \caption{Detailed classification error when adapting the model to test streams of \setting on {\bf DomainNet} with $\bm{\gamma = 10^{-4}}$.}
  \VspaceBefore
  
 \setlength{\tabcolsep}{0.085em}
 \resizebox{\linewidth}{!}{
 \renewcommand{\arraystretch}{0.9}
 {
 \centering
 
 \begin{tabular}{c|c c c c c c c || c | c c c c c c c || c |c c c c c c c}
  \toprule[1.2pt]

  Time & \multicolumn{6}{c}{$t\xrightarrow{\hspace*{3.5cm}}$} & & Time & \multicolumn{6}{c}{$t\xrightarrow{\hspace*{3.5cm}}$} & & Time & \multicolumn{6}{c}{$t\xrightarrow{\hspace*{3.5cm}}$} &  \\

  \hline

  Source & \it{clp} & \it{inf} & \it{pnt} & \it{qdr} & \it{rel} & \it{skt} & Avg. ($\downarrow$) & BN & \it{clp} & \it{inf} & \it{pnt} & \it{qdr} & \it{rel} & \it{skt} & Avg. ($\downarrow$) & TENT & \it{clp} & \it{inf} & \it{pnt} & \it{qdr} & \it{rel} & \it{skt} & Avg. ($\downarrow$) \\
  \midrule
  \it{clp} & N/A & 87.8 & 70.2 & 89.4 & 55.7 & 63.7 & 73.3 & \it{clp} & N/A & 97.2 & 95.6 & 98.1 & 94.3 & 94.3 & 95.9 & \it{clp} & N/A & 99.4 & 99.6 & 99.5 & 99.5 & 99.4 & 99.5 \\
  \it{inf} & 73.3 & N/A & 77.9 & 94.8 & 64.7 & 79.7 & 78.1 & \it{inf} & 97.3 & N/A & 97.5 & 99.3 & 97.0 & 97.5 & 97.7 & \it{inf} & 99.2 & N/A & 99.4 & 99.4 & 99.6 & 98.7 & 99.2 \\
  \it{pnt} & 65.2 & 84.9 & N/A & 96.7 & 55.5 & 69.0 & 74.3 & \it{pnt} & 95.6 & 96.8 & N/A & 99.0 & 94.5 & 95.9 & 96.3 & \it{pnt} & 98.8 & 99.4 & N/A & 99.5 & 99.3 & 99.9 & 99.4 \\
  \it{qdr} & 88.5 & 99.6 & 98.6 & N/A & 95.5 & 92.7 & 95.0 & \it{qdr} & 93.9 & 99.3 & 97.7 & N/A & 96.5 & 96.1 & 96.7 & \it{qdr} & 98.6 & 99.9 & 99.8 & N/A & 100.0 & 99.5 & 99.6 \\
  \it{rel} & 55.3 & 82.8 & 58.7 & 94.0 & N/A & 67.3 & 71.6 & \it{rel} & 94.6 & 96.5 & 94.2 & 98.9 & N/A & 95.1 & 95.9 & \it{rel} & 98.5 & 99.7 & 99.7 & 100.0 & N/A & 98.7 & 99.3 \\
  \it{skt} & 52.3 & 87.6 & 70.4 & 89.4 & 59.3 & N/A & 71.8 & \it{skt} & 94.5 & 96.7 & 95.0 & 98.3 & 94.4 & N/A & 95.8 & \it{skt} & 98.6 & 99.4 & 99.9 & 99.8 & 99.9 & N/A & 99.5 \\
\end{tabular}
 }
 }
 \VspaceAfter
\end{table*}

\end{document}